\documentclass[sigconf,nonacm]{acmart}

\AtBeginDocument{%
  }

\setcopyright{none}
\renewcommand\footnotetextcopyrightpermission[1]{}

\usepackage{microtype}
\usepackage{float}
\usepackage{graphicx}
\usepackage{booktabs}
\usepackage{tabularx}
\usepackage{longtable}
\usepackage{multirow}
\usepackage{colortbl}
\usepackage{amsmath}
\usepackage{mathtools}
\usepackage{amsthm}
\usepackage{newunicodechar}
\usepackage[capitalize,noabbrev]{cleveref}

\newcolumntype{L}{>{\raggedright\arraybackslash}X}
\newunicodechar{−}{\ensuremath{-}}

\title{The Human Creativity Benchmark}

\author{Aspen Hopkins}
\affiliation{%
  \institution{Contra; Massachusetts Institute of Technology}
  \city{New York, NY; Cambridge, MA}
  \country{USA}}

\author{Allison Nulty}
\affiliation{%
  \institution{Contra}
  \city{New York}
  \state{New York}
  \country{USA}}

\author{Alexandria Minetti}
\affiliation{%
  \institution{Contra}
  \city{New York}
  \state{New York}
  \country{USA}}

\author{Anoop Pakki}
\affiliation{%
  \institution{Contra}
  \city{New York}
  \state{New York}
  \country{USA}}

\author{Angad Singh}
\affiliation{%
  \institution{Contra}
  \city{New York}
  \state{New York}
  \country{USA}}

\renewcommand{\shortauthors}{The Human Creativity Benchmark}

\begin{document}

\begin{abstract}
Modern AI evaluation frameworks treat evaluator disagreement as noise to be resolved. In creative domains, professional disagreement reflects genuine differences in taste, not measurement error. We argue that evaluating creative AI requires preserving two distinct signals: \textbf{convergence}, where professionals align around shared best practices, and \textbf{divergence}, where individual taste legitimately varies. We present the Human Creativity Benchmark (HCB), a benchmark that operationalizes this separation by collecting pairwise preferences, scalar ratings on prompt adherence, usability, and visual appeal, and qualitative rationale from domain professionals. Across 15,000 professional judgments spanning five creative domains and three workflow phases (ideation, mockup, refinement), we find that convergence concentrates on verifiable dimensions like technical correctness and visual hierarchy, while divergence concentrates on taste-driven dimensions like aesthetic direction and conceptual risk. No model excels uniformly across all phases. Collapsing these signals into a single quality metric discards the most actionable information: where models must be correct versus where they should remain steerable.
\end{abstract}


\begin{teaserfigure}
  \centering
  \includegraphics[width=\textwidth]{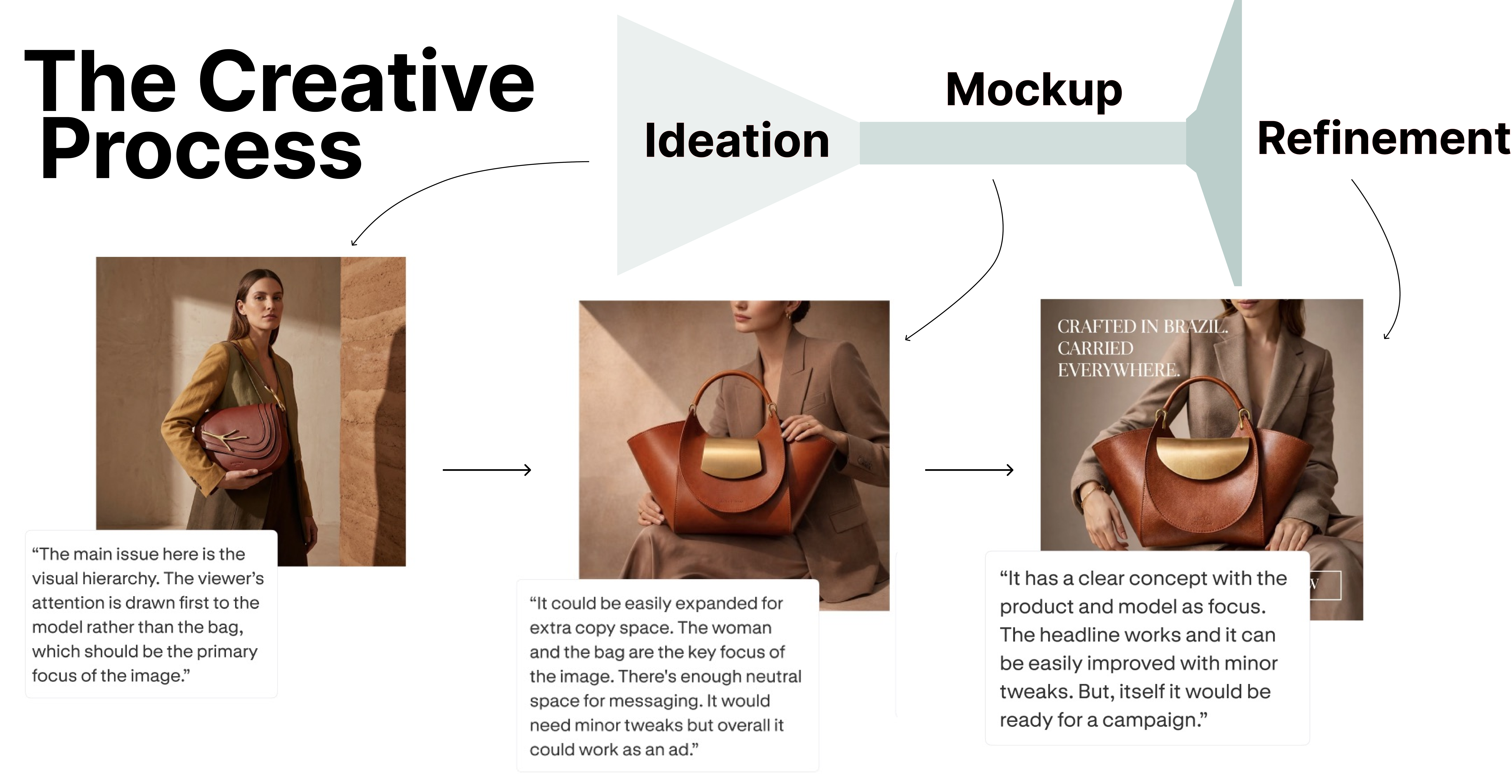}
  \caption{The creative process as a sideways martini: broad ideation narrows through mockup, ending in refinement. Example stills taken from Ad images rendered in response to prompts; prompts were derived from professional samples.}
  \label{fig:creative-overview}
\end{teaserfigure}

\maketitle

% \section{running list of to-dos}
% \begin{center}
    % \textcolor{red}{\textbf{TODOs before camera ready}}
    % \end{center}
    % \begin{itemize}
    %     \color{red}
    %     \item flesh out results section w/tldr headlines - Anoop - done
    %     \item swap out main overview figures, get rid of convergence-divergence figure, check that sizing of labels in these images are consistent (e.g., A B should look approximately the same across all figures.) Check all figure text is legible / not too small. - Aspen 
    %     \item need to emphasize further the idea that different steps in creative processes impose different requirements, constraints, needs - Aspen
    %     \item Currently working on discussion, but will wait to finish until results are updated! - Aspen
    %     \item update abstract with finished results + discussion tldr sentence. - Aspen
    % \end{itemize}

\section{Introduction}
Creativity has long resisted reduction to any single, stable operational definition. This has made measuring creativity challenging. Creative \textit{work}, which connects creative process and domain expertise with expected outcomes, is similarly heterogeneous and remains difficult to study. This continues in the age of AI.

% \textcolor{red}{Creative work is heterogeneous, task-dependent, and shaped by individual expertise and taste. A designer's output reflects not only technical competence but aesthetic judgment accumulated over years of practice---the perspective that makes one creative's work distinguishable from another's.}
A designer's output reflects not only technical competence but aesthetic judgment accumulated over years of practice. When professional creatives evaluate AI-generated work, they do so across potentially many axes. For some axes, evaluators agree on what works---readable typography, functional layout, correct visual hierarchy---revealing shared professional standards. For others, evaluators legitimately disagree: not because of measurement error, but because the criteria are personal---taste, aesthetic direction, or creative intent (Figure~\ref{fig:example-output}). AI benchmarks treat this disagreement as noise to be resolved. 
% Instead, it can offer important signal that is missed in current evaluation practices.

% This paper argues that both types of judgment must be preserved: professional disagreement is evidence of where models need to remain steerable rather than optimized toward any single creative direction.

In this work, we argue that professional disagreement is not \textit{always} noise to be resolved in creative domains. While some aspects of creative quality are objectively verifiable, others can and should be treated as inherently subjective. Our goal in creative evaluation is thus to not only ask if an output is ``good,'' but \textit{good according to whom, for what purpose, and at what stage of the creative process?} 

To this end, we contribute a framework that distinguishes \textbf{\textit{convergence}} and \textbf{\textit{divergence}} in creative responses as a way to preserve both kinds of judgment.\footnote{We use \textit{divergence} rather than \textit{disagreement} to avoid treating variation in expert judgment as implicitly negative.} Convergence describes dimensions where evaluators align around shared, checkable criteria; divergence describes dimensions where multiple creative judgments can be valid at once. We operationalize this distinction across three evaluation axes, \textit{Prompt Adherence}, \textit{Usability}, and \textit{Visual Appeal}, which vary in the extent to which expert judgments might converge around shared criteria or diverge with individual preference.

To test our framework, we introduce \textbf{The Human Creativity Benchmark (HCB)}, an exemplar dataset developed from an exploratory survey on how domain experts use AI in creative work. The HCB is designed to mirror real creative workflows across Ideation, Mockup, and Refinement stages (Figure~\ref{fig:creative-overview}) in five domains: landing pages, desktop apps, ad images, brand images, and product videos (Table~\ref{table:participant-data}).

Domain experts drawn from Contra, a network of independent professional creatives, evaluated the HCB's generated outputs. Their roughly 15,000 judgments span pairwise forced-rankings, scalar ratings on prompt adherence, usability, and visual appeal, and open-ended qualitative explanations. Through our analysis, we find evidence that creative AI outputs cannot be reduced to a single judgement of ``good'' or ``bad.'' Instead, what is defined as valuable varies by domain, by evaluation criterion, and in some cases, by individual preference. Separating these sources of judgment clarifies what model developers should optimize for: models should be reliably consistent on convergent axes while remaining steerable on divergent axes. 

In sum, we present three contributions:

\begin{figure*}[t]
    \centering
    \normalsize
    \setlength{\tabcolsep}{5pt}
    \renewcommand{\arraystretch}{1.3}
    
    \begin{minipage}[t]{0.315\textwidth}
    \begin{tabularx}{\linewidth}{|X|}
    \hline
    \cellcolor{gray!20}\textbf{Text-to-image / image-to-image} \\
    \hline
    Brand Design \textit{(Brand Image Assets)}\newline
    Content Designers \textit{(Ad Images)} \\
    \hline
    \texttt{gpt-image-1.5}\newline
    \texttt{gemini-3-pro-image}\newline
    \texttt{seedream-4.5}\newline
    \texttt{flux-2-pro} \\
    \hline
    \end{tabularx}
    \end{minipage}\hfill
    \begin{minipage}[t]{0.315\textwidth}
    \begin{tabularx}{\linewidth}{|X|}
    \hline
    \cellcolor{gray!20}\textbf{Image-to-video} \\
    \hline
    Video Editors \textit{(Product Video)} \\
    \hline
    \texttt{veo3.1}\newline
    \texttt{kling-v3.0-pro}\newline
    \texttt{seedance-v1.5-pro}\newline
    \texttt{grok-imagine-video} \\
    \hline
    \end{tabularx}
    \end{minipage}\hfill
    \begin{minipage}[t]{0.315\textwidth}
    \begin{tabularx}{\linewidth}{|X|}
    \hline
    \cellcolor{gray!20}\textbf{Text-to-code / code-to-code} \\
    \hline
    Product Designers \textit{(Desktop Applications)}\newline
    Web Designers \textit{(Landing Pages)} \\
    \hline
    \texttt{claude-opus-4.6}\newline
    \texttt{gemini-3.1-pro}\newline
    \texttt{gpt-5.3-codex}\newline
    \texttt{qwen3.5-397b} \\
    \hline
    \end{tabularx}
    \end{minipage}
    
    \refstepcounter{table}
    \label{table:participant-data}
    \par\vspace{0.5em}
    {\small\textbf{Table~\thetable:} Creative professional domains, output types, model modalities, and the specific models evaluated in this study. The benchmark spans image, video, and code generation settings, pairing each modality with the creative specialties that commonly produce those deliverables and the frontier models selected for comparison.}

    \vspace{0.75em}
    \includegraphics[width=\textwidth]{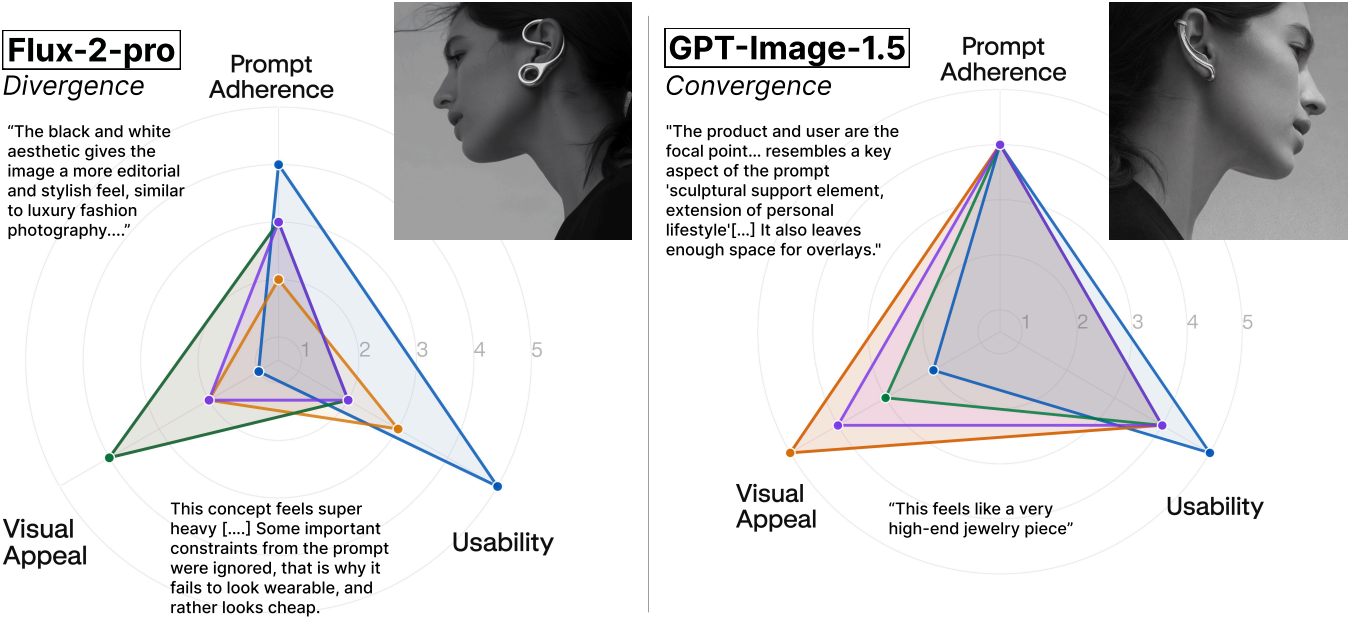}
    \caption{Inter-rater divergence vs.\ convergence on a single output. Examples of two model outputs for the same creative stage (Ideation) and prompt that illustrate divergence (\textit{left, Flux-2-Pro}) and convergence (\textit{right, GPT-Image-1.5}) for Ad Images. Tightly nested polygons indicate reviewer agreement and widely spread polygons indicate disagreement. For Flux-2-Pro, creatives \emph{diverge}: scores span nearly the full scale on Usability (2 -- 5) and Visual Appeal (1 -- 4), with the same image read as ``editorial and stylish'' like luxury fashion photography by one reviewer and as ``super heavy\ldots\ rather looks cheap'' by another. For GPT-Image-1.5, creatives \emph{converge}, most strongly on Prompt Adherence (unanimous) and Usability, with consistently favorable comments (``a very high-end jewelry piece; leaves enough space for overlays''). }
    \label{fig:example-output}
\end{figure*}

% The Human Creativity Benchmark (HCB) operationalizes this argument, capturing both signals through three complementary methods. Pairwise forced-ranking surfaces relative preference. Scalar ratings on prompt adherence, usability, and visual appeal surface where agreement concentrates. Open-ended qualitative follow-ups surface the reasoning behind each judgment. Together they produce data that distinguishes where a model must be correct from where it must be steerable---the most actionable distinction for practitioners that a single quality score would collapse.

\begin{enumerate}
    \item \textbf{The Human Creativity Benchmark (HCB)}, a dataset of prompt and sample images for creative work built with prompts seeded by real creatives' work artifacts at different steps of the creative workflow, including Ideation, Mockup, and Refinement, 
    along with domain-expert labels and annotations combining pairwise forced-ranking, scalar ratings, and open-ended qualitative responses.
    \item A framework differentiating \textbf{Convergence} and \textbf{Divergence} in creative outputs across three axes: \textbf{Visual Appeal}, \textbf{Prompt Adherence}, and \textbf{Usability}.
    \item An analysis of HCB results showing where expert judgments converge around shared best practices, where they diverge along subjective dimensions of creative quality, and how these patterns shift across the creative workflow.
\end{enumerate}

\section{Related Work}

% \subsection{Measuring Creativity}
Our interest in understanding the intersection of creative work and AI begins with the premise that creativity is heterogeneous, task-dependent, and unevenly expressed across stages of work. Creative producers may view aesthetic value and commercial value as harmonious or contradictory~\cite{lee2022}. They combine multiple modes of cognition in their process of production, and their perceptions of identity, such as genre, status, intellect, and profession shape their outputs. Broadly, creatives balance what is good (best practices) with something more personal, fundamental: taste.  

% If taste is a fingerprint, then the study of how these tendencies differ may be a close proxy to what underpins creative work; while there are certainly commonalities in what people like, their individual differences are what practically differentiate their outputs.

% \paragraph{AI in Creative Work}
% Creative work is labor-intensive~\cite{casey2020}: the constant iteration that takes domain experts from ideation to refinement to a final deliverable has led to increased adoption of AI tooling~\cite{googlestitch}.
% In parallel, a

As an array of AI-driven tools supporting creative work has emerged, however, creatives seeking to improve their workflows, expand the initial generative exploration, or save time and accumulative costs face frictions in systems that do not account for these otherwise distinct axes.
% \paragraph{Homogenization}
Perhaps most resonantly is the challenge that AI systems are imperfect samplers that tend to converge to an average. This leads to homogenization~\cite{bommasani2022} and mode collapse~\cite{thanhtung2020}, where models increasingly produce similar outputs---characteristics antithetical to the creative process. People spend years exploring, repeating, and improving their points of view, styles, and mediums. Converging their outputs through homogenization risks collapsing the plurality of creative practice, narrowing not only the range of outputs but the points of view available to creative fields.
% This makes steerability and optionality, rather than adherence to any single notion of ``best practices,'' central to creative AI systems.

\subsection{Evaluating AI in Creative Processes}

Existing norms of AI evaluation function largely in settings constrained by rules of what is correct. Benchmarks are generally built for domains where pre-specified heuristics, best practices or ground-truths, equate metrics of ``truthfulness'' or ``correctness'' with ``best''. 
% \textcolor{red}{These concepts do not necessarily hold in domains where creativity is fundamental to work, nor do they writ-large account for complex processes entailed in creative work. For example, what counts as effective practice in ideation may not be what matters in mockup or refinement, while claims about ``how creativity works'' that are informed by specific tasks and measurements may be overly prescriptive when applied to a new domain~\cite{lee2022}.} 
These methods of evaluation do not adequately account for diversity as an outcome and process value ~\cite{lee2022}: in creative work, preserving variation across styles, judgments, or workflows may be part of what makes a system useful in the first place.

A related body of work in data annotation and label adjudication has addressed the broader problem of expert disagreement in evaluation tasks. Methods such as Dawid-Skene modeling~\cite{dawid1979}, CrowdTruth~\cite{inel2014}, and perspectivist annotation frameworks~\cite{basile2021} treat annotator disagreement as informative rather than purely noisy. However, these approaches have been developed primarily for tasks with definable ground truth or narrow classification objectives. Creative evaluation presents a distinct challenge: there is no ground truth to approximate, and the dimensions on which experts disagree (aesthetic direction, mood, conceptual risk) are not reducible to miscalibration or error. The HCB builds on the insight that disagreement can be signal, but applies it to a domain where the standard resolution strategies like reconciliation, majority vote, gold-standard adjudication may be structurally inappropriate.

Prior to designing the benchmark, we conducted a paid formative survey to understand how professional creatives integrate AI tools into their workflows. The survey was fielded in November 2025 to creatives in the Contra network, and yielded 50 responses spanning design, video, development, and content disciplines. 
% The questionnaire covered creative workflows, AI adoption, tool-specific usage across eight generative categories, and attitudes toward AI's role in the future of creative work, its effect on their earning potential, barriers to adoption, and comfort disclosing AI use to clients.
The questionnaire was organized in four parts, exploring (1) their creative process, whether they currently use AI tools, and, if not, reasons against adoption; (2) how they incorporated AI, including how much of their creative process is AI-supported, how much of AI-content reaches final deliverables, and the perceived benefits or costs to adoption; (3) eight modules on  generative tool categories 
% (image generation, video generation, prototyping, code generation, copywriting, audio/voice/music, 3D/motion graphics, and LLM-based research) 
detailing specific tools and where in their workflow they applied them; and (4) beliefs on AI's role in the future of creative work, effects on earnings, barriers to adoption, and comfort disclosing AI use to clients.

At the time of the survey, the majority of respondents reported increased earnings with AI-use (66\% of the respondents), and mixed (skewing positive) attitudes towards the role AI in creative work (AI will enhance creativity (80\% of respondents) and create new forms of creativity (70\% of respondents), versus fewer who felt it would replace some creative tasks (50\% of respondents) or commoditize creativity (24\% of respondents)). Respondents generally adopted a ``co-creation'' or augmentation perspective. As one brand designer stated, \textit{``it’s become a kind of creative partner—helping me get from concept to something visually exciting much quicker,''} but, as another put it, \textit{``[w]ork with me to bring my ideas to life and help me do more, not take the wheel entirely.''} 
% at least at the stage of AI usage that we're in as a society, and furthermore 
% \ref{fig:ADD TO APPENDIX}.
While AI use was not monolithic, those that did use these tools tended to bucket AI use in a sequence of exploratory ideation, prototyping, followed by refining client deliverables.\footnote{ Additional details and data overviews are shown in Survey Insights (Appendix \ref{sec:survey-insights}).}  

Finally, we asked creatives to walk through their prompting process: what output they wanted, how they responded when an output fell short, and how acceptable outputs were incorporated into their own process or client-facing deliverables. Collectively, this data, along with themes of controllability, co-creation, ideation, and process-specific needs motivated the construction in the HCB, including the three-phase decomposition of the creative process (Ideation, Mockup, Refinement) described below.

\section{Methodology}
% \subsection{Creative Process}
The study structures creative workflows into three phases, shown in the teaser figure, validated against a prior survey of working creatives:
\begin{enumerate}
    \item \textbf{Ideation:} Discovery, exploration, and directional potential. At this stage, the creative is not looking for final production quality, but rather for exciting creative direction that is strategically appropriate and worth developing.

    \item \textbf{Mockup:} Creative direction has been decided, now it's time to make the vision come to life. The creative is actualizing the project's creative direction, creating product shots, stitching together scenes, incorporating brand identity, and bringing the campaign to life.

    \item \textbf{Refinement:} Designs are near production-ready, requiring only targeted tweaks to improve consistency across the design.
    \end{enumerate}

All prompts were seeded by expert-produced prompts and media, lightly edited to standardize prompt length and structure. The prompt sequence was designed to mirror a designer’s workflow based on survey responses: Ideation prompts generated new design directions, Mockup prompts used those directions to produce more stable concepts, and Refinement prompts built on the mockups to request specific edits.

% Prompts were built upon the previous phase, using input images for Mockup and Refinement to simulate a real designer's workflow. Ideation Prompts created new design directions, Mockup prompts used that vision and prompted for a more stable direction, and Refinement prompts used that direction and prompted for specific edits. [ASPEN FIX THIS ORDER / WORDING TO BE CLEARER AND SHORTER]
% Creative professionals from the network generated the prompts and input media. 

Participants were drawn from a network of independent professional creatives across design, video, development, and content projects, reflecting common deliverables in independent creative work. Participants were selected based on skillset and the generative model category most relevant to their workflow, then presented with guidelines contextualizing each phase of the creative process and outlining grading criteria for rubric alignment. Selected domains  meaningfully represent different evaluation conditions. For example, Ad images produce a single static composition with defined elements like a headline or product image, whereas a landing page is structurally more complex, with elements like layout and design fidelity competing for prioritization. These differences shape evaluator agreement patterns across phases.
% and why convergence patterns vary by domain. 

In total, after dropping incomplete responses, the study was conducted with 28 evaluators from 13 unique countries (Armenia, Belgium, Brazil, Canada, India, Malaysia, Netherlands, Poland, Portugal, Romania, Spain, United Kingdom, and United States) assessing 93 prompts across 80 sessions, yielding 5,940 pairwise judgments, 5,940 scalar ratings, and 3,675 qualitative responses.\footnote{The complete dataset, including expert responses, is publicly available on Hugging Face at  \url{https://huggingface.co/datasets/contra-labs/HCB} and is released under the \texttt{cc-by-4.0} license. Details of the data structure can be found in Table \ref{tab:dataset-files}.}

% \iffalse
% \begin{figure*}[t]
%     \centering
%     \includegraphics[width=.75\textwidth]{images/axes.pdf}
%     \caption{Schematic illustration of evaluator agreement patterns across two types of creative criteria. Convergent criteria (left) produce directional alignment across evaluators; divergent criteria (right) produce independent, taste-driven judgments.}
%     \label{fig:axes_fin}
% \end{figure*}
% \fi
% Designers were given high-level product and industry information and advised to design output appropriate for their use case. The prompt generation task guided creatives through each phase with baseline structural requirements covering prompt length, camera angles, color palettes, input media, and other domain-relevant attributes. Prompts were reviewed by our research team for clarity and alignment with real-world project briefs, then normalized for consistency. Prompts containing negative sentiment were removed to mitigate potential confounds.

% \subsection{Participants and input data}

% \begin{figure}
%     \centering
%     \includegraphics[width=1.0\linewidth]{images/axes_overview.pdf}
%     \caption{Caption}
%     \label{fig:axes_convergence}
% \end{figure}

\subsection{Evaluation design}

Five evaluators per domain completed six tasks per phase (called tournaments), with each tournament comprising two tasks centered on one prompt. Model ordering was randomized and identity anonymized throughout. All evaluations were conducted within the Contra Labs evaluation environment, a web-based interface that walked raters through each phase of the creative process under controlled conditions (Figure~\ref{fig:interface-extended}). The interface presented each tournament as a single guided flow built around one prompt, moving raters sequentially through pairwise comparison tasks,  scaled-rating tasks, and free-text responses. Additional information and sample interface views are available in the Appendix (Section~\ref{appendix:interface}).

\textbf{Task 1: Pairwise comparison.} Raters were presented with two outputs side-by-side across all possible pairings, producing six pairwise judgments per prompt. Rather than scoring against a predefined rubric, raters selected the output they preferred, isolating the subjective judgment a creative professional would actually apply in practice. After each selection, raters described the rationale in their choice. Pairwise results were aggregated using a Bradley-Terry model to produce ELO ratings for each model.

\textbf{Task 2: Scalar ratings.} Three Likert-scale subtasks were chosen to span the convergence--divergence spectrum, helping identify where models should be reliably correct and where they should remain steerable:
\label{sec:axes}

\begin{itemize}
    \item \textbf{Prompt Adherence:} How faithful is this output to the given prompt? The least subjective of the three scales, grounded in whether a model did what was asked.
    \item \textbf{Usability:} How well does this output function in the context of the prompt and campaign? This measures whether an output could realistically be used in a professional context.
    \item \textbf{Visual Appeal:} How visually interesting, cohesive, and polished is this output? This dimension targets taste: the aesthetic judgment that distinguishes work a creative would choose rather than merely accept.
\end{itemize}

\section{Hypotheses}
We structure this benchmark and our analyses in response to three hypotheses, motivated in part by our survey responses. 

\paragraph{\textit{Hypothesis \#1}: Convergence emerges when evaluating verifiable criteria; divergence emerges when tastes are misaligned.} While some axes of creative evaluation will reduce to verifiable metrics of performance (convergence), others will be intrinsically based on individual preferences (divergence). In such cases, divergence occurs when outputs are technically acceptable (or uniformly unacceptable) and evaluators can prioritize taste or creative direction. Figure~\ref{fig:example-output} previews this pattern for a single Ad Images prompt: reviewers converge on GPT-Image-1.5 and diverge on Flux-2-Pro across Usability and Visual Appeal.
% Our expectatation when a criterion is verifiable, we expect responses to concentrate; where it is taste-driven, ratings spread legitimately. The three scalar axes were chosen to span this predicted spectrum. 
We adopt the scalar axes listed above to assess this hypothesis. We expect \textbf{Prompt Adherence} to illustrate the greatest evaluator convergence because its criteria are checkable, and \textbf{Visual Appeal} to be the most divergent it lends to individual preference and stylistic differences; finally, \textbf{Usability} should fall between the two, combining shared professional standards with context-specific judgment. We treat inter-rater agreement as the operational indication of convergence, which should be further validated by context provided in free-text responses.

\paragraph{\textit{Hypothesis \#2}: Creative requirements for model outputs will change across the creative stages.} Creative work will involve different requirements across stages (Figure~\ref{fig:creative-overview}). During Ideation, we expect creatives will prefer outputs that are more generative; during Mockup, outputs that develop those directions into coherent specifications are rewarded; finally, Refinement evaluations will prioritize targeted correction, consistency, and polish. We thus expect model rankings, scalar labels, and evaluator agreement to shift across stages, as what is useful will change over the course of the workflow.

% Workflow phase changes the balance between convergence and divergence, so creative AI evaluation should measure both standards 

\paragraph{\textit{Hypothesis \#3}: Model comparison should capture convergence and divergence, not just overall quality.} If creative quality were one-dimensional, model comparison would reduce to a stable ranking of better and worse outputs, e.g., which model produces better outputs on average. We instead expect models to differ in how they satisfy shared standards and support divergent preferences: some models may be more reliable on convergent criteria such as prompt fidelity or production constraints, while others may better support varied creative directions. A convergence--divergence analysis would therefore surface meaningful differences in model behavior that disappear in a single overall ranking.

% models may differ in where they generate consensus and where they enable disagreement. Some may excel on shared standards such as prompt fidelity, coherence, or production feasibility, while others may better accommodate divergent tastes, interpretations, or creative goals. 

% If creative quality were reducible to a single dimension, then evaluating models would simply require identifying .

% Preserving both convergence and divergence should therefore surface model behaviors that a single aggregate score would collapse.

% We organize these results around a single hypothes
%          +is: evaluator agreement is governed less by the identity of a g
%          +iven model or domain than by the \emph{nature of the criterion}
%          + under evaluation. Following the convergence--divergence framew
%          +ork, we treat inter-rater agreement---Kendall's $W$ across eval
%          +uators and Krippendorff's $\alpha$ within models---as the opera
%          +tional signature of convergence. Judgments grounded in objectiv
%          +ely verifiable criteria concentrate (convergence), whereas judg
%          +ments grounded in aesthetic preference, among outputs that have
%          + already met a threshold of technical competence, spread legiti
%          +mately (divergence), reflecting variation in professional taste
%          + rather than measurement error.  

\subsection{Analysis}
Pairwise preference data was aggregated using a Bradley-Terry model to produce ELO ratings by domain and phase. Scalar ratings were analyzed across all three dimensions, with Kendall's W quantifying evaluator agreement at each phase. We additionally computed Krippendorff's $\alpha$ for each model within each domain and phase to assess inter-evaluator reliability and applied the Friedman test independently within each domain and phase to evaluate whether scalar ratings differed significantly among the four competing models.

% Qualitative feedback was analyzed in two steps. First, light coding was applied to all responses to surface re-occurring themes and to contextualize scalar and pair-wise comparisons. Second, we stripped all qualitative feedback of personally identifiable information and model identifiers. To normalize and parse raw text into structured data frames for cross-domain and cross-phase analysis, the responses were then processed through a deductive coding pass using GPT-4o with a predefined codebook, built in our first step, returning themes, per-theme sentiment, and key quotes.  

Qualitative feedback was analyzed in several stages. First, we applied light coding to all responses to surface recurring themes and to contextualize the scalar and pairwise comparisons; this pass also produced the codebook used in later analysis. Next, we stripped all qualitative feedback of personally identifiable information and model identifiers and ran the responses through a deductive coding pass using GPT-4o and the predefined codebook, which returned themes, per-theme sentiment, and key quotes, normalizing and parsing the raw text into structured data frames for cross-domain and cross-phase analysis. We then completed additional manual iteration over the raw text to refine and validate key findings.

\section{Results}
\begin{figure*}[t]
    \centering
    \begin{minipage}{\textwidth}
        \centering
        \includegraphics[width=\linewidth]{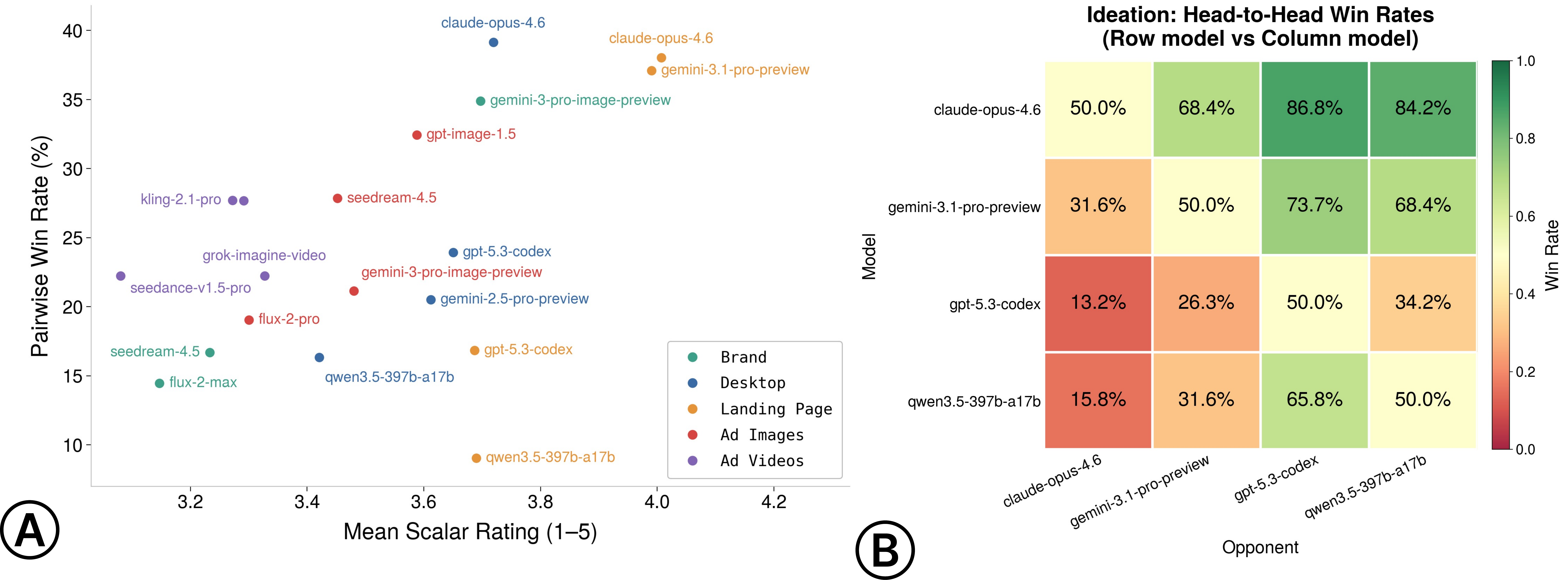}
        \caption{(A) \textit{Scalar ratings and pairwise win rates across domains.} Each point represents a model within a domain, plotted by its mean scalar quality rating (1--5 scale, horizontal axis) against its pairwise win rate (\%, vertical axis) aggregated over all head-to-head comparisons in that domain. The aggregation removes phase- and axis-level differences. (B) Head-to-head pairwise win rates in the Ideation stage for product-video models; each cell is the row model's win rate against the column model (diagonal self-matches excluded). Grok-Imagine-Video wins 50\% against Seedance-v1.5-pro, 47.2\% against Kling-v3.0-pro, and 41.7\% against Veo3.1. Kling-v3.0-pro wins most against Seedance-v1.5-pro (63.9\%), followed by Grok-Imagine-Video (52.8\%) and Veo3.1 (36.1\%). Seedance-v1.5-pro takes 50\% against Grok-Imagine-Video but loses to Veo3.1 (38.9\%) and Kling-v3.0-pro (36.1\%). Veo3.1 is the clear leader, winning all three of its matchups—63.9\% against Kling-v3.0-pro, 61.1\% against Seedance-v1.5-pro, and 58.3\% against Grok-Imagine-Video.}
        \label{fig:overview_dots}
    \end{minipage}
\end{figure*}

Figure~\ref{fig:overview_dots} shows an aggregate view of the HCB evaluation: each model-domain pair is reduced to a mean scalar rating and an aggregate pairwise win rate. This view is useful, but obfuscates the questions our hypotheses raise---including the axis-level convergence and divergence exemplified in Figure~\ref{fig:example-output}. The results below ask whether evaluation axes separate verifiable criteria from taste-driven criteria, whether creative requirements shift across Ideation, Mockup, and Refinement, and whether model comparisons change once axes and stages are kept separate. 

% It does not distinguish Prompt Adherence, Usability, and Visual Appeal; it aggregates across Ideation, Mockup, and Refinement; 

\begin{table*}[t]
    \centering
    \normalsize
    \setlength{\tabcolsep}{5pt}
    \renewcommand{\arraystretch}{1.3}
    \arrayrulecolor{black}
    \begin{tabularx}{\textwidth}{|l|l|X|X|X|}
    \hline
    \cellcolor{gray!20}\textbf{Domain} & \cellcolor{gray!20}\textbf{Model} & \cellcolor{gray!20}\textbf{Ideation Stage} & \cellcolor{gray!20}\textbf{Mockup Stage} & \cellcolor{gray!20}\textbf{Refinement Stage} \\
    \hline
    \multirow{4}{*}{\textbf{Desktop Apps}}
        & \texttt{Claude 4.6} & 0.600 \textit{(0.183; 0.0010)} & 0.400 \textit{(0.101; 0.3231)} & 0.171 \textit{(-0.042; 0.1134)} \\
        & \texttt{GPT 5.3} & 0.086 \textit{(0.568; 0.0010)} & 0.229 \textit{(0.116; 0.3231)} & 0.400 \textit{(0.119; 0.1134)} \\
        & \texttt{Gemini 3.1} & 0.257 \textit{(0.148; 0.0010)} & 0.171 \textit{(-0.031; 0.3231)} & 0.200 \textit{(0.243; 0.1134)} \\
        & \texttt{Qwen 3.5} & 0.057 \textit{(-0.066; 0.0010)} & 0.200 \textit{(0.231; 0.3231)} & 0.229 \textit{(0.234; 0.1134)} \\
    \hline
    \multirow{4}{*}{\textbf{Landing Pages}}
        & \texttt{Claude 4.6} & 0.533 \textit{(-0.060; 0.0000)} & 0.300 \textit{(0.072; 0.3697)} & 0.300 \textit{(-0.076; 0.4678)} \\
        & \texttt{Gemini 3.1} & 0.367 \textit{(-0.058; 0.0000)} & 0.467 \textit{(0.040; 0.3697)} & 0.267 \textit{(0.039; 0.4678)} \\
        & \texttt{GPT 5.3} & 0.100 \textit{(0.075; 0.0000)} & 0.167 \textit{(-0.011; 0.3697)} & 0.233 \textit{(0.209; 0.4678)} \\
        & \texttt{Qwen 3.5} & 0.000 \textit{(-0.043; 0.0000)} & 0.067 \textit{(0.165; 0.3697)} & 0.200 \textit{(0.026; 0.4678)} \\
    \hline
    \multirow{4}{*}{\textbf{Brand Assets}}
        & \texttt{Gemini 3 Image} & 0.381 \textit{(0.590; 0.0010)} & 0.500 \textit{(0.041; 0.0015)} & 0.417 \textit{(-0.031; 0.0033)} \\
        & \texttt{GPT Image 1.5} & 0.333 \textit{(0.054; 0.0010)} & 0.208 \textit{(0.280; 0.0015)} & 0.361 \textit{(0.353; 0.0033)} \\
        & \texttt{Seedream 4.5} & 0.119 \textit{(0.151; 0.0010)} & 0.167 \textit{(0.236; 0.0015)} & 0.139 \textit{(0.049; 0.0033)} \\
        & \texttt{Flux 2 Max} & 0.167 \textit{(0.135; 0.0010)} & 0.125 \textit{(0.770; 0.0015)} & 0.083 \textit{(0.343; 0.0033)} \\
    \hline
    \multirow{4}{*}{\textbf{Ad Images}}
        & \texttt{GPT Image 1.5} & 0.333 \textit{(0.476; 0.0010)} & 0.400 \textit{(0.183; 0.0056)} & 0.233 \textit{(0.112; 0.0398)} \\
        & \texttt{Seedream 4.5} & 0.233 \textit{(0.132; 0.0010)} & 0.233 \textit{(0.033; 0.0056)} & 0.367 \textit{(0.343; 0.0398)} \\
        & \texttt{Gemini 3 Image} & 0.233 \textit{(0.433; 0.0010)} & 0.267 \textit{(0.141; 0.0056)} & 0.133 \textit{(0.330; 0.0398)} \\
        & \texttt{Flux 2 Pro} & 0.200 \textit{(-0.054; 0.0010)} & 0.100 \textit{(-0.134; 0.0056)} & 0.267 \textit{(0.399; 0.0398)} \\
    \hline
    \multirow{4}{*}{\textbf{Ad Video}}
        & \texttt{Grok Imagine} & 0.111 \textit{(0.375; 0.0010)} & 0.194 \textit{(0.502; 0.5446)} & 0.361 \textit{(-0.042; 0.1375)} \\
        & \texttt{Veo 3.1} & 0.417 \textit{(0.195; 0.0010)} & 0.333 \textit{(0.130; 0.5446)} & 0.083 \textit{(0.176; 0.1375)} \\
        & \texttt{Kling 3.0} & 0.278 \textit{(0.273; 0.0010)} & 0.333 \textit{(0.068; 0.5446)} & 0.222 \textit{(0.417; 0.1375)} \\
        & \texttt{Seedance 1.5} & 0.194 \textit{(0.310; 0.0010)} & 0.139 \textit{(0.226; 0.5446)} & 0.333 \textit{(0.401; 0.1375)} \\
    \hline
    \end{tabularx}
    \caption{Pairwise preference outcomes across workflow phases. Each cell reports a model's pairwise win rate for the given domain and phase, with Krippendorff's $\alpha$ and Friedman $p$-value in parentheses.}
    \label{tab:domain-model-summary}
\end{table*}

\subsection{Hypothesis \#1: Evaluation axes separate standards from taste}
We first posited that evaluator convergence would depend on the type of criterion being assessed. Creative artifacts combine relatively verifiable standards, such as prompt fidelity, readability, and functional layout, with preference-driven judgments about style, tone, and direction. We therefore expected more agreement on \textbf{Prompt Adherence}, more variation on \textbf{Visual Appeal}, and an intermediate pattern on \textbf{Usability}.

The scalar results show that these dimensions are related, but not interchangeable (Table~\ref{table:appendix-scalar-stats}). For example, for the refinement stage of Ad Images, Seedream 4.5 and Gemini 3 Image are rated highly on Prompt Adherence (4.00 and 4.03), but diverge in Visual Appeal (3.90 and 2.97).
% In Product Video refinement, Seedance 1.5 has the highest Prompt Adherence score (3.44) but Grok Imagine  on both Usability (3.39) and Visual Appeal (3.72). 
The pattern is not a clean partition of objective and subjective criteria, but it does indicate a consistent trend: the axes capture different kinds of expert judgment rather than three versions of the same quality score.

Where criteria were objectively verifiable, such as illegible text or broken visual hierarchy, inter-rater agreement was high. For outputs above this threshold of technical competence, rankings increasingly reflected individual aesthetic preference, consistent with the distinction between convergent and divergent dimensions illustrated in Figure~\ref{fig:example-output}.
Agreement also differed by rating dimension. Agreement on Prompt Adherence was consistently higher than agreement on Visual Appeal, where criteria are personal rather than shared.

Free-text responses contextualize this separation. Evaluators cited failures such as unreadable text, prompt mismatch, or broken visual hierarchy, but they also described taste as important once outputs crossed a threshold of technical competence. One Desktop App Mockup evaluator summarized this directly: \textit{``I was judging based off of personal opinion and taste of what looks the best in my eyes,''}. In a similar vein, a brand designer stated, in reference to a set of outputs with similar usability, \textit{``[h]onestly, I feel like all four images could be used as brand visuals... What made me choose some over others was the sense of life, some felt more dynamic, realistic, and human.''} Together, these results support the first hypothesis, showing that while some axes produce creative agreement, others surface preference variation. Additional details can be found in the appendix, including the full scalar table (Table~\ref{table:appendix-scalar-stats}) and scalar plots (Figures~\ref{fig:ad-image-scalar-dist}, \ref{fig:ad-image-scalar-traj}, and \ref{fig:video-scalar-ribbons}).

\subsection{Hypothesis \#2: Creative requirements shift across stages and domains}
Our second hypothesis posited that what counts as a ``good'' output would vary across stages of the creative workflow and across domains. The HCB results support this prediction, but not as a uniform increase or decrease in agreement. Instead, evaluator agreement shifted with the domain, the phase of the workflow, and the dimension being assessed.

This is visible in the trajectory of Kendall's $W$ across phases. In the Ad Images domain, agreement increased steadily from Ideation ($W = 0.345$) to Mockup ($W = 0.436$) to Refinement ($W = 0.549$), as feedback narrowed toward more verifiable criteria such as typography, contrast, and production constraints. Landing Pages followed a different trajectory: agreement dropped from $W = 0.484$ in Ideation to $W = 0.293$ in Mockup, recovering only slightly in Refinement ($W = 0.333$). In Ideation, pairwise preferences concentrated strongly around a single output. By Refinement, however, all models produced functionally acceptable pages, and evaluators increasingly justified their rankings through personal preference rather than shared failure modes.

Creatives prioritized outputs that established a usable direction in the early stage of ideation: qualitative feedback was distributed across many themes, with structure and layout were raised most often. Landing Page evaluators, for example, prioritized visual hierarchy and layout coherence, while Desktop App comments centered on usability and hierarchy. In Mockup, the task became more constrained. Color \& Theme was the most frequent theme, with Prompt Adherence--Usability increased to $r = 0.65$. 
Landing Page evaluators most prominently discussed prompt adherence, grid structure, color consistency, and typographic pairing, versus Desktop App comments that emphasized text visibility, CTA clarity, and component tweaks.
By the Refinement stage, feedback concentrated on final-output criteria, but the effect differed by domain. Ad Images provide the most direct example: mentions of typography rose from $3\%$ in Ideation to approximately $34\%$ in Refinement, with rater agreement being the highest in Refinement ($W = 0.549$). In Product Videos, however, numerous refinement prompts asked for targeted edits but resulted in  outputs that introduced new elements instead. Free-text responses suggest a loose overall hierarchy in which usability acts as a threshold, prompt adherence orders viable outputs, and visual appeal resolves close contests

% The qualitative rationales clarify what changed across stages. In Ideation, creatives prioritized outputs that established a usable direction; feedback was distributed across many themes, with structure and layout mentioned most often. Landing Page evaluators emphasized visual hierarchy and layout coherence, while Desktop App evaluators focused on usability and hierarchy. In Mockup, the task became more constrained: Color & Theme became the most frequent theme, and the association between Prompt Adherence and Usability increased to $r = 0.65$. By Refinement, feedback concentrated on final-output criteria, though the dominant concerns still differed by domain. Ad Images provide the most direct example: mentions of typography rose from $3\%$ in Ideation to approximately $34\%$ in Refinement, coinciding with the highest observed agreement in that domain. Product Videos, by contrast, exposed a different refinement failure: many prompts requested targeted edits, but outputs often introduced new elements instead. Overall, the rationales suggest a loose evaluative hierarchy in which usability acts as a threshold, prompt adherence orders otherwise viable outputs, and visual appeal resolves close comparisons.

\subsection{Hypothesis \#3: Assessing Creative Artifacts Requires More Than One Dimension}
Our third hypothesis proposed that model comparison would change when phase and evaluation axis were kept rather than collapsed into a single aggregate score. Table~\ref{table:appendix-domain-stats} shows this pattern. Perhaps the clearest evidence supporting our hypothesis that single-dimension evaluations are not sufficient for creative evaluation is that no individual model lead all three phases in any domain (Figure~\ref{fig:video-scalar-ribbons}; see Appendix for equivalent charts per domain).

\begin{figure*}[t]
    \centering
    \includegraphics[width=\textwidth]{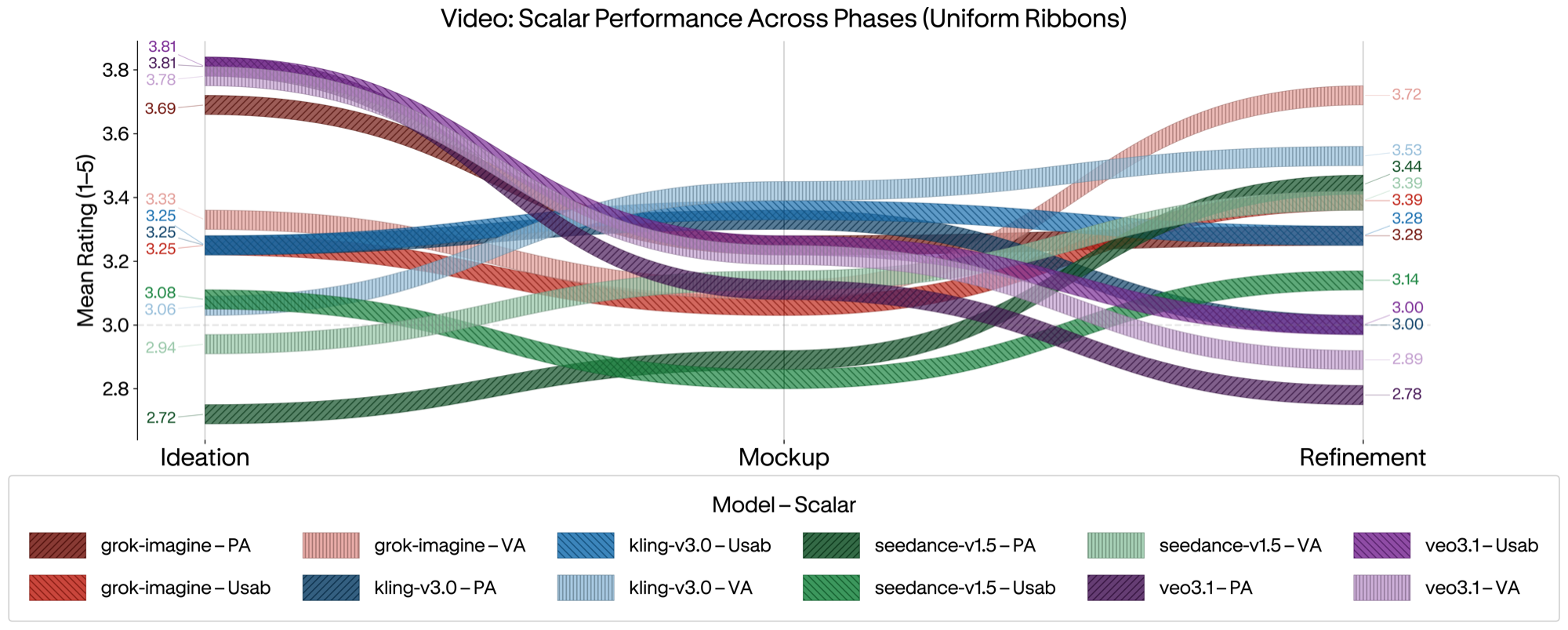}
    \caption{Scalar performance across the three phases for video generation, comprising one ribbon for each model--metric combination. Color encodes the model (Grok-Imagine in red, Kling-v3.0 in blue, Seedance-v1.5 in green, and Veo3.1 in purple), while shade encodes the evaluation metric, with the lightest ribbon denoting Visual Appeal (VA), the medium shade Usability (Usab), and the darkest shade Prompt Adherence (PA); these abbreviations (PA, Usab, VA) label the ribbons in the legend. Ribbon width is uniform. The overall range stays roughly constant across phases, spanning approximately 2.7 to 3.8 at both Ideation and Refinement, but the ordering of models changes considerably between them. Veo3.1 begins highest on Prompt Adherence at 3.84 but declines steadily to 2.78 at Refinement, the lowest score in the final phase. Conversely, Seedance-v1.5 on Prompt Adherence rises from 2.72 at Ideation to 3.44 at Refinement, and Grok-Imagine on Visual Appeal climbs to 3.72, the highest score overall, with Grok-Imagine on Usability close behind at 3.69. The dashed line denotes the neutral midpoint of 3.0.}
    \label{fig:video-scalar-ribbons}
\end{figure*}
% % The reason is that the expectations of a designer from a model change as their work progresses.

In Desktop Apps, Claude 4.6 has the highest win rate in Ideation and Mockup ($0.600$ and $0.400$), while GPT 5.3 is highest in Refinement ($0.400$) after the lowest Ideation value ($0.086$). In Ad Images, GPT Image 1.5 is highest in Ideation and Mockup ($0.333$ and $0.400$), while Seedream 4.5 is highest in Refinement ($0.367$). Product Videos have different leaders by phase: Veo 3.1 is highest in Ideation ($0.417$), Veo 3.1 and Kling 3.0 tie in Mockup ($0.333$), and Grok Imagine is highest in Refinement ($0.361$). 
% Because the reliability estimates are uneven, these should be read as descriptive shifts rather than stable rankings.

The extended domain results give the same pattern across phases. Landing Pages show phase-specific changes: Claude Opus 4.6 leads Ideation with an 80\% win rate, Gemini 3.1 leads in Mockup when a design system is introduced (68.9\%), and Claude returns to the lead in Refinement (60.0\%); by Refinement, all four models cluster between 3.9 and 4.4 across scalar dimensions. Product Videos show a different split: Veo 3.1 leads in Ideation (61.1\%) but receives negative refinement feedback for introducing new elements, while Grok Imagine leads in Refinement (56.5\%) and improves on fidelity-oriented themes. Kling 3.0 is the only video model above 50\% in all three phases (51\%, 61\%, and 52\%).

Ad Images show the same dependence on phase and axis. GPT Image 1.5 performs best earlier, while Seedream 4.5 improves by Refinement, especially on composition, usability, and typography; Flux 2 Pro also rises from Mockup to Refinement, while Gemini 3 Pro Image falls as typography and product accuracy become more important. These results are consistent with the third hypothesis in a limited sense. Model comparison changes depending on which phase and axis are being examined, but the data does not support a definitive taxonomy of models. Rather, collapsing the benchmark into a single overall ranking hides phase-specific differences where models appear stronger or weaker.

\section{Discussion \& Limitations}

As early pioneers~\cite{treffinger1972} in the field of creative studies wrote, \textit{``creativity may represent such a complex human phenomenon that we may never be able to represent it adequately as a single, unidimensional operational variable, or even as a small set of operations. ''} As we began to show in this work, evaluating creative quality cannot only be done through one axis of evaluation. Creative work is a hybrid of subjective and objective standards, and assessing only the objective loses important and useful signal in model evaluation and improving creatives' experiences. 
% the finding that no model led all three phases in any domain raises a practical question worth studying directly: does deliberate model switching improve outcomes, and can tools surface the right model at the right moment without adding friction? 

Current norms in AI evaluation do not account for the natural diversity and plurality present in creative labor, nor do they differentiate between the needs of creatives at different stages of their workflow.  We believe this can result in homogenized outputs that are less useful for creative work, leading to undifferentiated creative voices. We hope to build a shared vocabulary for creative workers engaging with these tools. 

To this end, model developers may use convergent and divergent criteria as distinct points of intervention: instances where experts converge highlight best practices that models can and should learn, while divergence identifies where a model should not optimize for one target but remain responsive to creative preference.

In building then evaluating this dataset, we find that the requirements of people \textit{evolve} throughout the stages of creative labor. Tool builders and creatives alike can explore phase-level needs shifts to indicate which models are better at a given moment, supporting deliberate model switching without adding friction to the creative workflow.

The question of what is quality remains unanswered in full, if only because the answer often lies in the eye of the beholder. As AI is increasingly integrated into creative processes, the future will need greater optionality for model selection and more investment in efforts preventing the loss of creative voices.

\paragraph{For Creatives}
This research provides language for something many creative professionals already feel: the frustration with AI tools is not necessarily that they produce bad work, but that they produce undifferentiated work that may be difficult to modulate. Understanding which models excel at exploration versus execution, and where in the process agreement breaks down into personal preference, gives creatives using AI tooling a basis for selecting models that fit their needs.

% We have the infrastructure to pursue this work: access to a large network of independent professional creatives, direct visibility into how professional work is generated, evaluated, and chosen, and an evaluation platform built to capture both signals at scale. The aim is to close the gap between how AI systems measure creative quality and how the people who make creative work judge it.

\paragraph{Limitations}
This study, although framed around the creative process, does not fully represent how creative work unfolds in practice. Creative work is rarely so linear. Designers iterate fluidly, move between tools, revisit stages, and often work across modalities within a single project. Future research will explore longer, less constrained creative arcs to better understand how these evaluation dynamics play out in practice.
% This study was conducted with 28 evaluators assessing 93 prompts across 80 sessions, yielding 5,940 pairwise judgments, 5,940 scalar ratings, and 3,675 qualitative responses. 
While the dataset we contribute provides a substantive basis for analysis, it represents only a starting point. For example, we do not control for differences in general model capability or the results of non-determinism, e.g., through iterative sampling, though temperature and various other parameters were standardized, and the prompts used were limited to a finite set of topics. We also collected responses from a relatively small participant group, limiting the statistical significance of our results. Future research should extend this work by expanding the evaluator pool, sampling prompts multiple times for robust coverage of model capabilities, and extending the prompt foci.

%  Phase-level performance shifts may partially reflect how varying constraint levels expose or compress baseline capability differences rather than creative workflow fit. 
% However, the consistency of evaluator agreement patterns across dimensions, where the same model produces high convergence on verifiable axes and high divergence on taste-driven axes, suggests that the structural separation between convergence and divergence holds independently of overall model capability.
% The prompts were authored by industry professionals and reviewed by our internal team for consistency and clarity, with a normalization process applied and negative-sentiment prompts removed. However, the prompt set was not subjected to external validation or independent expert review, and the possibility of latent bias in prompt construction cannot be excluded.

\clearpage
\bibliographystyle{plainnat}
\bibliography{example_paper}

\begin{thebibliography}{7}
\providecommand{\natexlab}[1]{#1}
\providecommand{\url}[1]{\texttt{#1}}
\expandafter\ifx\csname urlstyle\endcsname\relax
  \providecommand{\doi}[1]{doi: #1}\else
  \providecommand{\doi}{doi: \begingroup \urlstyle{rm}\Url}\fi

\bibitem[Basile et~al.(2021)Basile, Fell, Fornaciari, Hovy, Paun, Plank,
  Poesio, and Uma]{basile2021}
Valerio Basile, Michael Fell, Tommaso Fornaciari, Dirk Hovy, Silviu Paun,
  Barbara Plank, Massimo Poesio, and Anca Uma.
\newblock We need to consider disagreement in evaluation.
\newblock In \emph{Proceedings of the 1st Workshop on Evaluation and Comparison
  of {NLP} Systems}, pages 15--21, 2021.

\bibitem[Bommasani et~al.(2022)Bommasani, Creel, Kumar, Jurafsky, and
  Liang]{bommasani2022}
Rishi Bommasani, Kathleen~A. Creel, Arvind Kumar, Dan Jurafsky, and Percy
  Liang.
\newblock Picking on the same person: Does algorithmic monoculture lead to
  outcome homogenization?
\newblock In \emph{Advances in Neural Information Processing Systems},
  volume~35, pages 3663--3678, 2022.

\bibitem[Dawid and Skene(1979)]{dawid1979}
A.~Philip Dawid and Allan~M. Skene.
\newblock Maximum likelihood estimation of observer error-rates using the {EM}
  algorithm.
\newblock \emph{Journal of the Royal Statistical Society: Series C (Applied
  Statistics)}, 28\penalty0 (1):\penalty0 20--28, 1979.

\bibitem[Inel et~al.(2014)Inel, Khamkham, Cristea, Dumitrache, Rutjes, Ploeg,
  Romaszko, Aroyo, and Sips]{inel2014}
Oana Inel, Khalifeh Khamkham, Tatiana Cristea, Anca Dumitrache, Heine Rutjes,
  Jelle Ploeg, Lukasz Romaszko, Lora Aroyo, and Robert-Jan Sips.
\newblock {CrowdTruth}: Machine-human computation framework for harnessing
  disagreement in gathering annotated data.
\newblock In \emph{The Semantic Web -- {ISWC} 2014}, volume 8797 of
  \emph{Lecture Notes in Computer Science}, pages 486--504. Springer, 2014.

\bibitem[Lee(2022)]{lee2022}
Hye-Kyung Lee.
\newblock Rethinking creativity: Creative industries, {AI} and everyday
  creativity.
\newblock \emph{Media, Culture \& Society}, 44\penalty0 (3):\penalty0 601--612,
  2022.

\bibitem[Thanh-Tung and Tran(2020)]{thanhtung2020}
Hoang Thanh-Tung and Thanh Tran.
\newblock Catastrophic forgetting and mode collapse in {GANs}.
\newblock In \emph{2020 International Joint Conference on Neural Networks
  ({IJCNN})}, pages 1--10. IEEE, 2020.

\bibitem[Treffinger and Poggio(1972)]{treffinger1972}
Donald~J. Treffinger and John~P. Poggio.
\newblock Needed research on the measurement of creativity.
\newblock \emph{The Journal of Creative Behavior}, 6\penalty0 (4):\penalty0
  263--267, 1972.

\end{thebibliography}

\clearpage
\appendix
\onecolumn

\section{Dataset}
We release the full set of prompts, model outputs, and human evaluations underlying this benchmark as a public dataset. The data captures professional creative practitioners comparing the outputs of frontier generative models across a realistic, multi-stage creative workflow. Rather than collecting isolated one-shot judgments, the benchmark follows a three-stage pipeline, Ideation, Mockup, and Refinement, that mirrors how creative work progresses from initial concept to polished deliverable. Each model output is evaluated through three complementary signals: pairwise preference judgments, scalar ratings on three quality dimensions, and coded free-text rationales.
 
\subsection{Pipeline, Domains, and Models}
 
The benchmark spans five creative domains: Ad Images, Brand Design, Ad Video, Landing Pages, and Desktop App. Each domain is exercised across all three pipeline stages, yielding $95$ distinct prompts in total. Every prompt is generated by four candidate models drawn from a domain-appropriate pool, producing $380$ model outputs. In total the dataset covers $13$ frontier models, with four competing within each domain. The image domains (Ad Images, Brand Design) are served by image-generation models such as \texttt{gpt-image-1.5}, \texttt{gemini-3-pro-image-preview}, \texttt{seedream-4.5}, and the \texttt{flux-2} family; the Ad Video domain is served by video models including \texttt{veo3.1}, \texttt{kling-v3.0-pro}, \texttt{seedance-v1.5-pro}, and \texttt{grok-imagine-video}; and the code-based domains (Landing Pages, Desktop App) are served by general-purpose models including \texttt{claude-opus-4.6}, \texttt{gemini-3.1-pro-preview}, \texttt{gpt-5.3-codex}, and \texttt{qwen3.5-397b-a17b}. Prompts at the Mockup and Refinement stages may additionally supply an input image, reflecting the iterative, edit-driven nature of later pipeline stages.

\subsection{Data Files}
 
The dataset comprises five CSV files, summarized in Table~\ref{tab:dataset-files}. The files share \texttt{prompt\_id} as a common key and can be joined to reconstruct the full evaluation context for any output.

\begin{table}[H]
\centering
\small
\setlength{\tabcolsep}{5pt}
\renewcommand{\arraystretch}{1.3}
\begin{tabularx}{\textwidth}{|l|c|L|}
\hline
\cellcolor{gray!20}\textbf{File} & \cellcolor{gray!20}\textbf{Records} & \cellcolor{gray!20}\textbf{Description} \\
\hline
\texttt{prompts\_workflow.csv} & $95$ & Prompt specifications, including domain, pipeline stage, prompt text, and an optional input image for later stages. \\
\hline
\texttt{model\_outputs.csv} & $380$ & Model generations keyed by prompt and model. Visual outputs are stored as asset references; code-based outputs are stored inline. \\
\hline
\texttt{pairwise\_comparisons.csv} & $3{,}174$ & Head-to-head preference judgments recording the two competing models and the chosen model, along with the evaluator's core skill. \\
\hline
\texttt{scalar\_feedback.csv} & $2{,}116$ & Per-output ratings on a $1$ to $5$ scale across three dimensions: Prompt Adherence, Usability, and Visual Appeal. \\
\hline
\texttt{qualitative\_feedback.csv} & $2{,}247$ & Free-text rationales annotated with assigned themes, per-theme sentiment, and representative key quotes. \\
\hline
\end{tabularx}
\caption{Files in the released dataset. All files share \texttt{prompt\_id} as a join key.}
\label{tab:dataset-files}
\end{table}

% \subsection{Availability}

% We intend this release to support research on human preference modeling, evaluation of generative models in realistic creative workflows, and analysis of how professional judgment varies across domains, pipeline stages, and practitioner backgrounds.

% \section{Model Categories}
% \label{appendix:models}

% \begin{enumerate}
    % \item \textbf{Ideation specialists.} Claude Opus 4.6 and Veo 3.1 generate strong first drafts but struggle when asked to iterate, leading ideation and falling behind by refinement.
    % \item \textbf{Refinement climbers.} GPT 5.3 Codex, Grok Imagine Video, Seedream 4.5, and Qwen 3.5 start weak in ideation but improve as tasks become more constrained and specific, with GPT 5.3 Codex (Desktop Apps) and Grok Imagine Video each reaching first place in refinement despite starting last or third.
    % \item \textbf{Mockup specialists.} Gemini 3.1 Pro Preview and Gemini 3 Pro Image excel at introducing design system elements like color palette, grid, and typography, but struggle once iteration takes over.

% \end{enumerate}

\section{Evaluation Interface Extended}
\label{appendix:interface}
All evaluations were conducted within the Contra Labs evaluation environment, a web-based interface that walked raters through each phase of the creative process under controlled conditions. The interface presented each tournament as a single guided flow built around one prompt, moving raters sequentially through the pairwise comparison task and then the scaled-rating task.

\begin{figure}[H]
    \centering
    \begin{minipage}[t]{0.32\textwidth}
        \centering
        \textbf{(A)}\\[-0.25em]
        \includegraphics[width=\linewidth]{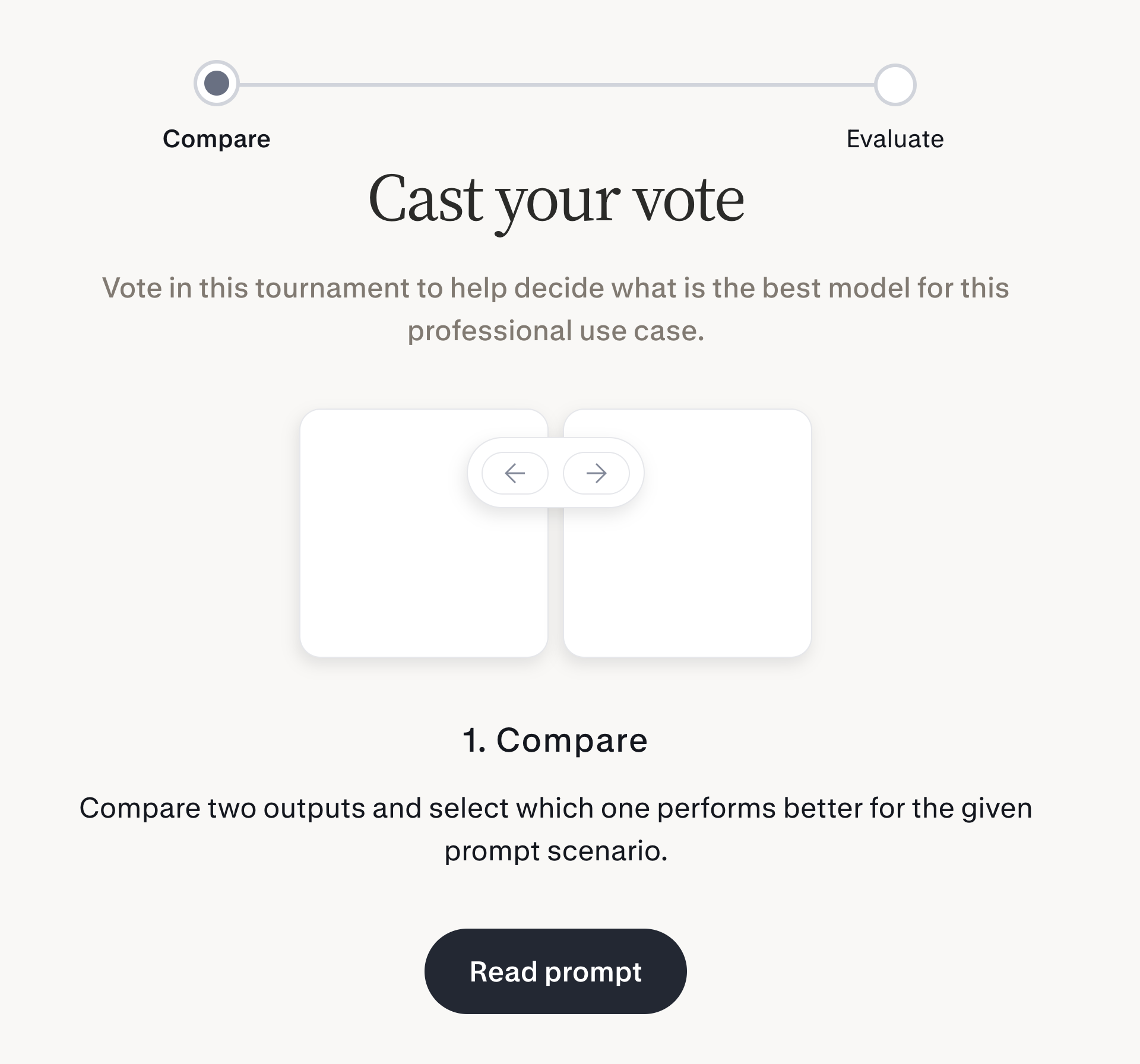}
    \end{minipage}\hfill
    \begin{minipage}[t]{0.32\textwidth}
        \centering
        \textbf{(B)}\\[-0.25em]
        \includegraphics[width=\linewidth]{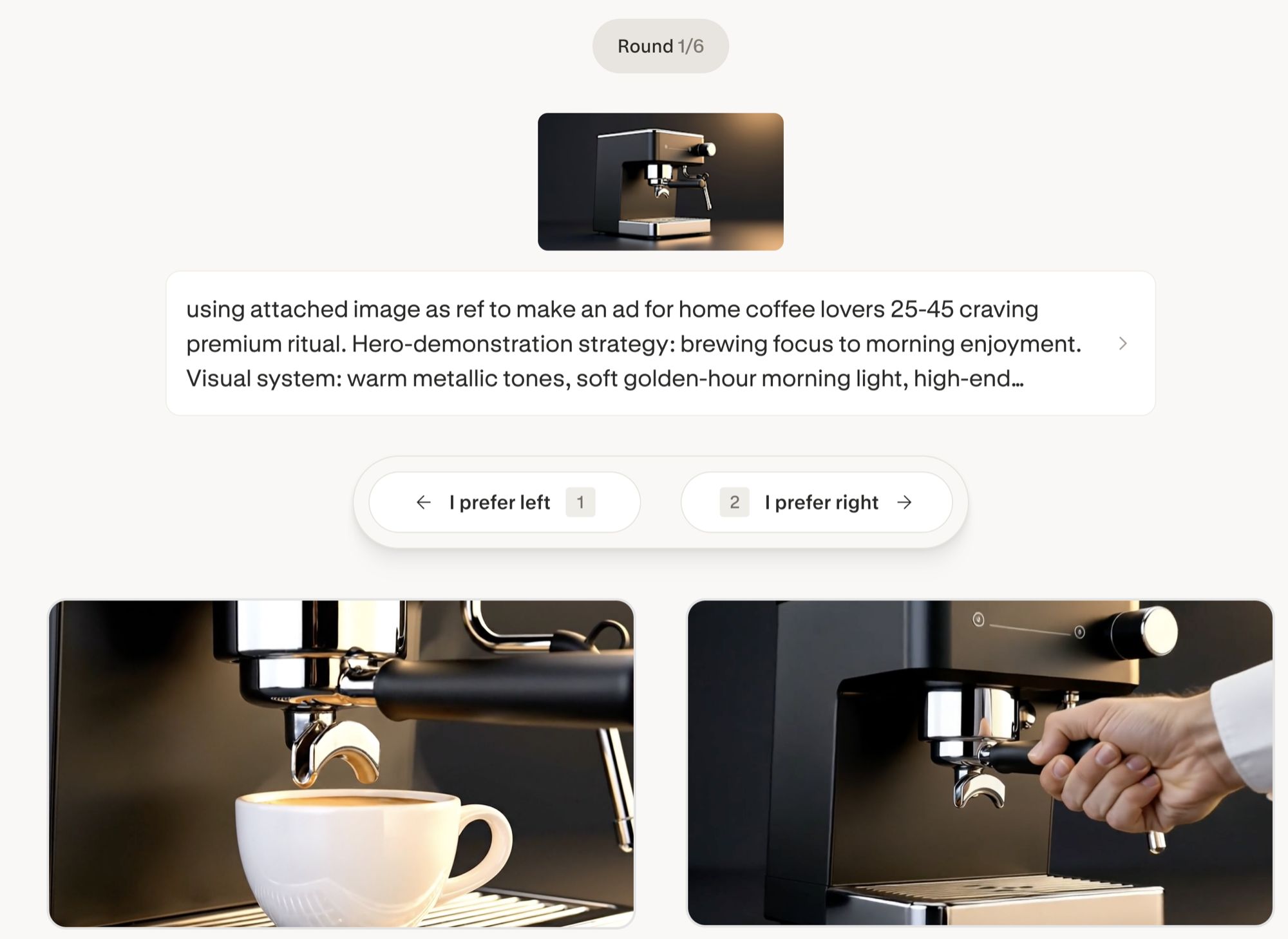}
    \end{minipage}\hfill
    \begin{minipage}[t]{0.32\textwidth}
        \centering
        \textbf{(C)}\\[-0.25em]
        \includegraphics[width=\linewidth]{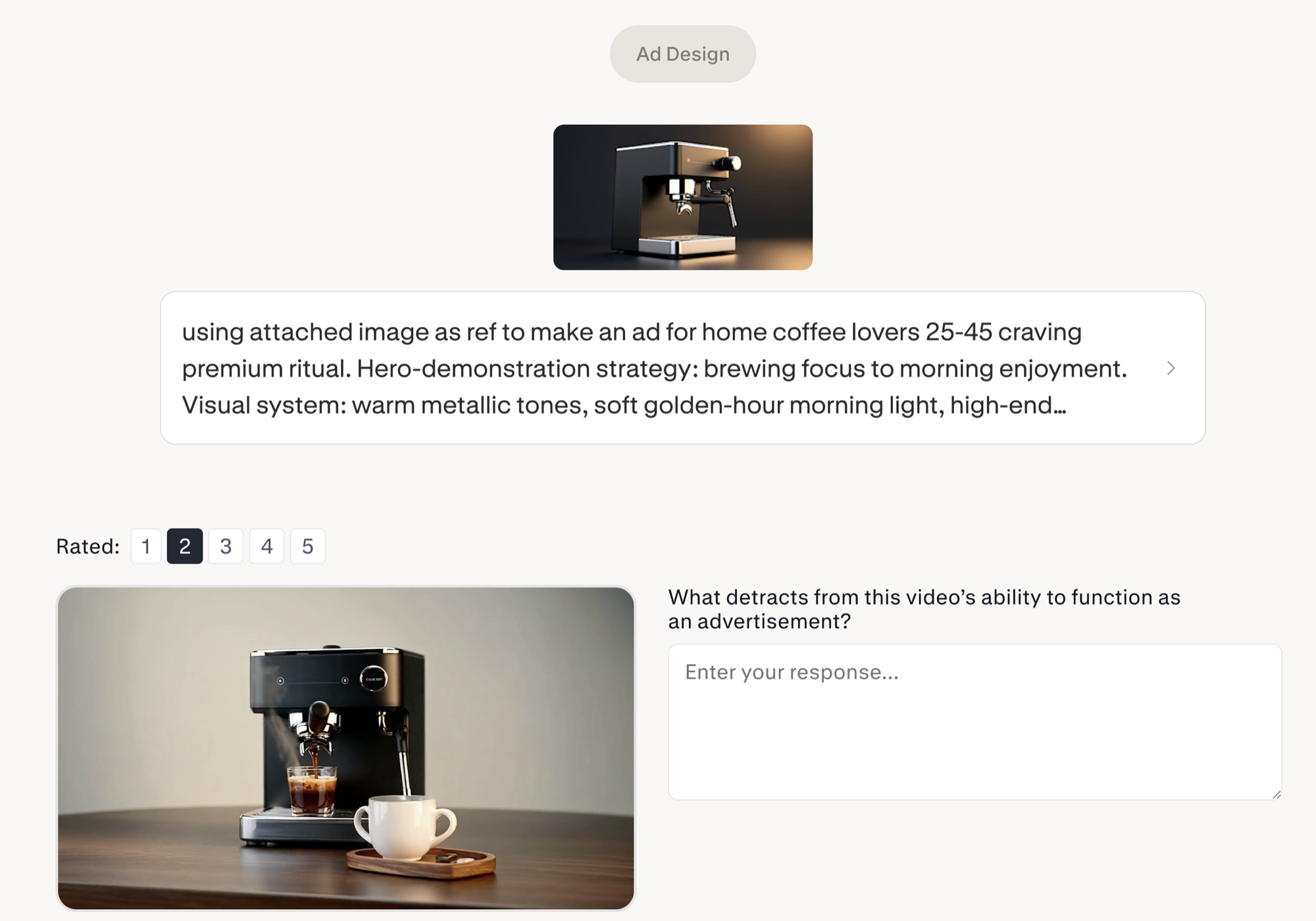}
    \end{minipage}
    \caption{Contra Labs evaluation interface, extended views. (A) Introductory screen for the pairwise comparison task, with access to instructions and the prompt. (B) Side-by-side pairwise comparison with the prompt and reference images visible. (C) Scalar rating task for prompt adherence, usability, and visual appeal.}
    \label{fig:interface-extended}
\end{figure}

Before any judgments were made, raters were shown a link to access the evaluation instructions, same as the one also provided up front in the participant instruction document, followed by an introductory screen identifying the first task as a pairwise comparison (Figure~\ref{fig:interface-extended}(A)). From this intro screen, a "Read prompt" control surfaced the prompt together with its reference images, after which the rater began the tournament. Throughout the pairwise task, the prompt and reference images remained visible while two outputs were presented at a time, and the rater selected the preferred output between the left and right options (Figure~\ref{fig:interface-extended}(B)). The interface advanced through successive pairings until all combinations for the prompt were complete. It then displayed a ranking of the outputs with model identities anonymized, and asked the rater, via a free-text field, what had influenced their decision.

After the pairwise task, the rater was brought to an introductory screen for the Likert rating task. Each of the three rating dimensions was presented on its own screen, showing the prompt and reference images, the dimension being assessed, a short sub-question framing that dimension, the output images, a 1-through-5 grading scale, and the corresponding grading rubric (Figure~\ref{fig:interface-extended}(C)). The first dimension, prompt adherence, was rated against this layout alone. For the subsequent dimensions, usability and visual appeal, the same layout was extended with a free-text field in which raters described the strengths and weaknesses of each image relative to the grading criteria. Completing the visual appeal ratings concluded the tournament.

Across both tasks, model identities were never shown: outputs appeared only under blinded codenames and their on-screen position was randomized to neutralize ordering and recognition effects. Keeping the prompt and reference images persistently in view ensured raters assessed each generation against the original creative intent rather than in isolation, while the free-text capture preserved each evaluator's own language for the subsequent thematic coding analysis.

\section{Survey Details}
\label{appendix:survey}

Each participant received \$350 as compensation for completing the questionnaire. The questionnaire was organized in four parts, exploring (1) respondents' creative process, whether they currently use AI tools, and, if not, reasons against adoption; (2) how they incorporated AI, including how much of their creative process is AI-supported, how much AI-generated content reaches final deliverables, and the perceived benefits of adoption (e.g., delivery speed, creative exploration, production cost, access to new capabilities); (3) eight modules on generative tool categories (image generation, video generation, prototyping, code generation, copywriting, audio/voice/music, 3D/motion graphics, and LLM-based research), detailing specific tools and where in their workflow they applied them; and (4) attitudes toward AI's role in the future of creative work, its effect on their earning potential, barriers to adoption, and comfort disclosing AI use to clients.

The majority of respondents reported 1-25\% of their creative process involved AI tools [add details] across a wide spread of applications or domains. When asked how they incorporated AI into their workflow, a video editor starting \textit{``loose [....] From there I refine and hone in on what I'm not getting, and adjust prompts to be more in line with my ideas. I think doing this leaves a lot of opportunity for surprise and serendipity.''} This reflects a general pattern across respondents: while AI use and adoption was not monolithic, those that did use these tools bucketed their use in a sequence of exploratory ideation, prototyping, followed by refining client deliverables. 

\section{Extended Analysis}
\label{sec:extended-analysis}

% No model leads all three phases in any domain (Figure~\ref{fig:brand-scalar-ribbons}; see Appendix for equivalent charts per domain).
% % The reason is that the expectations of a designer from a model change as their work progresses.

\paragraph{Landing Pages} Landing Pages shows the clearest phase-by-phase handoff. Claude Opus 4.6 leads Ideation with outputs rated highly on visual hierarchy and layout coherence. When a design system is introduced, Gemini 3.1 Pro Preview takes over (68.9\%), with evaluators citing its stronger execution on prompt adherence, grid structure, color consistency, and typographic pairing; this advantage reverses in Refinement, where incremental editing favors Claude Opus 4.6, which reclaims the lead (60.0\%). By Refinement, all four models cluster between 3.9--4.4 across all scalar dimensions and preference comes back down to taste, with GPT 5.3 Codex and Qwen 3.5 each showing steady improvement without leading any phase (Figure~\ref{fig:lp-scalar-ribbons-appendix}).

\paragraph{Product Videos} No model leads more than one phase in Product Videos, producing a three-phase handoff: Veo 3.1 leads Ideation (61.1\%), Kling 3.0 Pro leads Mockup (61.1\%), and Grok Imagine Video leads Refinement (56.5\%), with Kling 3.0 Pro the only model competitive across all three. Veo 3.1 is the only model that degrades on every measured dimension across all three phases: strong in Ideation when generating from scratch, it draws negative evaluator sentiment in Refinement for introducing new elements rather than applying targeted edits---a pattern reflected directly in realism sentiment, which moves from net +6 in Ideation to $-3$ in Refinement, while Grok Imagine Video improves from $-15$ to +20. Theme co-occurrence analysis reflects this split: Veo 3.1's evaluation profile clusters around generation themes like Motion \& Blur, while Grok Imagine's clusters around fidelity themes like Realism and Scene Coherence; Scene Coherence is net negative across all four models, suggesting temporal consistency remains the most persistent challenge in AI video generation. Prompt Adherence correlates independently with Usability (0.64) and Visual Appeal (0.58)---an output can look and feel right while still missing what the brief asked for (Figure~\ref{fig:video-win-rates}).

\paragraph{Ad Design} Ad Images has the most reliable convergence arc of any domain, with evaluator agreement rising at every phase transition (0.345 $\rightarrow$ 0.436 $\rightarrow$ 0.549) as criteria become progressively more verifiable. Analysis suggests evaluators follow a strict decision hierarchy: usability acts as a hard gate, prompt adherence is the primary ordering criterion among outputs that clear it, and visual appeal resolves close contests as a tiebreaker---criteria close to objective that evaluators reach without coordination. GPT Image 1.5 leads ideation and mockup but drops to third by refinement, with Seedream 4.5 following the opposite trajectory---climbing from third to first---while Flux 2 Pro makes a similar rise and Gemini 3 Pro, steady through the first two phases, finishes last. Seedream 4.5's ascent tracks sharply improved sentiment in composition, usability, and typography by refinement; Gemini 3 Pro, strong in mockup, collapses in refinement as typography and product accuracy turn negative (Figure~\ref{fig:ad-image-win-rates}).

\paragraph{Desktop Apps} Desktop App evaluation surfaces the broadest theme set of any domain, spanning prompt adherence, usability, layout quality, visual hierarchy, readability, and design conversion efficiency. Evaluator focus shifts consistently across phases: Ideation centers on usability and hierarchy; Mockup narrows to prompt adherence---key text visibility, CTA clarity, component refinement; Refinement surfaces structural consistency, with flaws in lower-ranked outputs becoming more apparent as evaluator attention shifts from layout to visual hierarchy. Theme co-occurrence analysis reveals distinct model signatures: Claude Opus 4.6 shows tight coupling between prompt adherence and perceived usability; Gemini 3.1 Pro's coupling is weaker, with occasional mismatches between adherence and usability scores; Qwen 3.5 is strong on high-level layout but clusters negatively around granular execution themes like typography and spacing. While models perform well at ideation and structural layout, Refinement shifts evaluator attention to interaction patterns, icon quality, and accessibility---areas where sentiment turns negative and models consistently underperform (Figure~\ref{fig:desktop-scalar-appendix}).

\subsection{Phase Insights}
Evaluator focus shifts in a consistent arc across all domains: Ideation centers on structure and layout, with moderate agreement reflecting the range of valid directions before a reference is established (Landing Page $W = 0.484$, Ad Image $W = 0.345$). In Mockup, Color \& Theme displaces Layout as the top concern and the Prompt Adherence--Usability correlation strengthens to $r = 0.65$---when a specification is explicit, following it closely produces a more usable output almost automatically. By Refinement, criteria narrow sharply: typography rises from $3\%$ of mentions to $\sim 34\%$ in Ad Images, evaluator agreement peaks ($W = 0.549$), and usability becomes the single strongest predictor of competitive success---outputs scoring 5 on usability finish in the top 2 ranks $84\%$ of the time versus $10\%$ for score-1 outputs. Early-stage feedback spreads across many dimensions; by Refinement it reduces to a few production-ready criteria, and the Usability--Visual Appeal correlation reaches $0.818$.

\section{Domain insights}

\begin{scriptsize}
\setlength{\tabcolsep}{4pt}
\arrayrulecolor{gray!40}
\begin{longtable}{|l|l|l|c|c|c|}
\caption{Model performance and evaluator agreement statistics across domains and phases.}
\label{table:appendix-domain-stats}\\
\hline
\cellcolor{gray!20}\textbf{Domain} & \cellcolor{gray!20}\textbf{Phase} & \cellcolor{gray!20}\textbf{Model} & \cellcolor{gray!20}\textbf{Win Rate} & \cellcolor{gray!20}\textbf{Krippendorff $\alpha$} & \cellcolor{gray!20}\textbf{Friedman $p$-value} \\
\hline
\endfirsthead
\caption[]{Model performance and evaluator agreement statistics across domains and phases (continued).}\\
\hline
\cellcolor{gray!20}\textbf{Domain} & \cellcolor{gray!20}\textbf{Phase} & \cellcolor{gray!20}\textbf{Model} & \cellcolor{gray!20}\textbf{Win Rate} & \cellcolor{gray!20}\textbf{Krippendorff $\alpha$} & \cellcolor{gray!20}\textbf{Friedman $p$-value} \\
\hline
\endhead
\hline
\multicolumn{6}{r}{Continued on next page}\\
\endfoot
\hline
\endlastfoot
Desktop Apps & Ideation & Claude 4.6 & 0.600 & 0.183 & 0.0010 \\
Desktop Apps & Ideation & GPT 5.3 & 0.086 & 0.568 & 0.0010 \\
Desktop Apps & Ideation & Gemini 3.1 & 0.257 & 0.148 & 0.0010 \\
Desktop Apps & Ideation & Qwen 3.5 & 0.057 & -0.066 & 0.0010 \\
\addlinespace
Desktop Apps & Mockup & Claude 4.6 & 0.400 & 0.101 & 0.3231 \\
Desktop Apps & Mockup & GPT 5.3 & 0.229 & 0.116 & 0.3231 \\
Desktop Apps & Mockup & Gemini 3.1 & 0.171 & -0.031 & 0.3231 \\
Desktop Apps & Mockup & Qwen 3.5 & 0.200 & 0.231 & 0.3231 \\
\addlinespace
Desktop Apps & Refinement & Claude 4.6 & 0.171 & -0.042 & 0.1134 \\
Desktop Apps & Refinement & GPT 5.3 & 0.400 & 0.119 & 0.1134 \\
Desktop Apps & Refinement & Gemini 3.1 & 0.200 & 0.243 & 0.1134 \\
Desktop Apps & Refinement & Qwen 3.5 & 0.229 & 0.234 & 0.1134 \\
\addlinespace
Landing Pages & Ideation & Claude 4.6 & 0.533 & -0.060 & 0.0000 \\
Landing Pages & Ideation & Gemini 3.1 & 0.367 & -0.058 & 0.0000 \\
Landing Pages & Ideation & GPT 5.3 & 0.100 & 0.075 & 0.0000 \\
Landing Pages & Ideation & Qwen 3.5 & 0.000 & -0.043 & 0.0000 \\
\addlinespace
Landing Pages & Mockup & Claude 4.6 & 0.300 & 0.072 & 0.3697 \\
Landing Pages & Mockup & Gemini 3.1 & 0.467 & 0.040 & 0.3697 \\
Landing Pages & Mockup & GPT 5.3 & 0.167 & -0.011 & 0.3697 \\
Landing Pages & Mockup & Qwen 3.5 & 0.067 & 0.165 & 0.3697 \\
\addlinespace
Landing Pages & Refinement & Claude 4.6 & 0.300 & -0.076 & 0.4678 \\
Landing Pages & Refinement & Gemini 3.1 & 0.267 & 0.039 & 0.4678 \\
Landing Pages & Refinement & GPT 5.3 & 0.233 & 0.209 & 0.4678 \\
Landing Pages & Refinement & Qwen 3.5 & 0.200 & 0.026 & 0.4678 \\
\addlinespace
Brand Assets & Ideation & Gemini 3 Image & 0.381 & 0.590 & 0.0010 \\
Brand Assets & Ideation & GPT Image 1.5 & 0.333 & 0.054 & 0.0010 \\
Brand Assets & Ideation & Seedream 4.5 & 0.119 & 0.151 & 0.0010 \\
Brand Assets & Ideation & Flux 2 Max & 0.167 & 0.135 & 0.0010 \\
\addlinespace
Brand Assets & Mockup & Gemini 3 Image & 0.500 & 0.041 & 0.0015 \\
Brand Assets & Mockup & GPT Image 1.5 & 0.208 & 0.280 & 0.0015 \\
Brand Assets & Mockup & Seedream 4.5 & 0.167 & 0.236 & 0.0015 \\
Brand Assets & Mockup & Flux 2 Max & 0.125 & 0.770 & 0.0015 \\
\addlinespace
Brand Assets & Refinement & Gemini 3 Image & 0.417 & -0.031 & 0.0033 \\
Brand Assets & Refinement & GPT Image 1.5 & 0.361 & 0.353 & 0.0033 \\
Brand Assets & Refinement & Seedream 4.5 & 0.139 & 0.049 & 0.0033 \\
Brand Assets & Refinement & Flux 2 Max & 0.083 & 0.343 & 0.0033 \\
\addlinespace
Ad Images & Ideation & GPT Image 1.5 & 0.333 & 0.476 & 0.0010 \\
Ad Images & Ideation & Seedream 4.5 & 0.233 & 0.132 & 0.0010 \\
Ad Images & Ideation & Gemini 3 Image & 0.233 & 0.433 & 0.0010 \\
Ad Images & Ideation & Flux 2 Pro & 0.200 & -0.054 & 0.0010 \\
\addlinespace
Ad Images & Mockup & GPT Image 1.5 & 0.400 & 0.183 & 0.0056 \\
Ad Images & Mockup & Seedream 4.5 & 0.233 & 0.033 & 0.0056 \\
Ad Images & Mockup & Gemini 3 Image & 0.267 & 0.141 & 0.0056 \\
Ad Images & Mockup & Flux 2 Pro & 0.100 & -0.134 & 0.0056 \\
\addlinespace
Ad Images & Refinement & GPT Image 1.5 & 0.233 & 0.112 & 0.0398 \\
Ad Images & Refinement & Seedream 4.5 & 0.367 & 0.343 & 0.0398 \\
Ad Images & Refinement & Gemini 3 Image & 0.133 & 0.330 & 0.0398 \\
Ad Images & Refinement & Flux 2 Pro & 0.267 & 0.399 & 0.0398 \\
\addlinespace
Ad Video & Ideation & Grok Imagine & 0.111 & 0.375 & 0.0010 \\
Ad Video & Ideation & Veo 3.1 & 0.417 & 0.195 & 0.0010 \\
Ad Video & Ideation & Kling 3.0 & 0.278 & 0.273 & 0.0010 \\
Ad Video & Ideation & Seedance 1.5 & 0.194 & 0.310 & 0.0010 \\
\addlinespace
Ad Video & Mockup & Grok Imagine & 0.194 & 0.502 & 0.5446 \\
Ad Video & Mockup & Veo 3.1 & 0.333 & 0.130 & 0.5446 \\
Ad Video & Mockup & Kling 3.0 & 0.333 & 0.068 & 0.5446 \\
Ad Video & Mockup & Seedance 1.5 & 0.139 & 0.226 & 0.5446 \\
\addlinespace
Ad Video & Refinement & Grok Imagine & 0.361 & -0.042 & 0.1375 \\
Ad Video & Refinement & Veo 3.1 & 0.083 & 0.176 & 0.1375 \\
Ad Video & Refinement & Kling 3.0 & 0.222 & 0.417 & 0.1375 \\
Ad Video & Refinement & Seedance 1.5 & 0.333 & 0.401 & 0.1375 \\
\end{longtable}
\arrayrulecolor{black}
\end{scriptsize}

\begin{scriptsize}
\setlength{\tabcolsep}{4pt}
\arrayrulecolor{gray!40}
\begin{longtable}{|l|l|l|c|c|c|}
\caption{Mean scalar ratings (1--5) for Prompt Adherence, Usability, and Visual Appeal across domains and phases.}
\label{table:appendix-scalar-stats}\\
\hline
\cellcolor{gray!20}\textbf{Domain} & \cellcolor{gray!20}\textbf{Phase} & \cellcolor{gray!20}\textbf{Model} & \cellcolor{gray!20}\textbf{Prompt Adherence} & \cellcolor{gray!20}\textbf{Usability} & \cellcolor{gray!20}\textbf{Visual Appeal} \\
\hline
\endfirsthead
\caption[]{Mean scalar ratings (1--5) for Prompt Adherence, Usability, and Visual Appeal across domains and phases (continued).}\\
\hline
\cellcolor{gray!20}\textbf{Domain} & \cellcolor{gray!20}\textbf{Phase} & \cellcolor{gray!20}\textbf{Model} & \cellcolor{gray!20}\textbf{Prompt Adherence} & \cellcolor{gray!20}\textbf{Usability} & \cellcolor{gray!20}\textbf{Visual Appeal} \\
\hline
\endhead
\hline
\multicolumn{6}{r}{Continued on next page}\\
\endfoot
\hline
\endlastfoot
Desktop Apps & Ideation & Claude 4.6 & 3.86 & 3.66 & 3.40 \\
Desktop Apps & Ideation & GPT 5.3 & 3.49 & 2.97 & 3.09 \\
Desktop Apps & Ideation & Gemini 3.1 & 3.46 & 3.26 & 3.11 \\
Desktop Apps & Ideation & Qwen 3.5 & 3.26 & 2.94 & 2.71 \\
\addlinespace
Desktop Apps & Mockup & Claude 4.6 & 3.97 & 3.69 & 3.83 \\
Desktop Apps & Mockup & GPT 5.3 & 3.94 & 3.71 & 3.69 \\
Desktop Apps & Mockup & Gemini 3.1 & 3.89 & 3.69 & 3.91 \\
Desktop Apps & Mockup & Qwen 3.5 & 3.63 & 3.49 & 3.63 \\
\addlinespace
Desktop Apps & Refinement & Claude 4.6 & 3.69 & 3.77 & 3.71 \\
Desktop Apps & Refinement & GPT 5.3 & 4.03 & 3.86 & 3.89 \\
Desktop Apps & Refinement & Gemini 3.1 & 3.74 & 3.66 & 3.71 \\
Desktop Apps & Refinement & Qwen 3.5 & 3.74 & 3.69 & 3.60 \\
\addlinespace
Landing Pages & Ideation & Claude 4.6 & 3.77 & 4.23 & 3.97 \\
Landing Pages & Ideation & Gemini 3.1 & 3.83 & 3.87 & 3.67 \\
Landing Pages & Ideation & GPT 5.3 & 3.10 & 3.20 & 3.03 \\
Landing Pages & Ideation & Qwen 3.5 & 3.07 & 3.43 & 3.13 \\
\addlinespace
Landing Pages & Mockup & Claude 4.6 & 3.60 & 3.90 & 3.63 \\
Landing Pages & Mockup & Gemini 3.1 & 3.97 & 4.03 & 3.73 \\
Landing Pages & Mockup & GPT 5.3 & 4.00 & 3.73 & 3.57 \\
Landing Pages & Mockup & Qwen 3.5 & 3.87 & 4.00 & 3.60 \\
\addlinespace
Landing Pages & Refinement & Claude 4.6 & 4.43 & 4.13 & 4.27 \\
Landing Pages & Refinement & Gemini 3.1 & 4.13 & 3.93 & 4.13 \\
Landing Pages & Refinement & GPT 5.3 & 4.17 & 4.07 & 4.23 \\
Landing Pages & Refinement & Qwen 3.5 & 4.07 & 4.03 & 3.97 \\
\addlinespace
Brand Assets & Ideation & Gemini 3 Image & 3.62 & 3.43 & 3.52 \\
Brand Assets & Ideation & GPT Image 1.5 & 4.26 & 4.07 & 4.14 \\
Brand Assets & Ideation & Seedream 4.5 & 3.55 & 3.24 & 3.17 \\
Brand Assets & Ideation & Flux 2 Max & 3.81 & 3.57 & 3.83 \\
\addlinespace
Brand Assets & Mockup & Gemini 3 Image & 3.79 & 4.00 & 4.17 \\
Brand Assets & Mockup & GPT Image 1.5 & 3.54 & 3.17 & 3.38 \\
Brand Assets & Mockup & Seedream 4.5 & 2.92 & 2.88 & 2.92 \\
Brand Assets & Mockup & Flux 2 Max & 2.62 & 2.58 & 2.96 \\
\addlinespace
Brand Assets & Refinement & Gemini 3 Image & 4.08 & 4.06 & 3.61 \\
Brand Assets & Refinement & GPT Image 1.5 & 3.72 & 3.64 & 3.39 \\
Brand Assets & Refinement & Seedream 4.5 & 3.22 & 3.33 & 3.28 \\
Brand Assets & Refinement & Flux 2 Max & 2.92 & 2.81 & 2.61 \\
\addlinespace
Ad Images & Ideation & GPT Image 1.5 & 3.50 & 3.03 & 2.93 \\
Ad Images & Ideation & Seedream 4.5 & 2.93 & 2.93 & 3.27 \\
Ad Images & Ideation & Gemini 3 Image & 3.13 & 2.93 & 3.20 \\
Ad Images & Ideation & Flux 2 Pro & 3.27 & 3.03 & 3.23 \\
\addlinespace
Ad Images & Mockup & GPT Image 1.5 & 3.97 & 3.93 & 3.87 \\
Ad Images & Mockup & Seedream 4.5 & 3.43 & 3.17 & 3.17 \\
Ad Images & Mockup & Gemini 3 Image & 3.57 & 3.60 & 3.77 \\
Ad Images & Mockup & Flux 2 Pro & 3.33 & 3.33 & 3.03 \\
\addlinespace
Ad Images & Refinement & GPT Image 1.5 & 4.13 & 3.73 & 3.23 \\
Ad Images & Refinement & Seedream 4.5 & 4.00 & 4.23 & 3.90 \\
Ad Images & Refinement & Gemini 3 Image & 4.03 & 3.53 & 2.97 \\
Ad Images & Refinement & Flux 2 Pro & 3.63 & 3.67 & 3.00 \\
\addlinespace
Ad Video & Ideation & Grok Imagine & 3.69 & 3.25 & 3.33 \\
Ad Video & Ideation & Veo 3.1 & 3.81 & 3.81 & 3.78 \\
Ad Video & Ideation & Kling 3.0 & 3.25 & 3.25 & 3.06 \\
Ad Video & Ideation & Seedance 1.5 & 2.72 & 3.08 & 2.94 \\
\addlinespace
Ad Video & Mockup & Grok Imagine & 3.25 & 3.06 & 3.11 \\
Ad Video & Mockup & Veo 3.1 & 3.11 & 3.25 & 3.22 \\
Ad Video & Mockup & Kling 3.0 & 3.33 & 3.36 & 3.42 \\
Ad Video & Mockup & Seedance 1.5 & 2.89 & 2.83 & 3.14 \\
\addlinespace
Ad Video & Refinement & Grok Imagine & 3.28 & 3.39 & 3.72 \\
Ad Video & Refinement & Veo 3.1 & 2.78 & 3.00 & 2.89 \\
Ad Video & Refinement & Kling 3.0 & 3.00 & 3.28 & 3.53 \\
Ad Video & Refinement & Seedance 1.5 & 3.44 & 3.14 & 3.39 \\
\end{longtable}
\arrayrulecolor{black}
\end{scriptsize}

\begin{figure*}[t]
    \centering
    \includegraphics[width=1.1\textwidth]{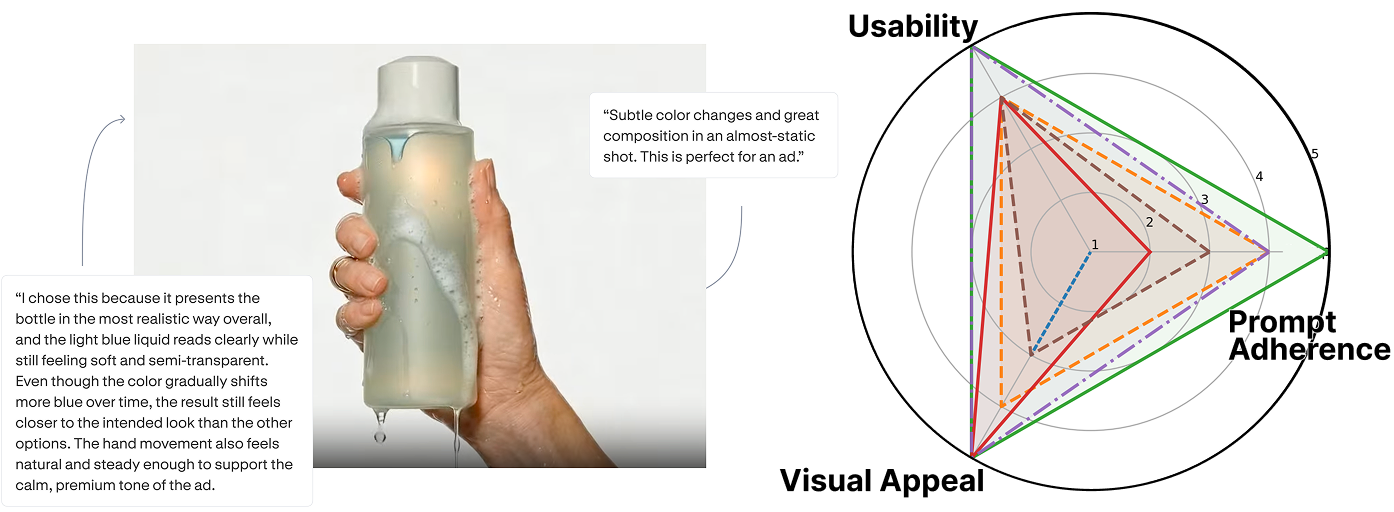}
    \caption{(A) shows an example output for brand images, along with a sampled quote included as explanation of domain expert quotes. (B) illustrates a summary ranking of the image across 6 experts--trends are relatively monotonic, illustrating directional alignment (or ``convergence'') in opinion.}
    \label{fig:brand-example-overview}
\end{figure*}

Figure~\ref{fig:brand-example-overview} provides a supplementary brand-design example, including a sampled expert quote and summary ranking across six evaluators.

\section{Survey Insights}
\label{sec:survey-insights}
\begin{figure}[H]
    \centering
    \includegraphics[width=.6\textwidth]{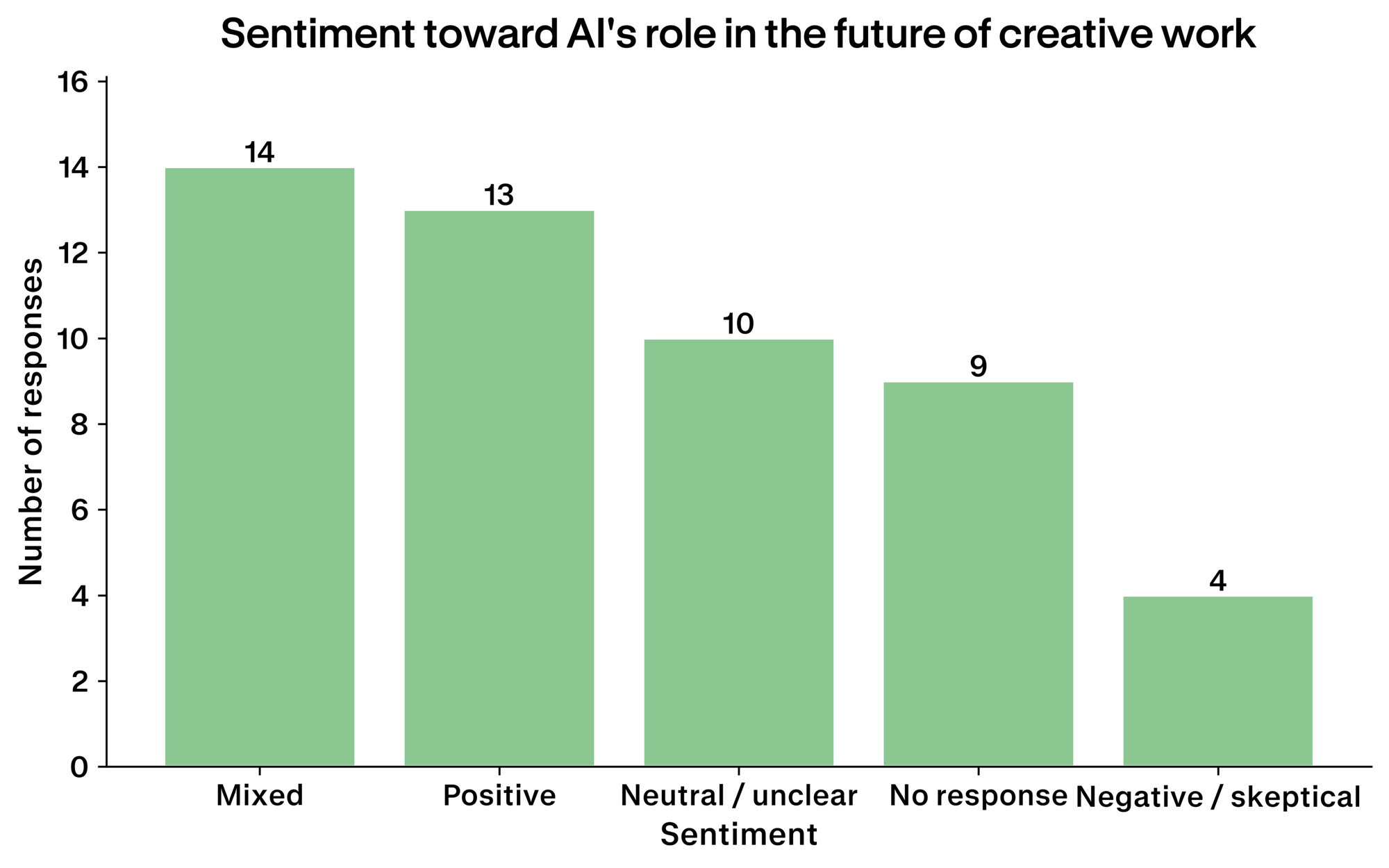}
    % \caption{Model win rates across the three-stage creative pipeline for Ad Image generation. Results illustrate phase-dependent performance fluctuations.}
    \caption{Distribution of respondent sentiment toward AI's role in the future of creative work (50 responses). Mixed sentiment is most common at 14 responses, closely followed by Positive (13). Neutral or unclear responses account for 10, and 9 respondents gave no response. Negative or skeptical sentiment is the least common, at 4 responses.}
    \label{fig:survey-ai-sentiment}
\end{figure}

\begin{figure}[htbp]
    \centering
    \includegraphics[width=.6\textwidth]{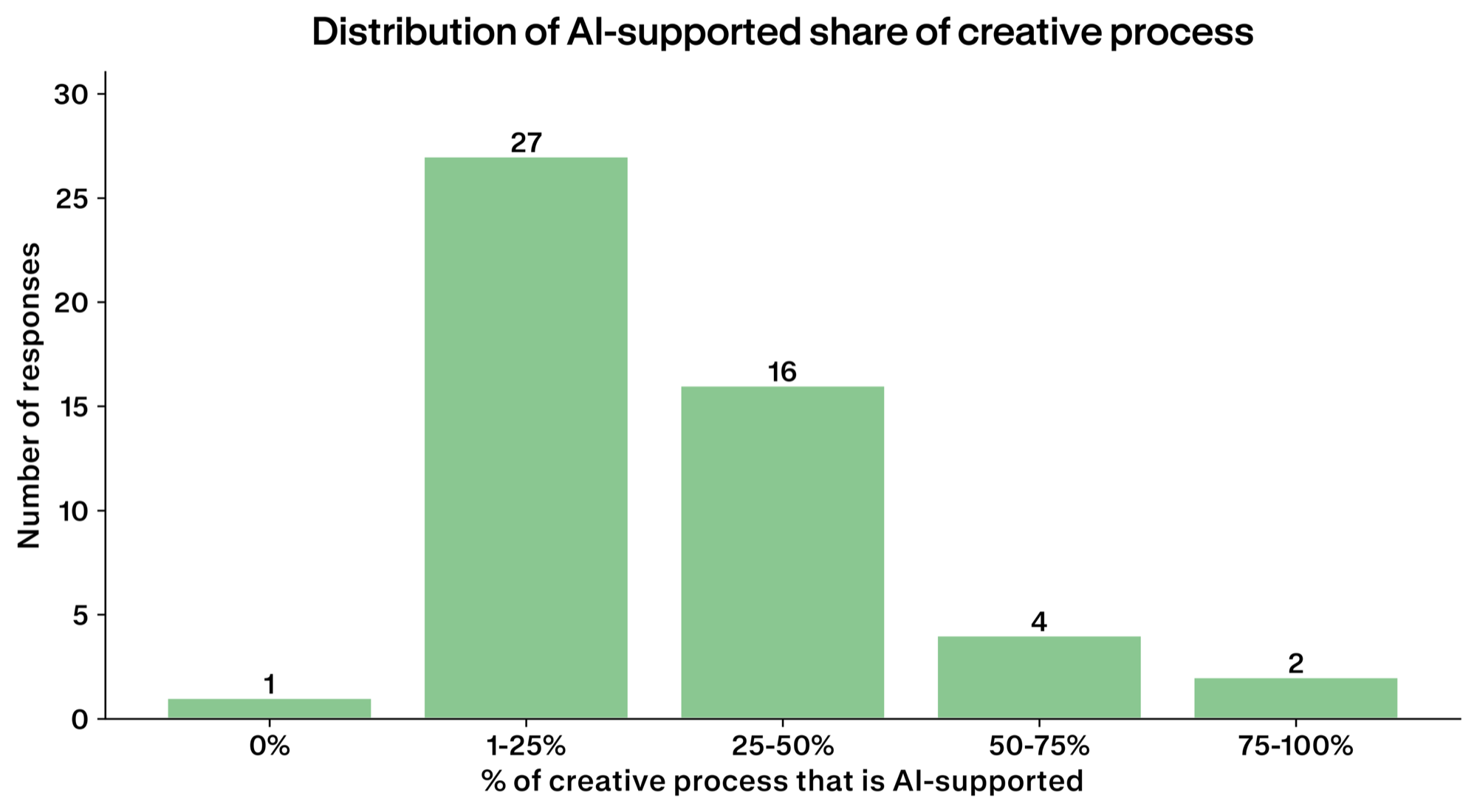}
    % \caption{Model win rates across the three-stage creative pipeline for Ad Image generation. Results illustrate phase-dependent performance fluctuations.}
    \caption{Distribution of the self-reported share of the creative process that is AI-supported (50 responses). The most common range is 1 to 25\%, reported by 27 respondents, followed by 25 to 50\% (16). Higher levels of reliance are uncommon, with 4 respondents reporting 50 to 75\% and 2 reporting 75 to 100\%, while a single respondent reported no AI support (0\%).}
    \label{fig:survey-ai-share}
\end{figure}

\begin{figure}[htbp]
    \centering
    \includegraphics[width=1\textwidth]{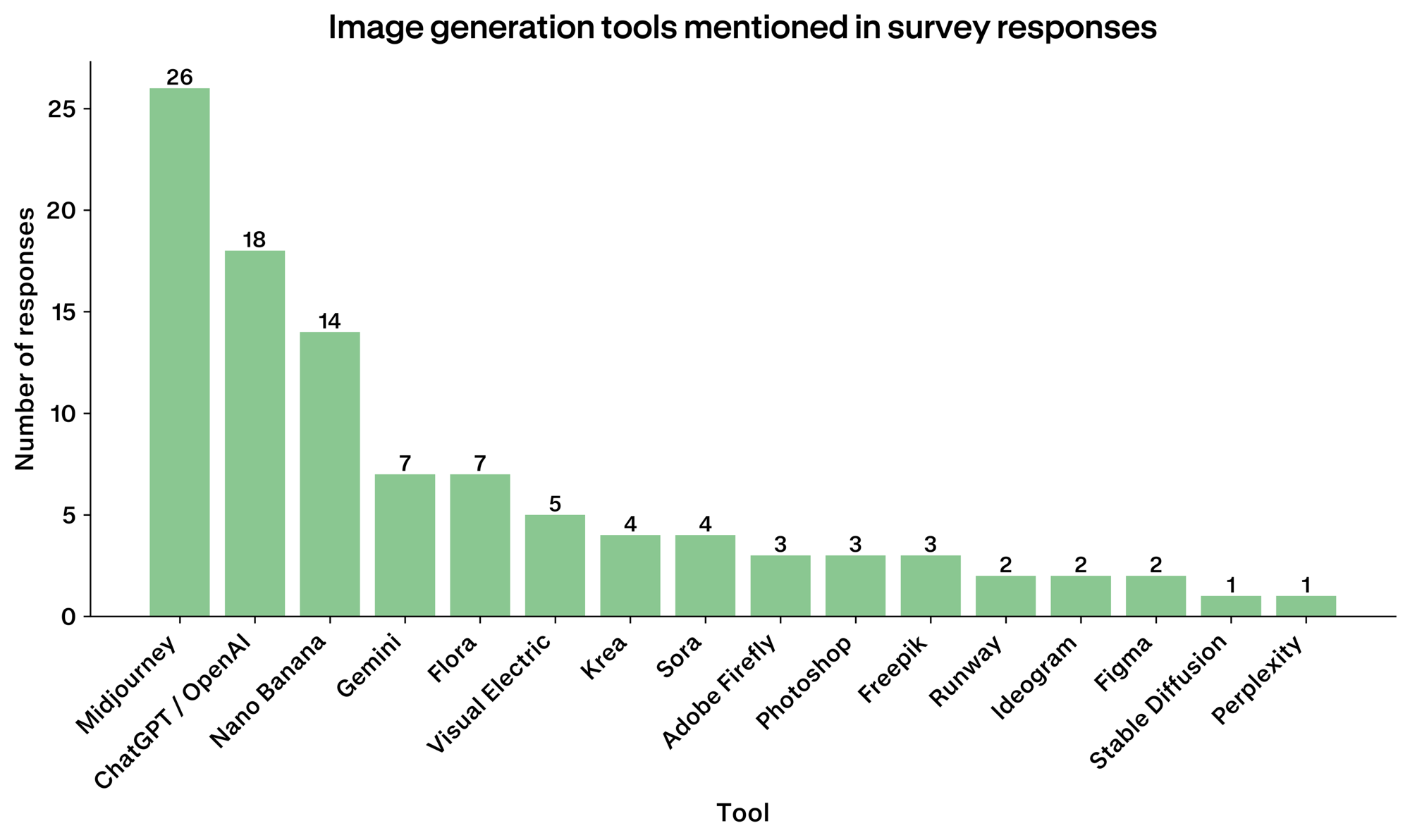}
    % \caption{Model win rates across the three-stage creative pipeline for Ad Image generation. Results illustrate phase-dependent performance fluctuations.}
    \caption{Image generation tools mentioned in survey responses. Midjourney is cited most frequently at 26 responses, followed by ChatGPT / OpenAI (18) and Nano Banana (14). A middle tier comprises Gemini and Flora (7 each), Visual Electric (5), and Krea and Sora (4 each). The remaining tools, including Adobe Firefly, Photoshop, Freepik, Runway, Ideogram, Figma, Stable Diffusion, and Perplexity, were each named three or fewer times.}
    \label{fig:survey-image-tools}
\end{figure}

\begin{figure}[htbp]
    \centering
    \includegraphics[width=1\textwidth]{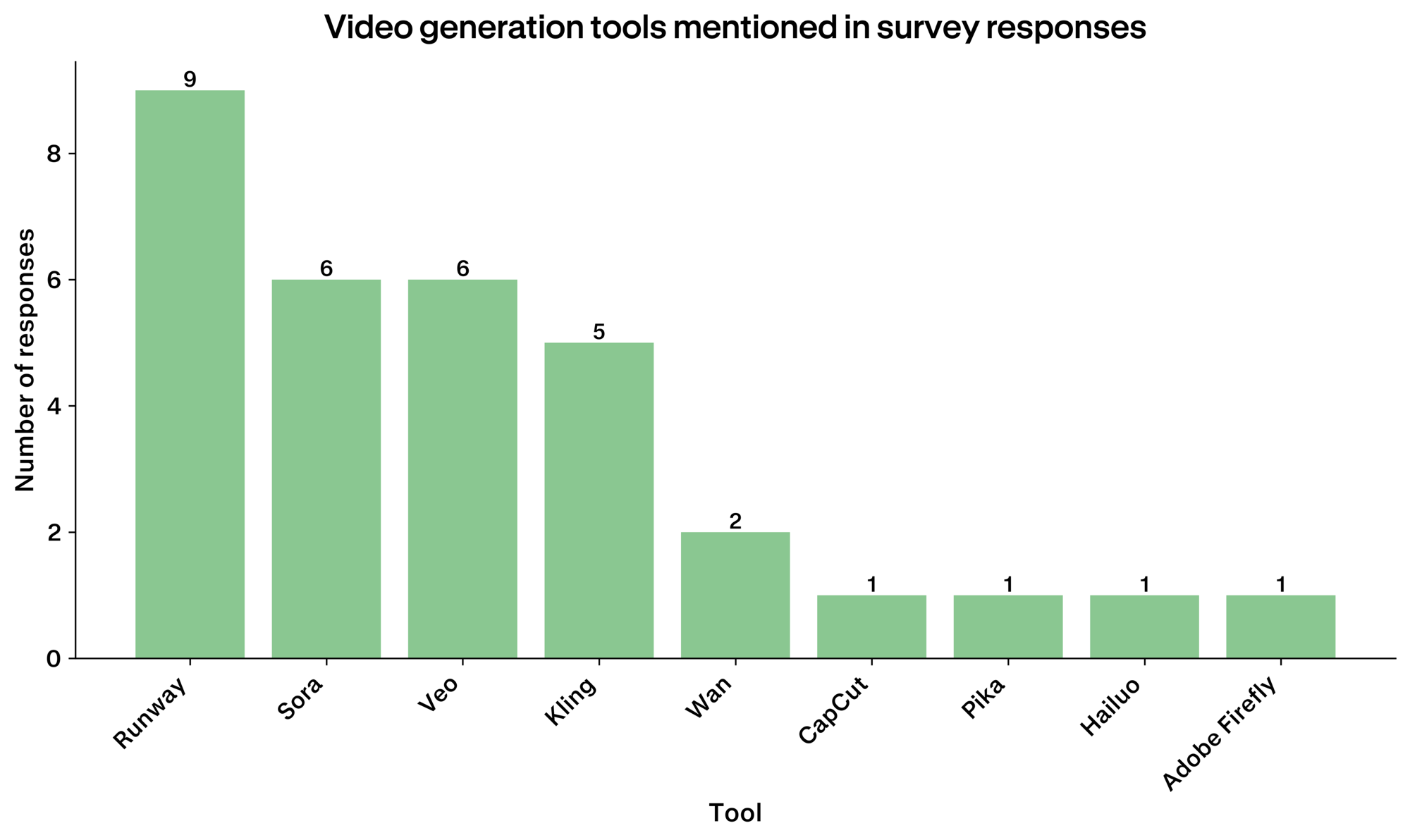}
    % \caption{Model win rates across the three-stage creative pipeline for Ad Image generation. Results illustrate phase-dependent performance fluctuations.}
    \caption{Video generation tools mentioned in survey responses. Runway is the most frequently cited at 9 responses, followed by Sora and Veo (6 each) and Kling (5). The remaining tools were named far less often: Wan (2), and CapCut, Pika, Hailuo, and Adobe Firefly (1 each).}
    \label{fig:survey-video-tools}
\end{figure}

\begin{figure}[htbp]
    \centering
    \includegraphics[width=1\textwidth]{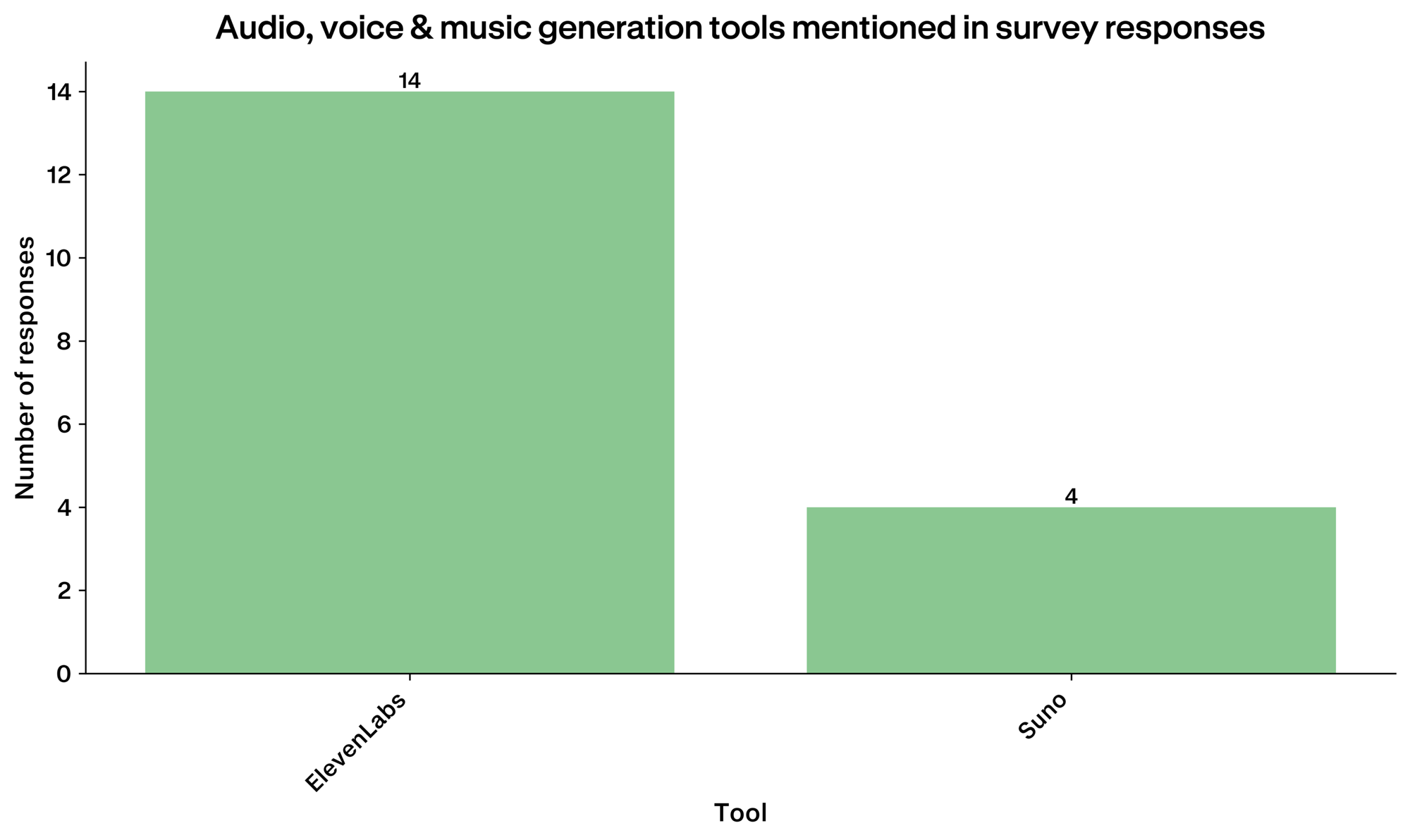}
    % \caption{Model win rates across the three-stage creative pipeline for Ad Image generation. Results illustrate phase-dependent performance fluctuations.}
    \caption{Audio, voice, and music generation tools mentioned in survey responses. Only two tools were named: ElevenLabs, cited in 14 responses, and Suno, cited in 4.}
    \label{fig:survey-audio-tools}
\end{figure}

Figure~\ref{fig:survey-ai-sentiment} summarizes respondents' attitudes toward AI in creative work, and Figure~\ref{fig:survey-ai-share} reports the self-reported share of each respondent's process that is AI-supported. Figures~\ref{fig:survey-image-tools}, \ref{fig:survey-video-tools}, and \ref{fig:survey-audio-tools} list the image, video, and audio tools named most frequently in survey responses.

\section{Domain Insights}
\subsection{Ad Images}
\begin{figure}[htbp]
    \centering
    \includegraphics[width=1\textwidth]{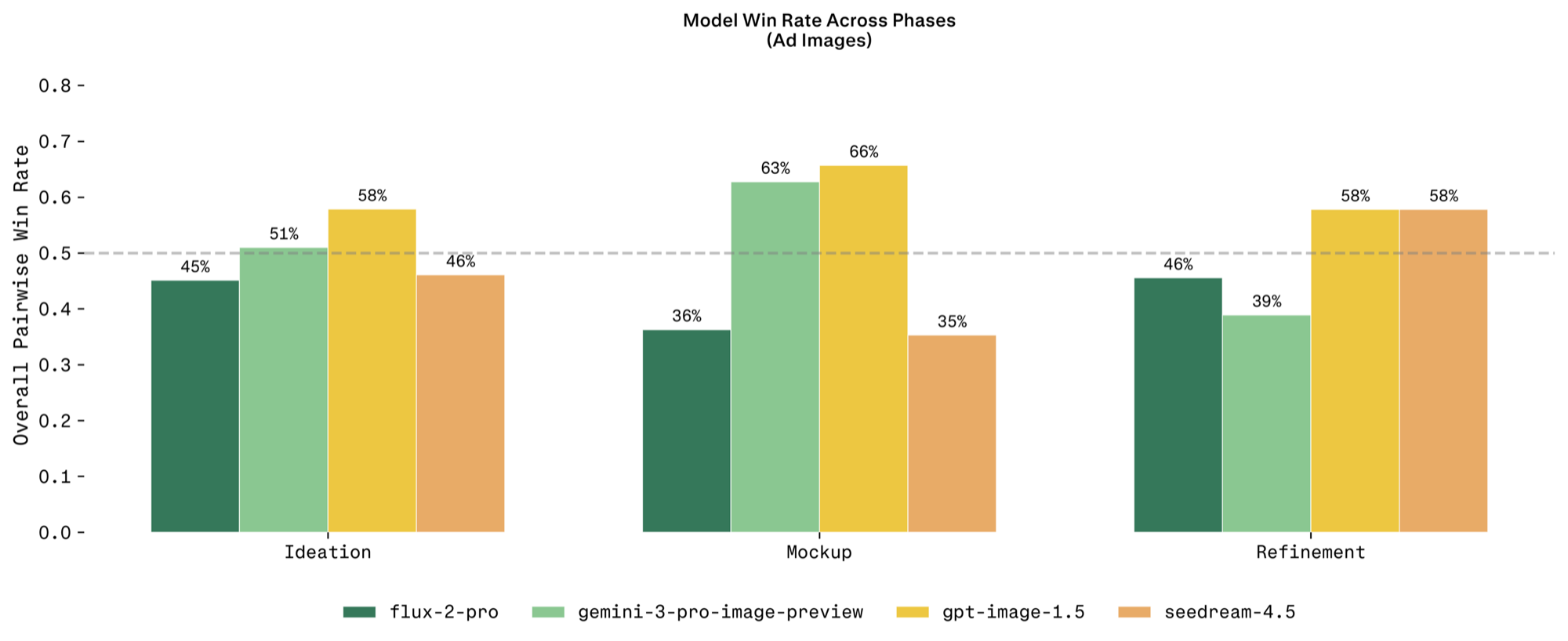}
    % \caption{Model win rates across the three-stage creative pipeline for Ad Image generation. Results illustrate phase-dependent performance fluctuations.}
    \caption{Model win rates across the three ad-image generation stages identified in the paper: Ideation, Mockup, and Refinement. GPT-Image-1.5 leads in Ideation (58\%), followed by Gemini-3-Pro-Image-Preview (51\%), Seedream-4.5 (46\%), and Flux-2-Pro (45\%). Mockup shows the same ordering at the top, with GPT-Image-1.5 (66\%) ahead of Gemini-3-Pro-Image-Preview (63\%), while Seedream-4.5 (35\%) and Flux-2-Pro (36\%) trail closely together. In Refinement, GPT-Image-1.5 and Seedream-4.5 tie for the lead (58\% each), followed by Flux-2-Pro (46\%) and then Gemini-3-Pro-Image-Preview (39\%). The dashed line marks the 50\% break-even point.}
    \label{fig:ad-image-win-rates}
\end{figure}

\begin{figure}[htbp]
    \centering
    \includegraphics[width=1\textwidth]{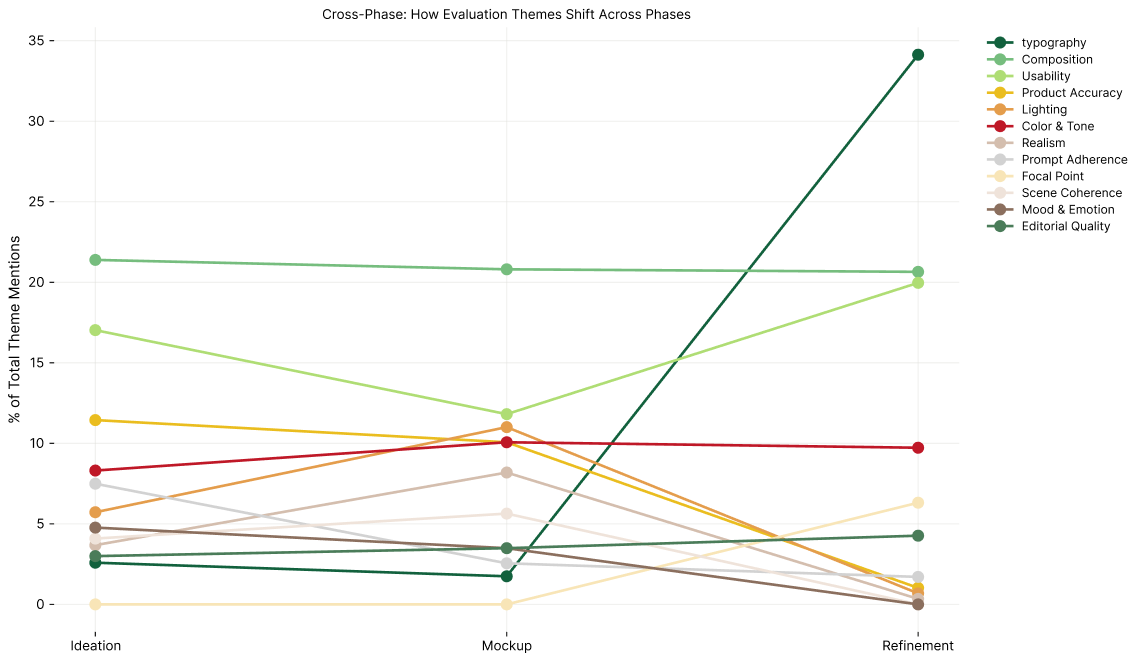}
    % \caption{Distribution of evaluation themes across the three-stage creative pipeline. Trends illustrate a shift in human evaluator priority from conceptual alignment (Prompt Adherence, Product Accuracy) in the Ideation phase to technical precision (Typography, Editorial Quality) during final Refinement.}
    \caption{Model win rates across the three ad-image generation stages identified in the paper: Ideation, Mockup, and Refinement. GPT-Image-1.5 leads in Ideation (58\%), followed by Gemini-3-Pro-Image-Preview (51\%), Seedream-4.5 (46\%), and Flux-2-Pro (45\%). Mockup shows the same ordering at the top, with GPT-Image-1.5 (66\%) ahead of Gemini-3-Pro-Image-Preview (63\%), while Seedream-4.5 (35\%) and Flux-2-Pro (36\%) trail closely together. In Refinement, GPT-Image-1.5 and Seedream-4.5 tie for the lead (58\% each), followed by Flux-2-Pro (46\%) and then Gemini-3-Pro-Image-Preview (39\%). The dashed line marks the 50\% break-even point.}
    \label{fig:eval-themes-pipeline}
\end{figure}

\begin{figure}[H]
    \centering
    \includegraphics[width=1\textwidth]{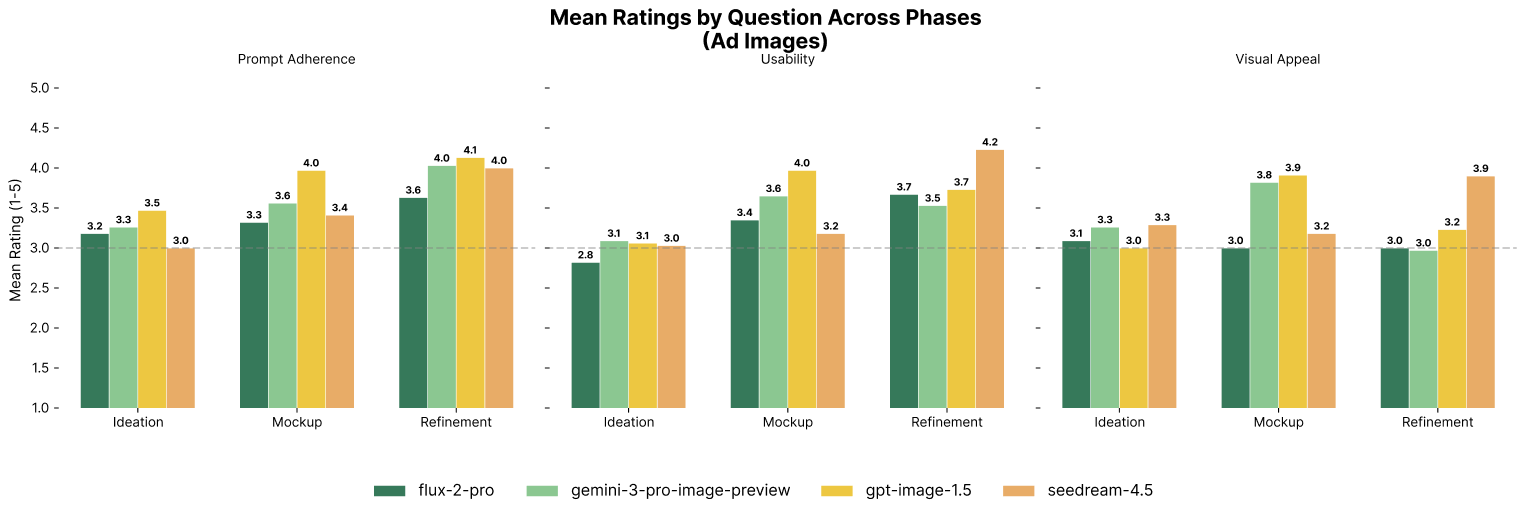}
    % \caption{Scalar rating distribution for Ad Image deliverables. Mean scores for Prompt Adherence, Usability, and Visual Appeal track model improvement from Ideation through Refinement. }
    \caption{Mean scalar ratings (1–5) for ad-image deliverables across the three pipeline stages, broken out by evaluation question. The dashed line marks the neutral midpoint (3.0).
Prompt Adherence. GPT-Image-1.5 leads at every stage: 3.5 in Ideation (ahead of Gemini-3-Pro-Image-Preview at 3.3, Flux-2-Pro at 3.2, and Seedream-4.5 at 3.0), 4.0 in Mockup (Gemini 3.6, Seedream 3.4, Flux 3.3), and 4.1 in Refinement, where Gemini and Seedream both reach 4.0 and Flux trails at 3.6.
Usability. In Ideation, Gemini and GPT-Image-1.5 tie at 3.1, with Seedream just behind at 3.0 and Flux lowest at 2.8. GPT-Image-1.5 jumps to 4.0 in Mockup, followed by Gemini (3.6), Flux (3.4), and Seedream (3.2). By Refinement, Seedream takes the lead at 4.2, ahead of GPT-Image-1.5 and Flux (tied at 3.7) and Gemini (3.5).
Visual Appeal. Gemini and Seedream share the Ideation lead at 3.3, ahead of Flux (3.1) and GPT-Image-1.5 (3.0). In Mockup, GPT-Image-1.5 (3.9) and Gemini (3.8) lead, followed by Seedream (3.2) and Flux (3.0). In Refinement, Seedream scores highest at 3.9, followed by GPT-Image-1.5 (3.2) and then Flux and Gemini, tied at 3.0.}
    \label{fig:ad-image-scalar-dist}
\end{figure}

\begin{figure}[H]
    \centering
    \includegraphics[width=1\textwidth]{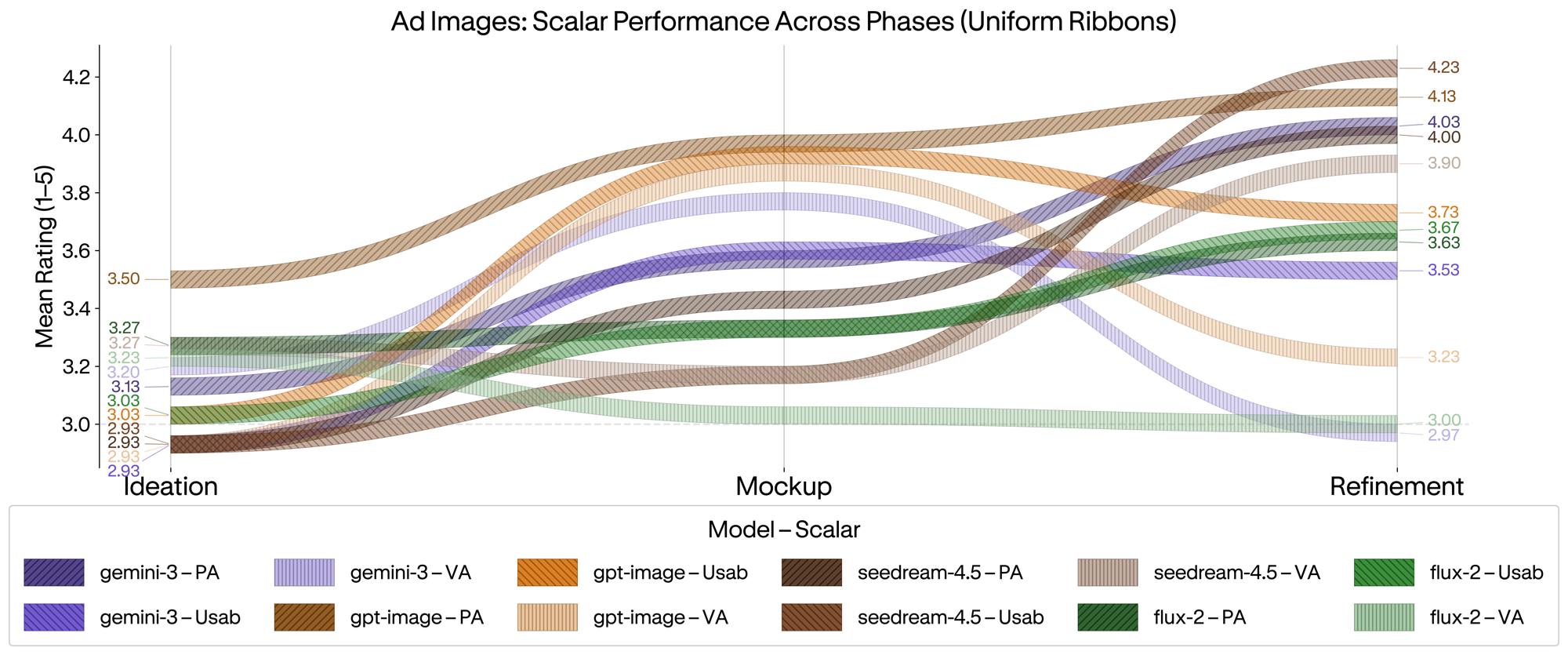}
    % \caption{Scalar performance trajectory for the Ad Images domain. Mean expert ratings are tracked through Ideation, Mockup, and Refinement phases. }
    \caption{Scalar performance across the three phases for ad-image generation, comprising one ribbon for each model–metric combination. Color encodes the model (Gemini-3 in purple, GPT-Image in orange, Seedream-4.5 in brown, and Flux-2 in green), while shade encodes the evaluation metric, with the lightest ribbon denoting Visual Appeal (VA), the medium shade Usability (Usab), and the darkest shade Prompt Adherence (PA); these abbreviations (PA, Usab, VA) label the ribbons in the legend. Ribbon width is uniform and does not encode uncertainty.Most ribbons rise from a compressed band at Ideation, spanning roughly 2.9 to 3.5, and broaden into a wider spread of approximately 3.0 to 4.2 by Refinement. Seedream-4.5 on Visual Appeal shows the largest improvement, climbing from 3.50 at Ideation to 4.23 at Refinement, the highest score overall, with GPT-Image on Prompt Adherence close behind at 4.13. Several metrics decline in the final phase, most notably GPT-Image on Visual Appeal, which falls to 3.23, alongside Flux-2 and Gemini-3 on Visual Appeal, which settle near the 3.0 midpoint. The dashed line denotes the neutral midpoint of 3.0.}
    \label{fig:ad-image-scalar-traj}
\end{figure}

\begin{figure}[htbp]
    \centering
    \includegraphics[width=1\textwidth]{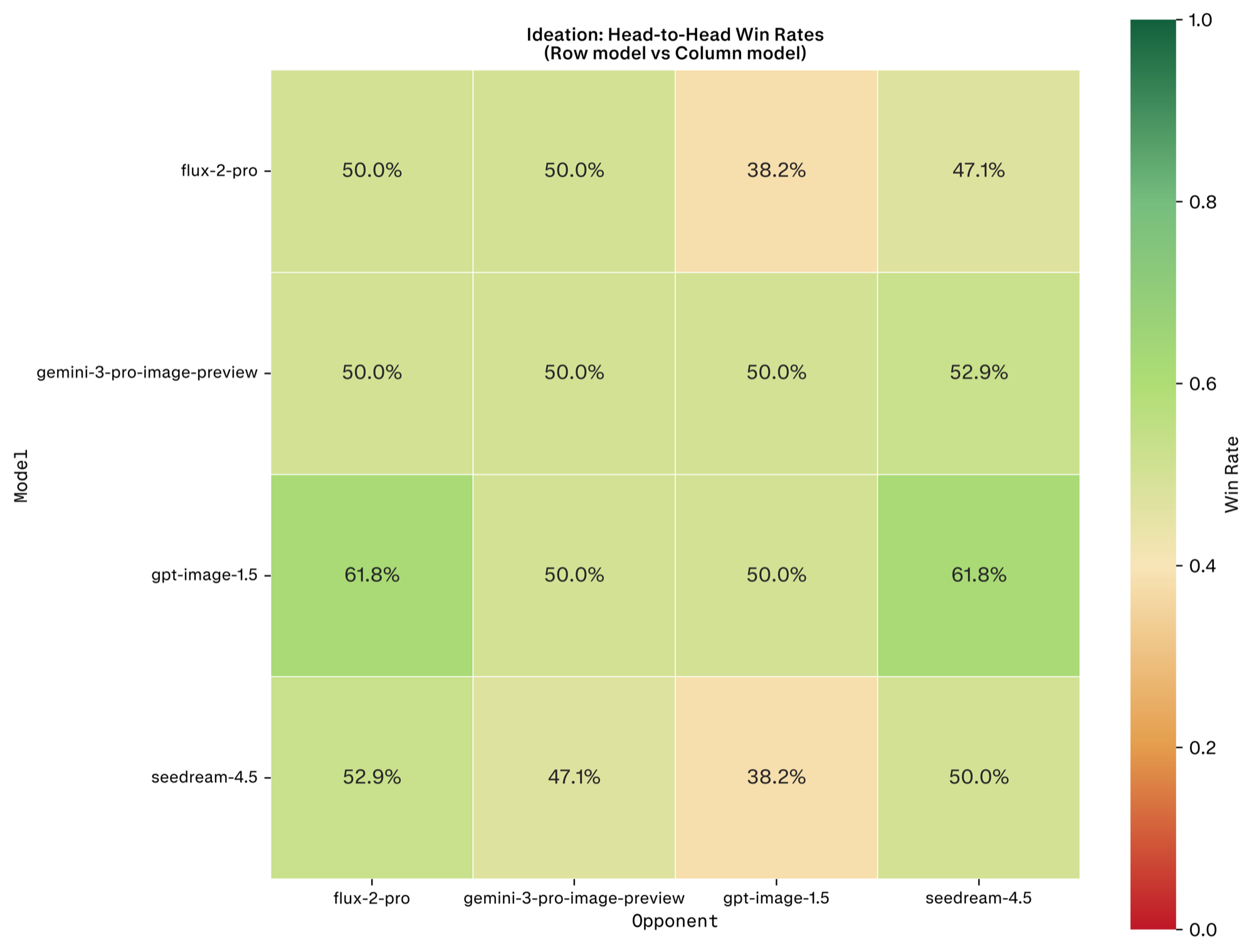}
    \caption{Head-to-head pairwise win rates for ad-image generation in the Ideation stage. Each cell reports the row model's win rate against the column model; diagonal self-matches are excluded.}
    \label{fig:adimg-h2h-ideation}
\end{figure}

\begin{figure}[htbp]
    \centering
    \includegraphics[width=1\textwidth]{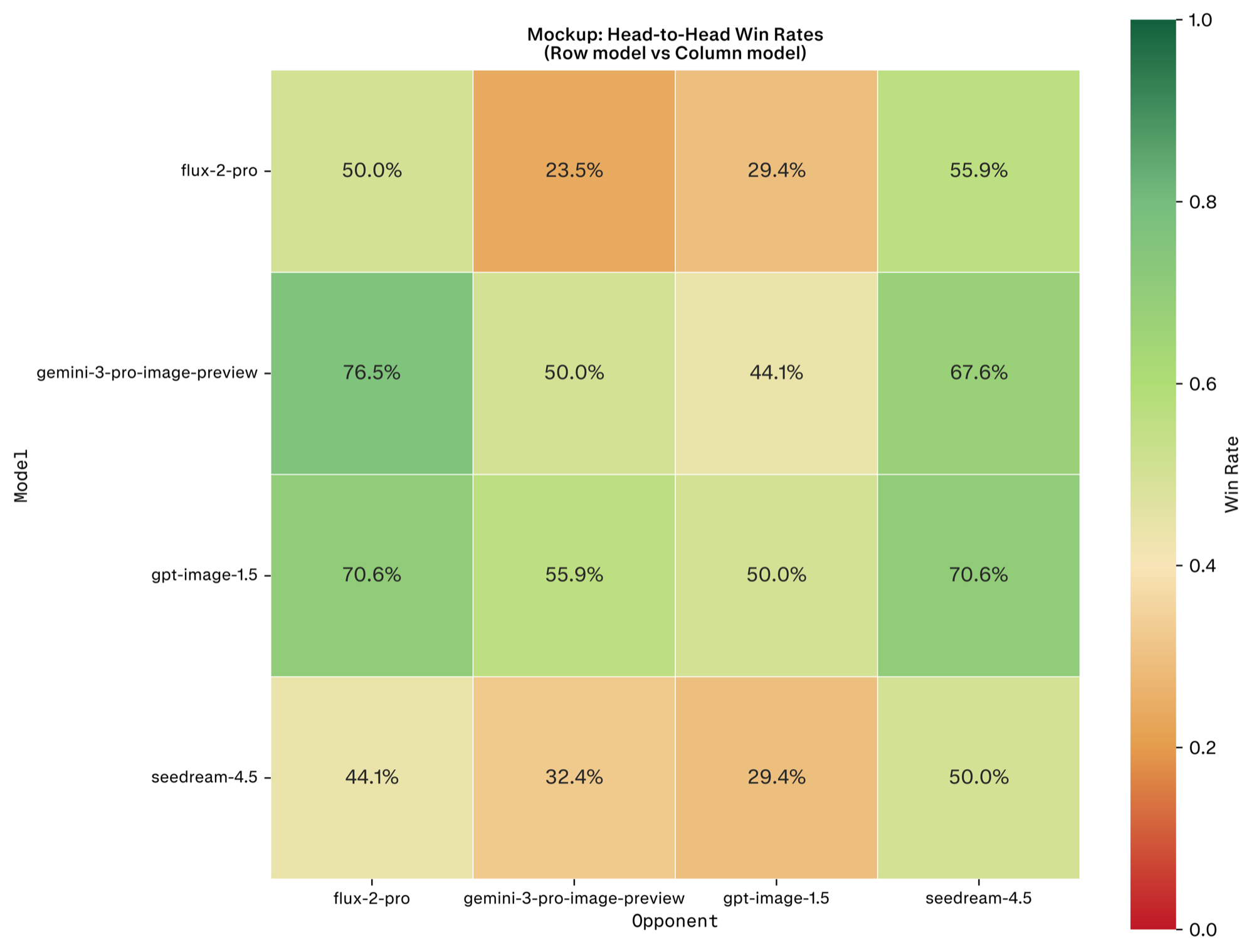}
    \caption{Head-to-head pairwise win rates for ad-image generation in the Mockup stage. Each cell reports the row model's win rate against the column model; diagonal self-matches are excluded.}
    \label{fig:adimg-h2h-mockup}
\end{figure}

\begin{figure}[htbp]
    \centering
    \includegraphics[width=1\textwidth]{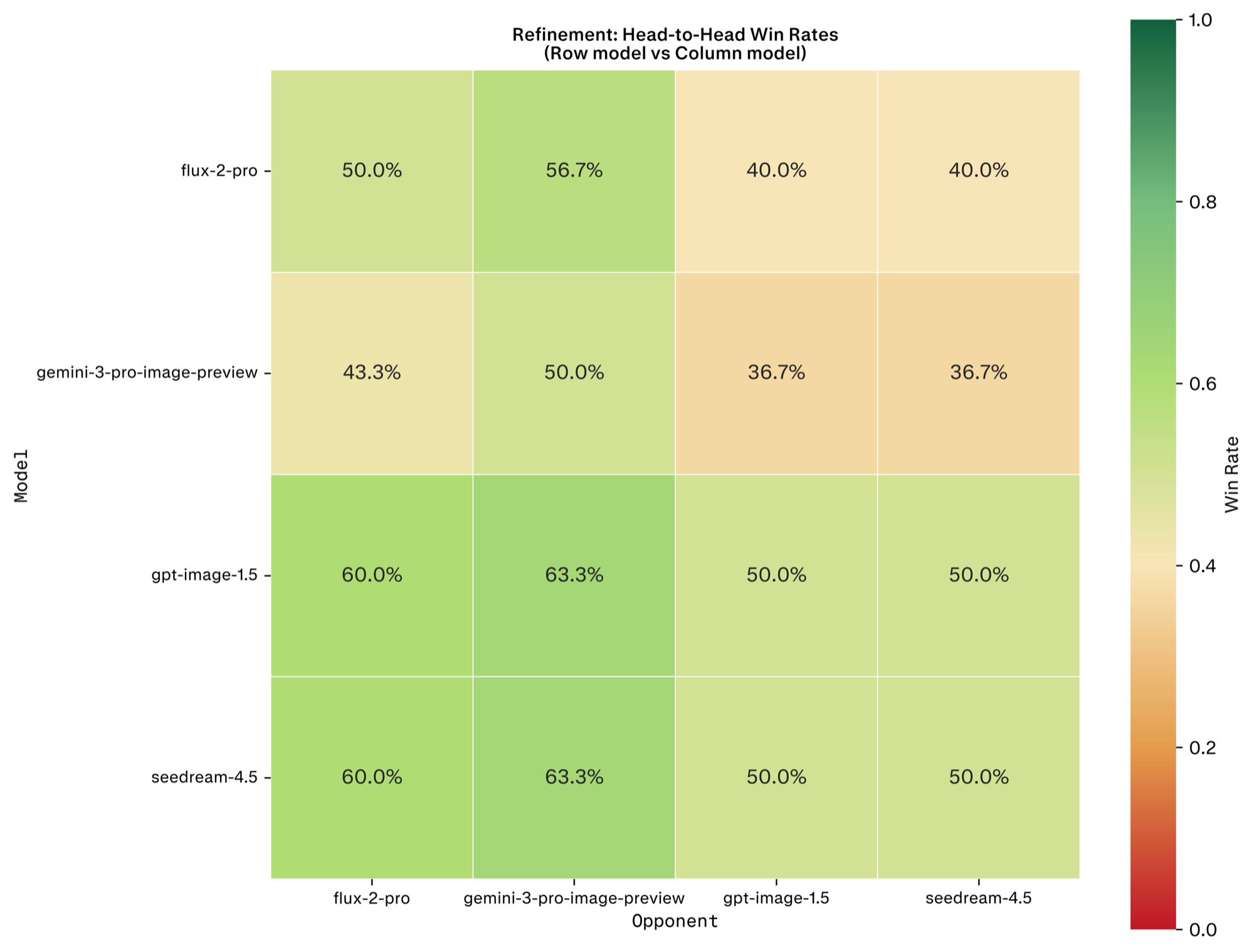}
    \caption{Head-to-head pairwise win rates for ad-image generation in the Refinement stage. Each cell reports the row model's win rate against the column model; diagonal self-matches are excluded.}
    \label{fig:adimg-h2h-refinement}
\end{figure}

Figure~\ref{fig:ad-image-win-rates} reports phase-level win rates. Figure~\ref{fig:eval-themes-pipeline} tracks how evaluation themes shift across phases. Figures~\ref{fig:ad-image-scalar-dist} and~\ref{fig:ad-image-scalar-traj} summarize scalar ratings and trajectories, and Figures~\ref{fig:adimg-h2h-ideation}, \ref{fig:adimg-h2h-mockup}, and~\ref{fig:adimg-h2h-refinement} give head-to-head pairwise win rates for the Ideation, Mockup, and Refinement stages.

\subsection{Ad Video}
\begin{figure}[htbp]
    \centering
    \includegraphics[width=1\textwidth]{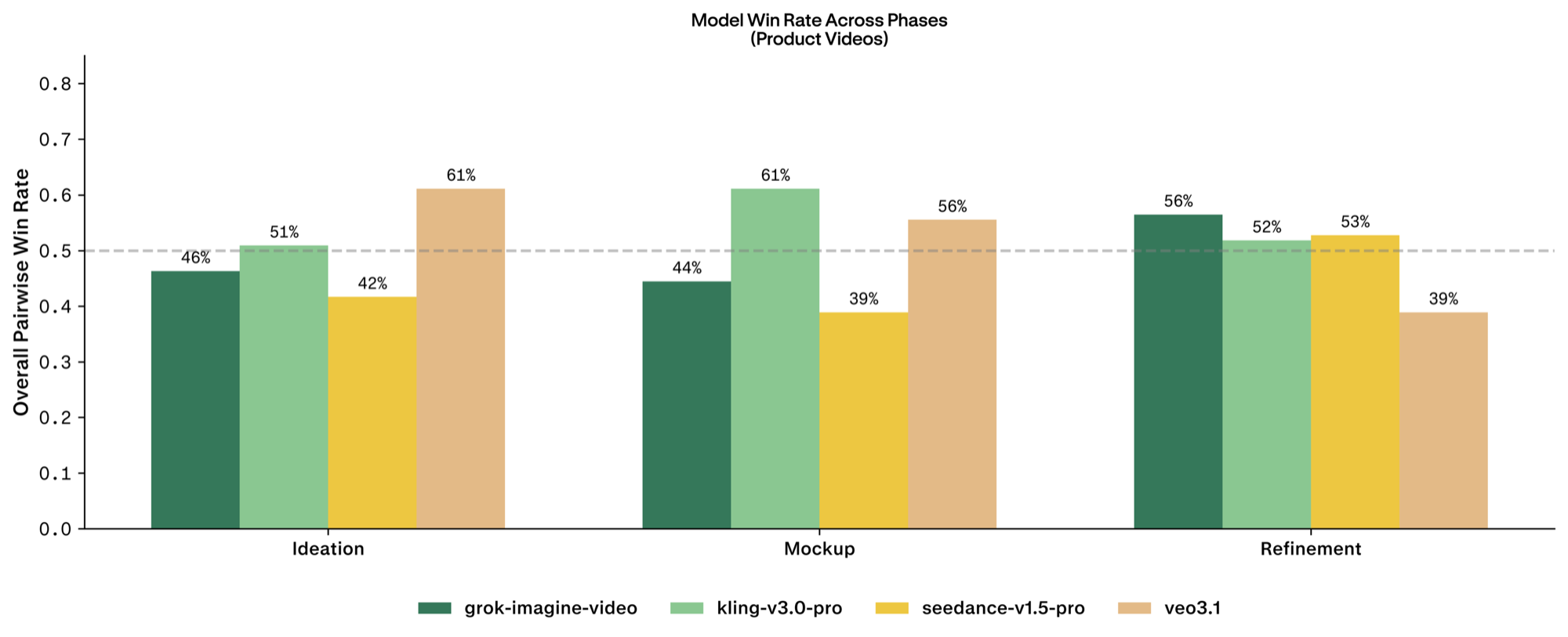}
    % \caption{Model win rates for Product Video generation across workflow phases.}
    \caption{Overall pairwise win rates across the three pipeline stages for product-video generation. Veo3.1 leads Ideation at 61\%, followed by Kling-v3.0-pro (51\%), Grok-Imagine-Video (46\%), and Seedance-v1.5-pro (42\%). In Mockup, Kling-v3.0-pro rises to the top at 61\%, ahead of Veo3.1 (56\%), Grok-Imagine-Video (44\%), and Seedance-v1.5-pro (39\%). By Refinement the order reshuffles again: Grok-Imagine-Video leads at 56\%, with Kling-v3.0-pro and Seedance-v1.5-pro nearly tied just above break-even (52\% and 53\%), and Veo3.1 falling to last at 39\%. The dashed line marks the 50\% break-even point.}
    \label{fig:video-win-rates}
\end{figure}

\begin{figure}[htbp]
    \centering
    \includegraphics[width=1\textwidth]{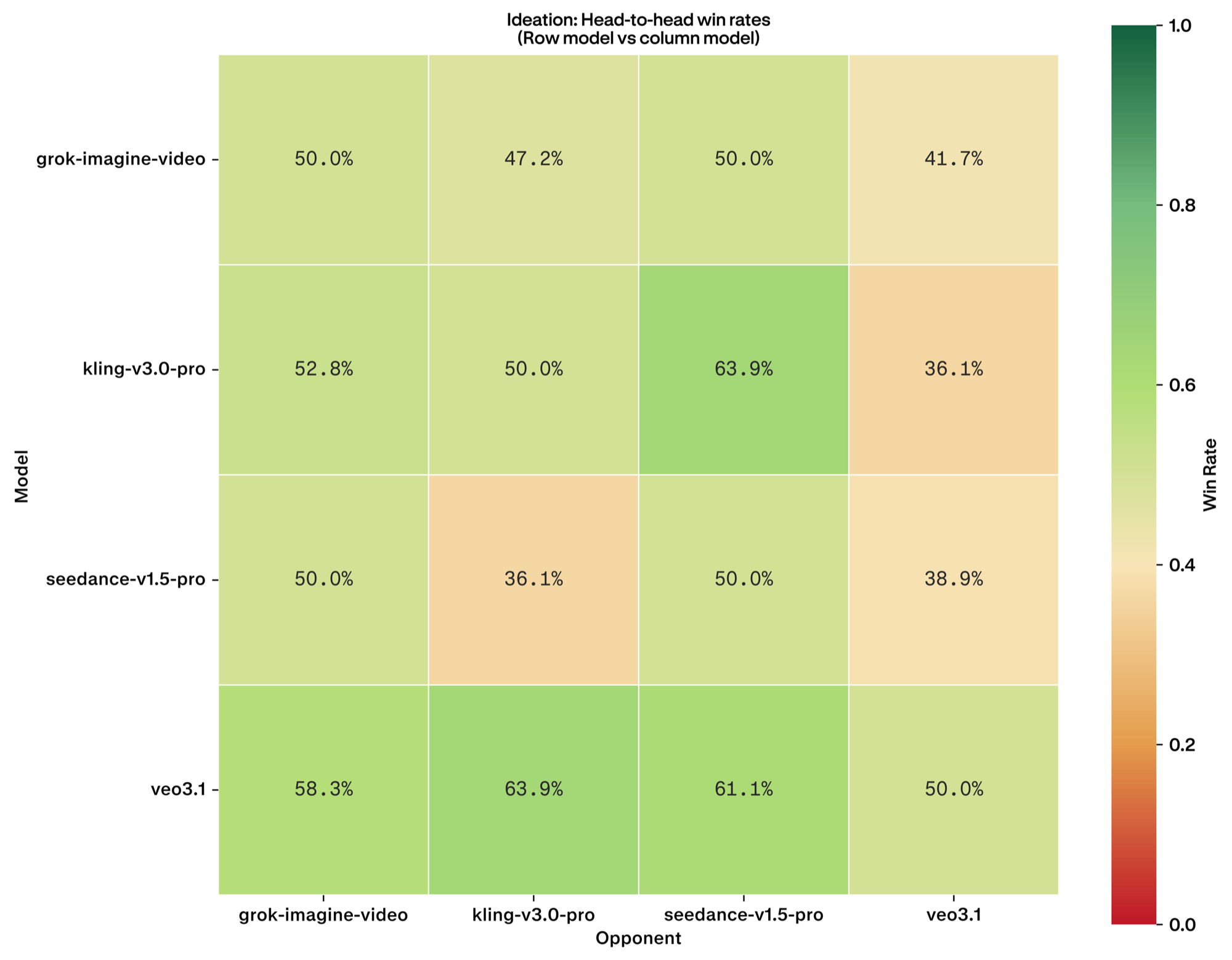}
    \caption{Head-to-head pairwise win rates for product-video generation in the Ideation stage. Each cell reports the row model's win rate against the column model; diagonal self-matches are excluded.}
    \label{fig:advideo-h2h-ideation}
\end{figure}

\begin{figure}[htbp]
    \centering
    \includegraphics[width=1\textwidth]{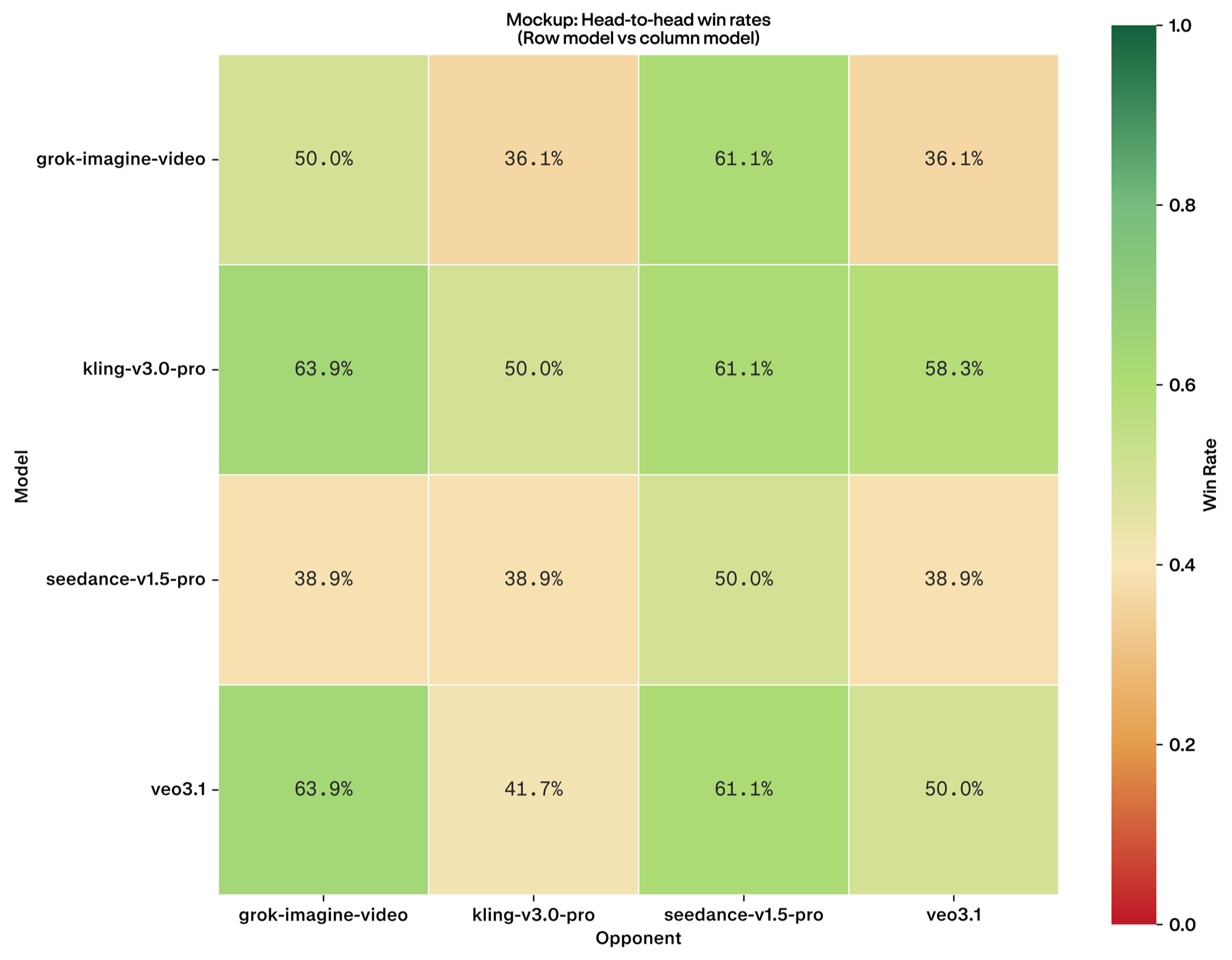}
    \caption{Head-to-head pairwise win rates for product-video generation in the Mockup stage. Each cell reports the row model's win rate against the column model; diagonal self-matches are excluded.}
    \label{fig:advideo-h2h-mockup}
\end{figure}

\begin{figure}[htbp]
    \centering
    \includegraphics[width=1\textwidth]{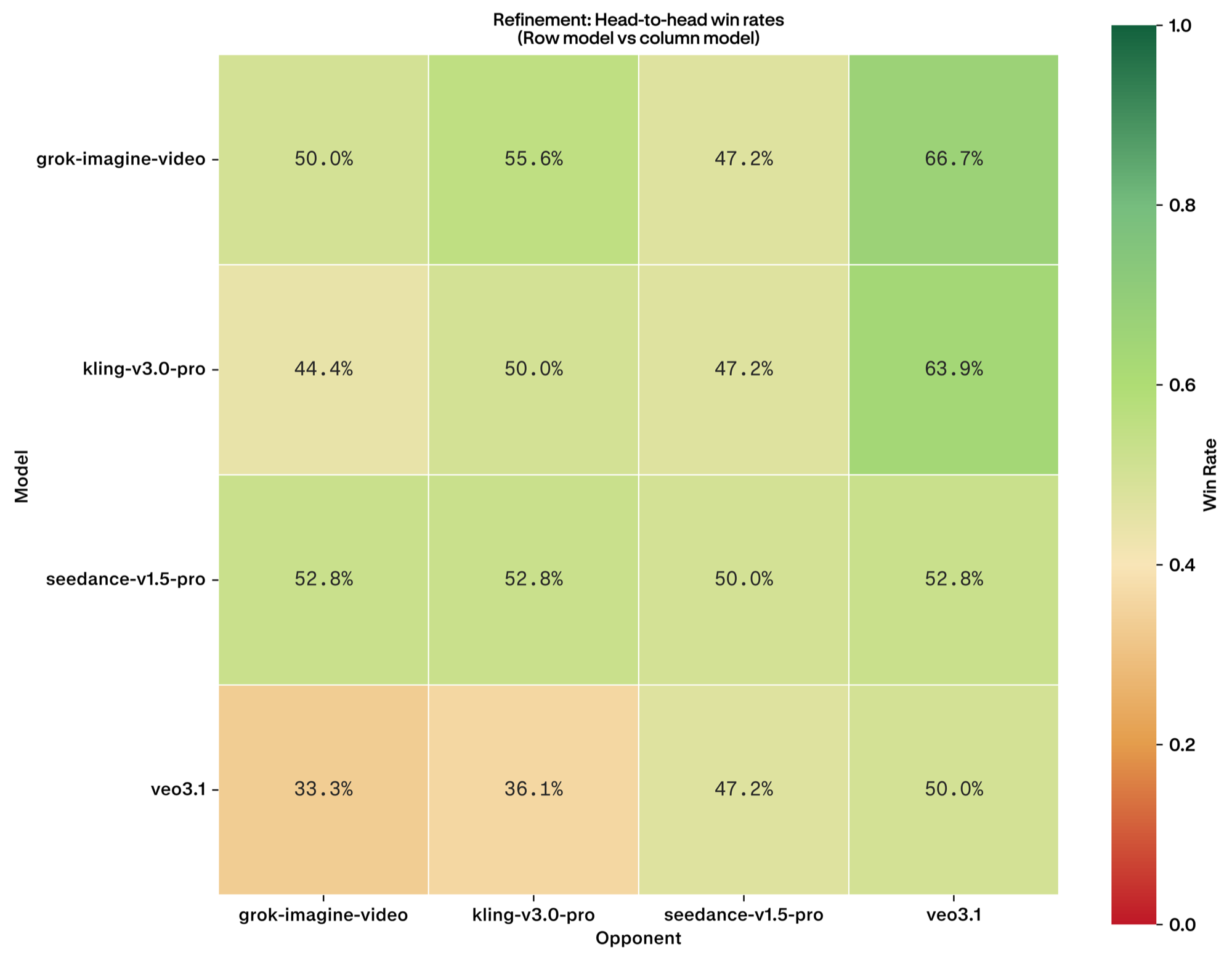}
    \caption{Head-to-head pairwise win rates for product-video generation in the Refinement stage. Each cell reports the row model's win rate against the column model; diagonal self-matches are excluded.}
    \label{fig:advideo-h2h-refinement}
\end{figure}

Figure~\ref{fig:video-win-rates} reports pairwise win rates across the three pipeline stages. Figures~\ref{fig:advideo-h2h-ideation}, \ref{fig:advideo-h2h-mockup}, and~\ref{fig:advideo-h2h-refinement} give head-to-head pairwise win rates for the Ideation, Mockup, and Refinement stages.

\subsection{Brand Assets}
\begin{figure}[H]
    \centering
    \includegraphics[width=1\textwidth]{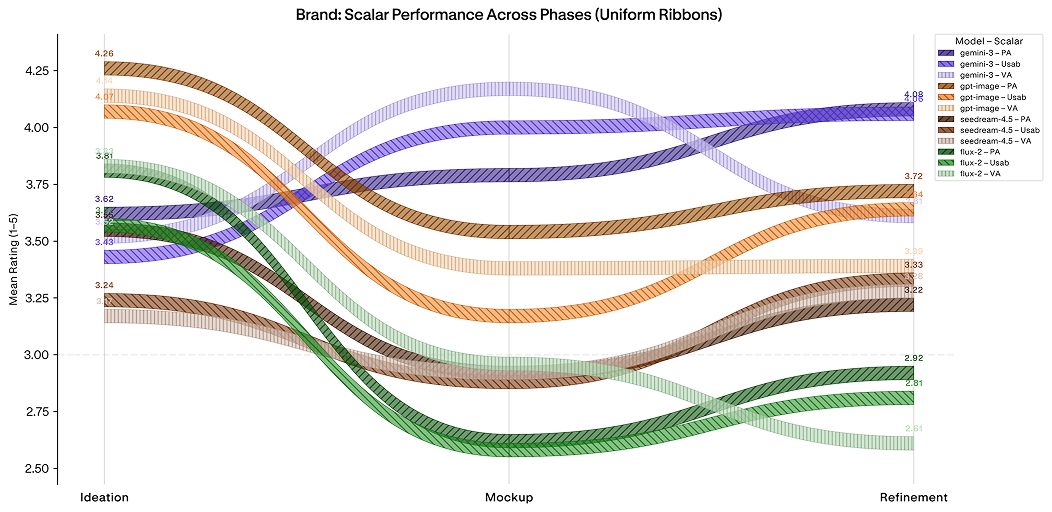}
    \caption{Example overview of scalar performance for Brand Design content; y-axis marks the stage, while x-axis marks mean rating, separated by metric (Prompt Adherence, Usability, and Visual Appeal) and model. Similar charts are shown in Appendix for each category produced.}
    \label{fig:brand-scalar-ribbons}
\end{figure}

\begin{figure}[H]
    \centering
    \includegraphics[width=1\textwidth]{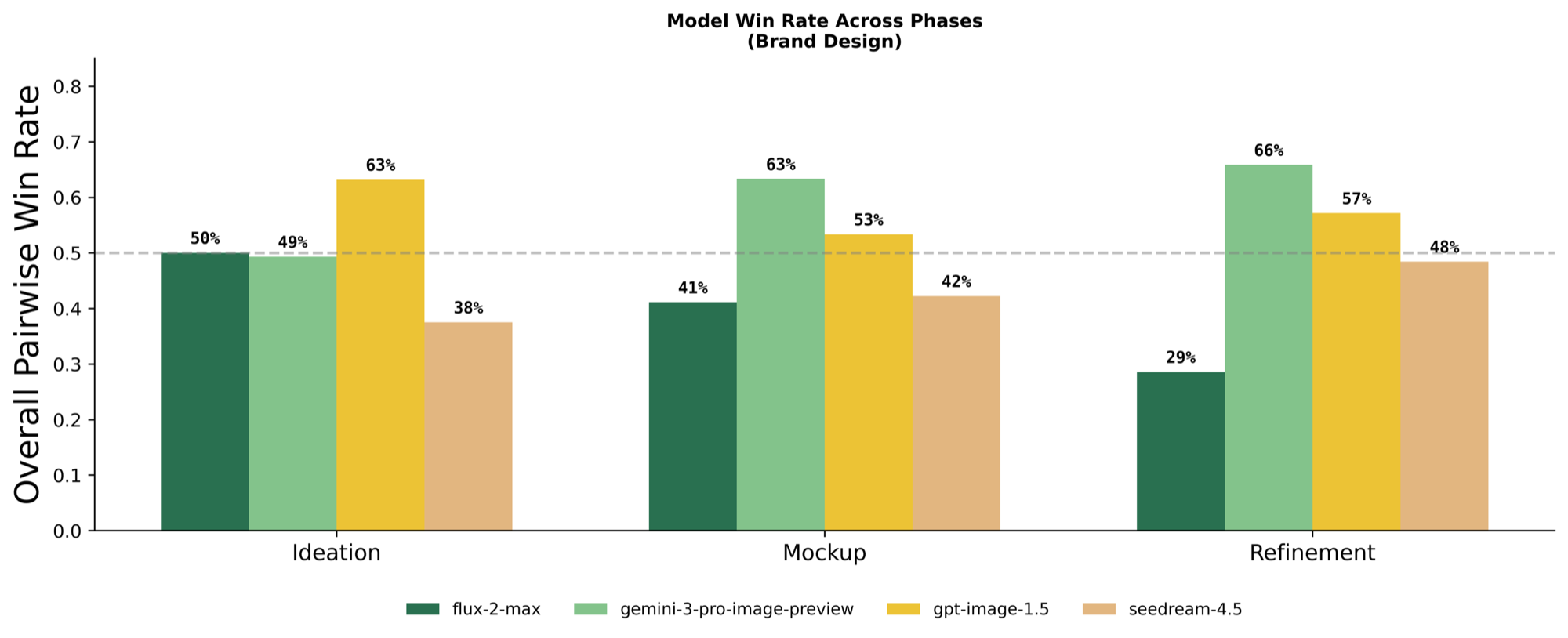}
    % \caption{Model win rates for the Brand Design domain across workflow phases.}
    \caption{Model win rates across the three phases for brand-design generation. GPT-Image-1.5 leads Ideation at 63\%, followed by Flux-2-max at 50\%, Gemini-3-Pro-Image-Preview at 49\%, and Seedream-4.5 at 38\%. In Mockup, Gemini-3-Pro-Image-Preview takes the leads with 63\%, followed by GPT-Image-1.5 at 53\%, followed by Seedream-4.5 and Flux-2-max at 42\%, and 41\%. By Refinement, Gemini-3-Pro-Image-Preview leads again at 66\%, followed by GPT-Image-1.5 at 57\%, Seedream-4.5 at 48\%, and finally Flux-2-max at 29\%.}
    \label{fig:brand-win-rates}
\end{figure}

\begin{figure}[htbp]
    \centering
    \includegraphics[width=1\textwidth]{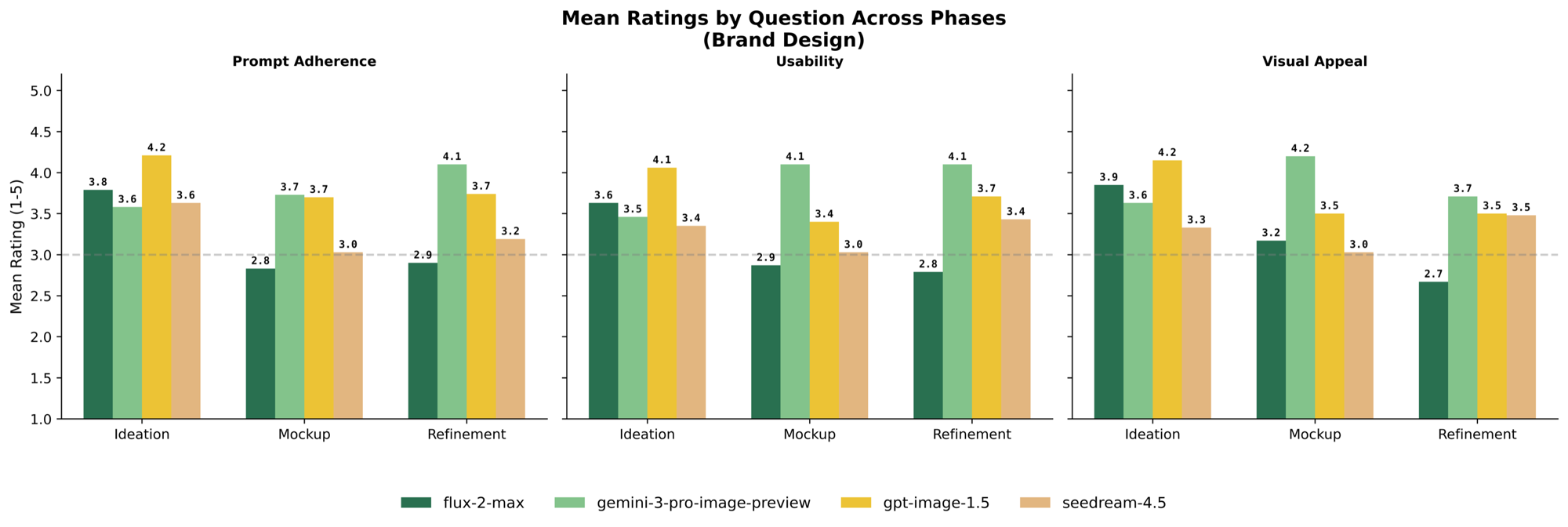}
    % \caption{Scalar rating distribution for Brand Design deliverables by phase. Mean scores illustrate the progression of model performance.}
    \caption{Overall pairwise win rates across the three pipeline stages for brand-design generation. GPT-Image-1.5 leads Ideation at 63\%, followed by Flux-2-max (50\%), Gemini-3-Pro-Image-Preview (49\%), and Seedream-4.5 (38\%). In Mockup, Gemini-3-Pro-Image-Preview rises to the top at 63\%, ahead of GPT-Image-1.5 (53\%), Seedream-4.5 (42\%), and Flux-2-max (41\%). By Refinement, Gemini-3-Pro-Image-Preview extends its lead to 66\%, followed by GPT-Image-1.5 (57\%) and Seedream-4.5 (48\%), while Flux-2-max falls to 29\%. The dashed line marks the 50\% break-even point.}
    \label{fig:brand-scalar-dist}
\end{figure}

\begin{figure}[htbp]
    \centering
    \includegraphics[width=1\textwidth]{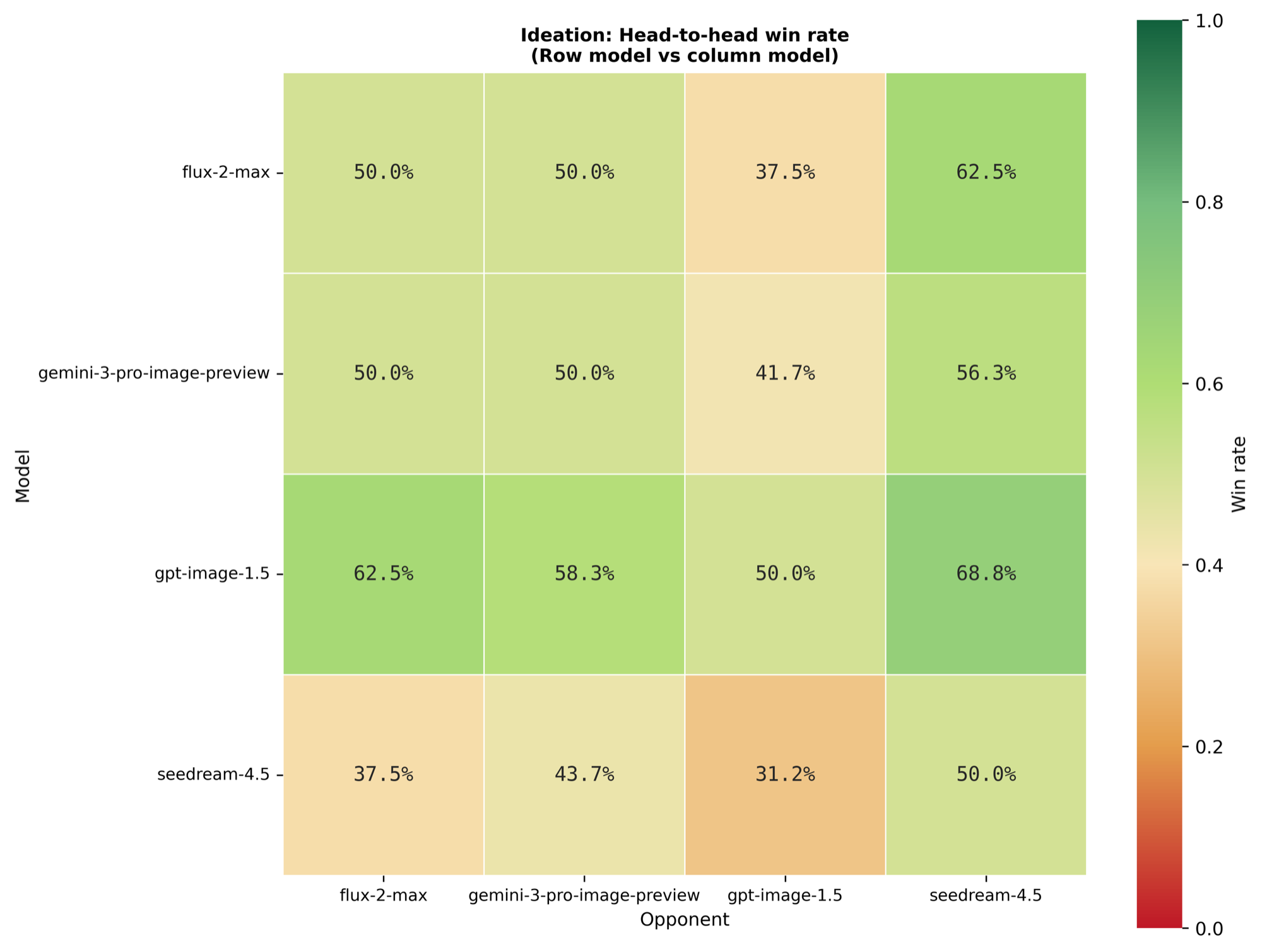}
    \caption{Head-to-head pairwise win rates for brand-design generation in the Ideation stage. Each cell reports the row model's win rate against the column model; diagonal self-matches are excluded.}
    \label{fig:brand-h2h-ideation}
\end{figure}

\begin{figure}[htbp]
    \centering
    \includegraphics[width=1\textwidth]{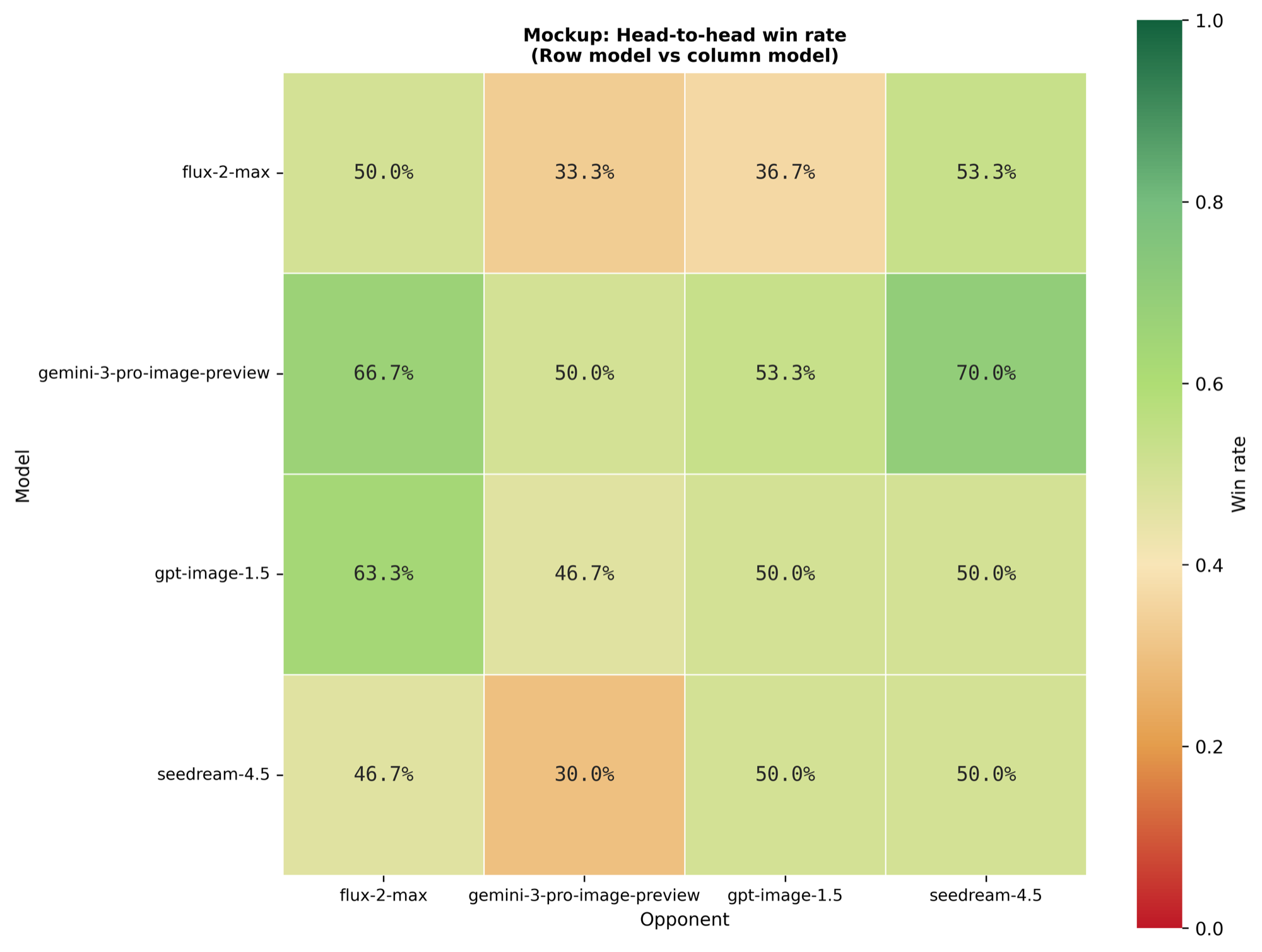}
    \caption{Head-to-head pairwise win rates for brand-design generation in the Mockup stage. Each cell reports the row model's win rate against the column model; diagonal self-matches are excluded.}
    \label{fig:brand-h2h-mockup}
\end{figure}

\begin{figure}[htbp]
    \centering
    \includegraphics[width=1\textwidth]{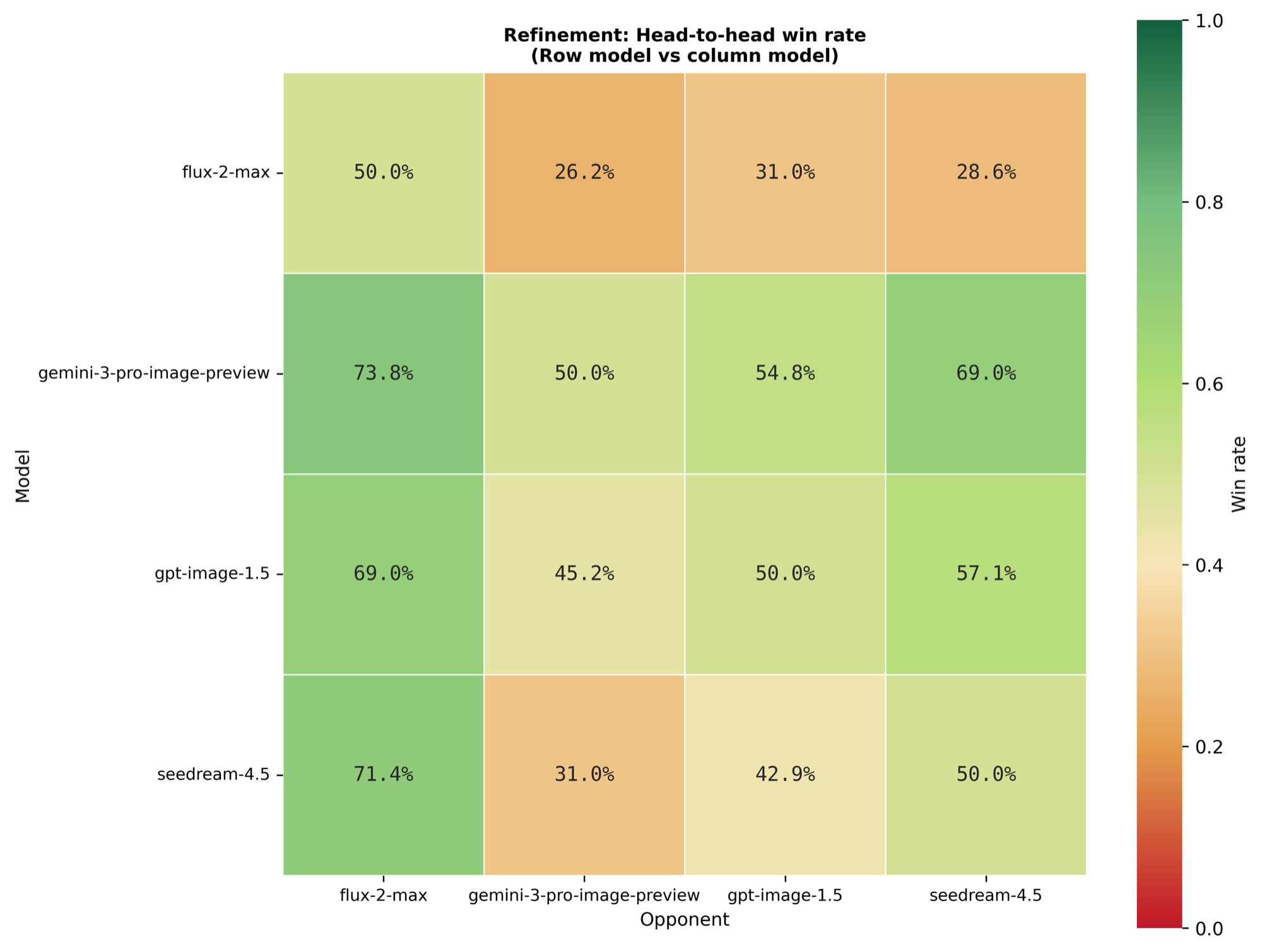}
    \caption{Head-to-head pairwise win rates for brand-design generation in the Refinement stage. Each cell reports the row model's win rate against the column model; diagonal self-matches are excluded.}
    \label{fig:brand-h2h-refinement}
\end{figure}

Figures~\ref{fig:brand-scalar-ribbons}, \ref{fig:brand-win-rates}, and~\ref{fig:brand-scalar-dist} summarize scalar performance, win rates, and mean ratings for brand design, and Figures~\ref{fig:brand-h2h-ideation}, \ref{fig:brand-h2h-mockup}, and~\ref{fig:brand-h2h-refinement} give head-to-head pairwise win rates for the Ideation, Mockup, and Refinement stages.

\subsection{Desktop Apps}
\begin{figure}[htbp]
    \centering
    \includegraphics[width=1\textwidth]{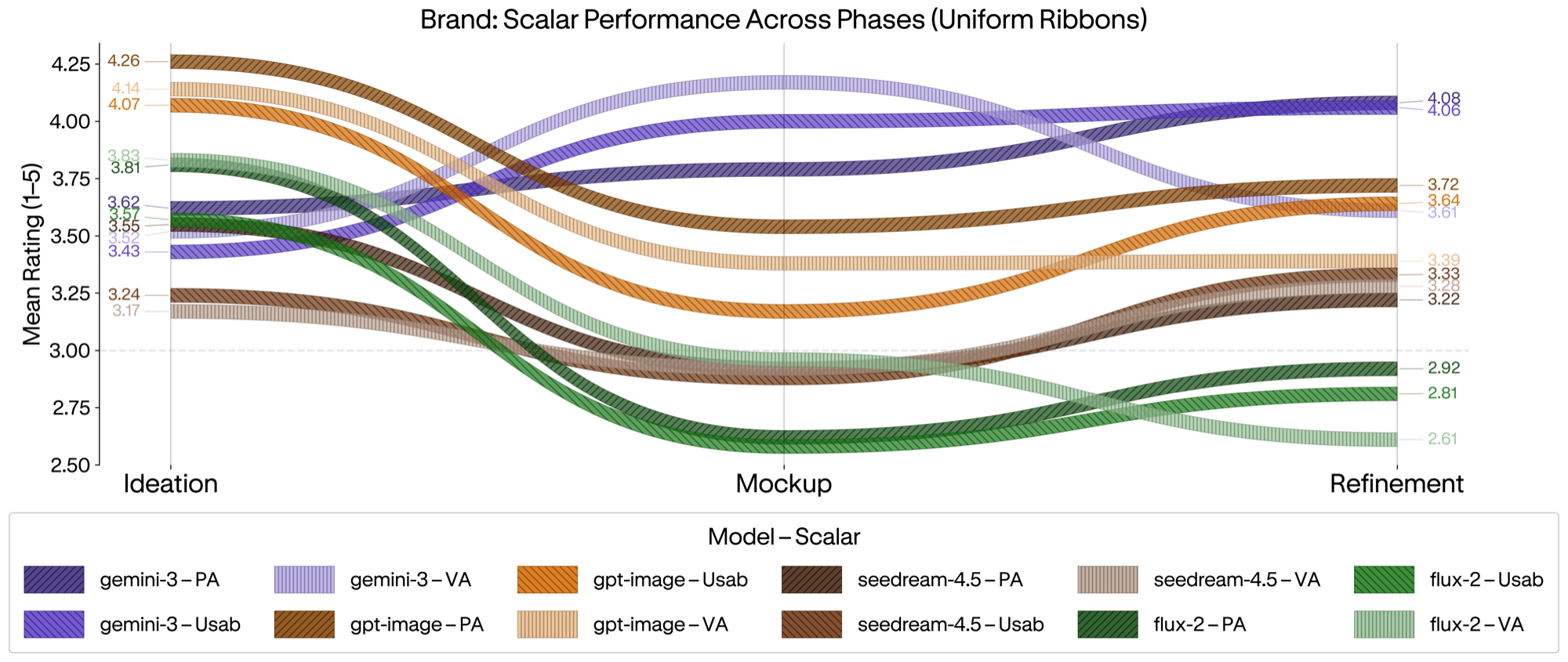}
    % \caption{Longitudinal mean scalar ratings for the Desktop Apps domain. Colored ribbons represent model--dimension pairs across the three-stage creative pipeline; the three evaluated dimensions are Prompt Adherence (PA), Usability (Usab), and Visual Appeal (VA).}
    \caption{Scalar performance across the three phases for brand-design generation, comprising one ribbon for each model–metric combination (twelve in total). Color encodes the model (Gemini-3 in purple, GPT-Image in orange, Seedream-4.5 in brown, and Flux-2 in green), while shade encodes the evaluation metric, with the lightest ribbon denoting Visual Appeal (VA), the medium shade Usability (Usab), and the darkest shade Prompt Adherence (PA); these abbreviations (PA, Usab, VA) label the ribbons in the legend. Ribbon width is uniform.GPT-Image records the highest scores at Ideation, led by Prompt Adherence at 4.26, but its three metrics decline over the following phases. Gemini-3 moves in the opposite direction, rising to the top by Refinement with Prompt Adherence at 4.08 and Usability at 4.06. Flux-2 falls steadily across phases to the lowest scores in the final phase, reaching 2.92 on Prompt Adherence, 2.81 on Usability, and 2.61 on Visual Appeal. Seedream-4.5 remains in the middle of the range throughout. The dashed line denotes the neutral midpoint of 3.0.}
    \label{fig:desktop-scalar-appendix}
\end{figure}

\begin{figure}[htbp]
    \centering
    \includegraphics[width=1\textwidth]{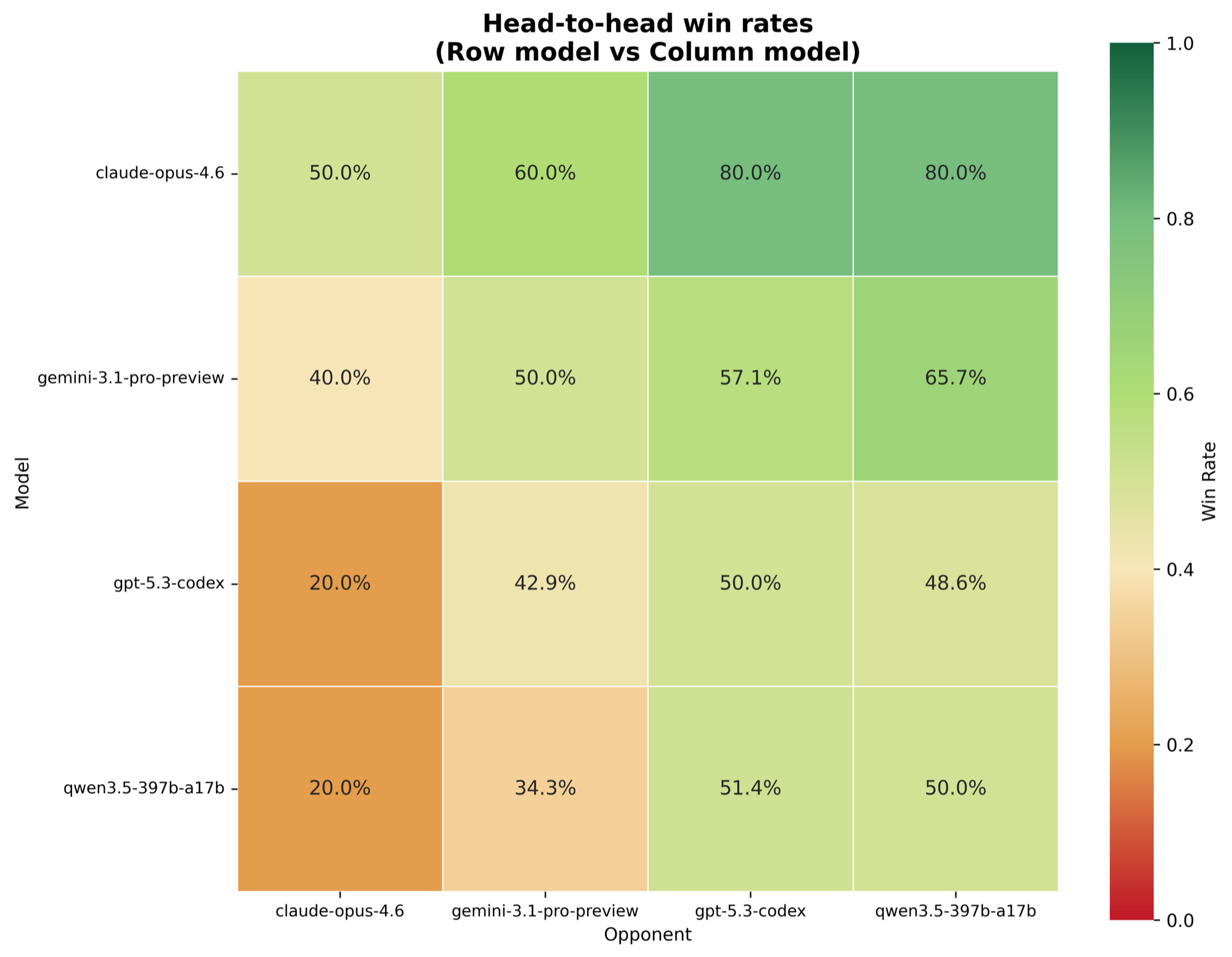}
    \caption{Head-to-head pairwise win rates for desktop-app generation in the Ideation stage. Each cell reports the row model's win rate against the column model; diagonal self-matches are excluded.}
    \label{fig:desktop-h2h-ideation}
\end{figure}

\begin{figure}[htbp]
    \centering
    \includegraphics[width=1\textwidth]{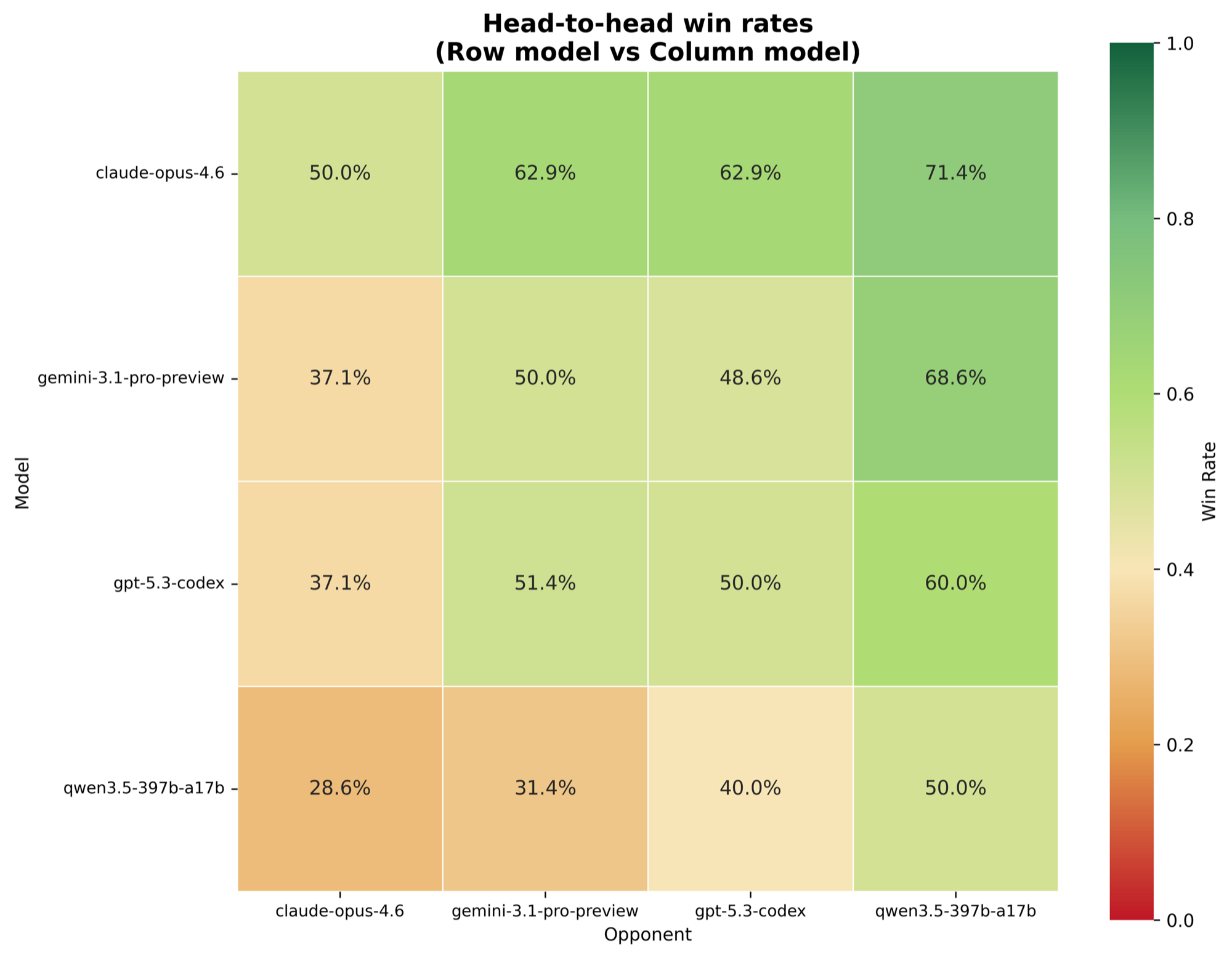}
    \caption{Head-to-head pairwise win rates for desktop-app generation in the Mockup stage. Each cell reports the row model's win rate against the column model; diagonal self-matches are excluded.}
    \label{fig:desktop-h2h-mockup}
\end{figure}

\begin{figure}[htbp]
    \centering
    \includegraphics[width=1\textwidth]{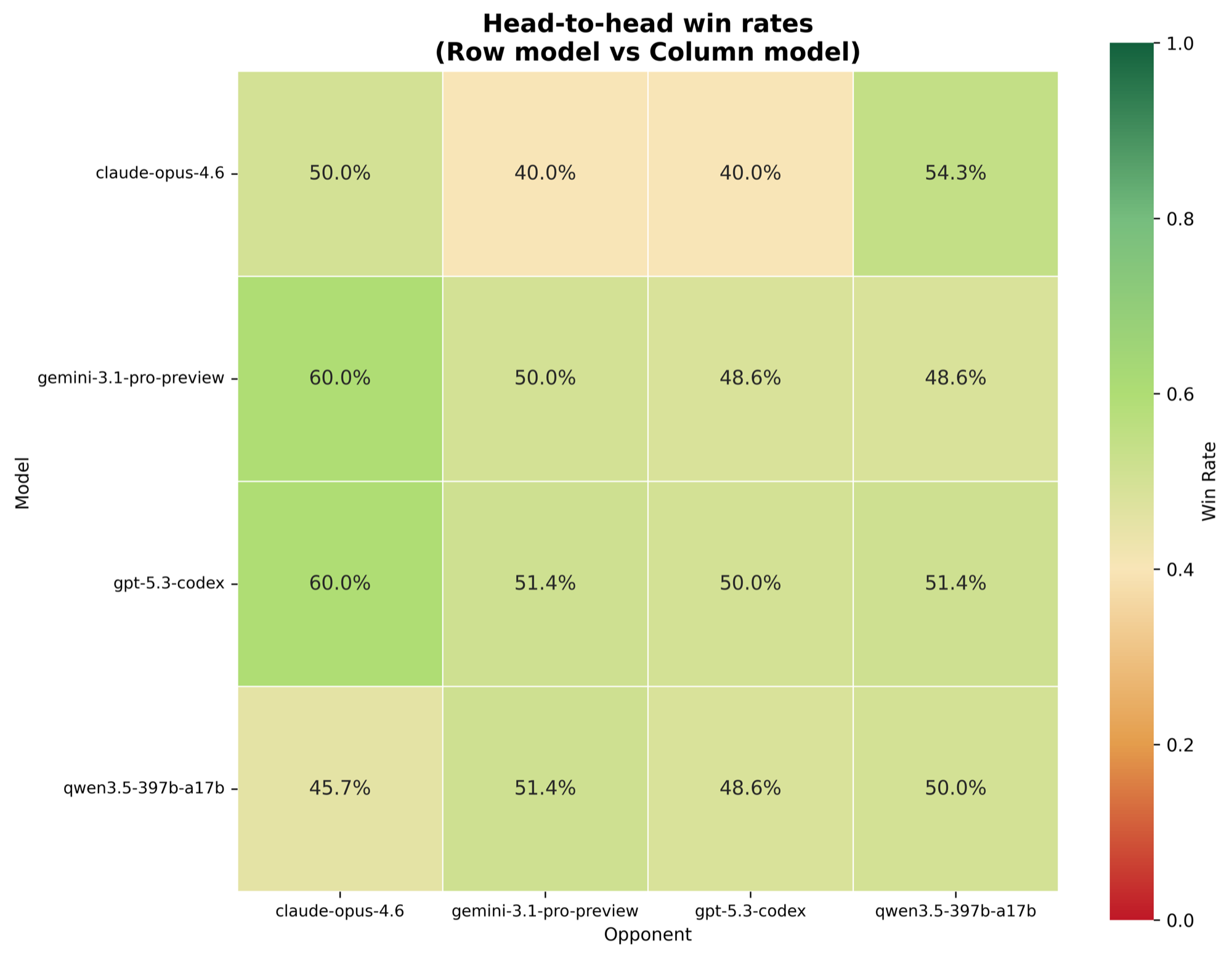}
    \caption{Head-to-head pairwise win rates for desktop-app generation in the Refinement stage. Each cell reports the row model's win rate against the column model; diagonal self-matches are excluded.}
    \label{fig:desktop-h2h-refinement}
\end{figure}

Figure~\ref{fig:desktop-scalar-appendix} tracks scalar performance across the three pipeline stages. Figures~\ref{fig:desktop-h2h-ideation}, \ref{fig:desktop-h2h-mockup}, and~\ref{fig:desktop-h2h-refinement} give head-to-head pairwise win rates for the Ideation, Mockup, and Refinement stages.

\subsection{Landing Pages}
\begin{figure}[H]
    \centering
    \includegraphics[width=1\textwidth]{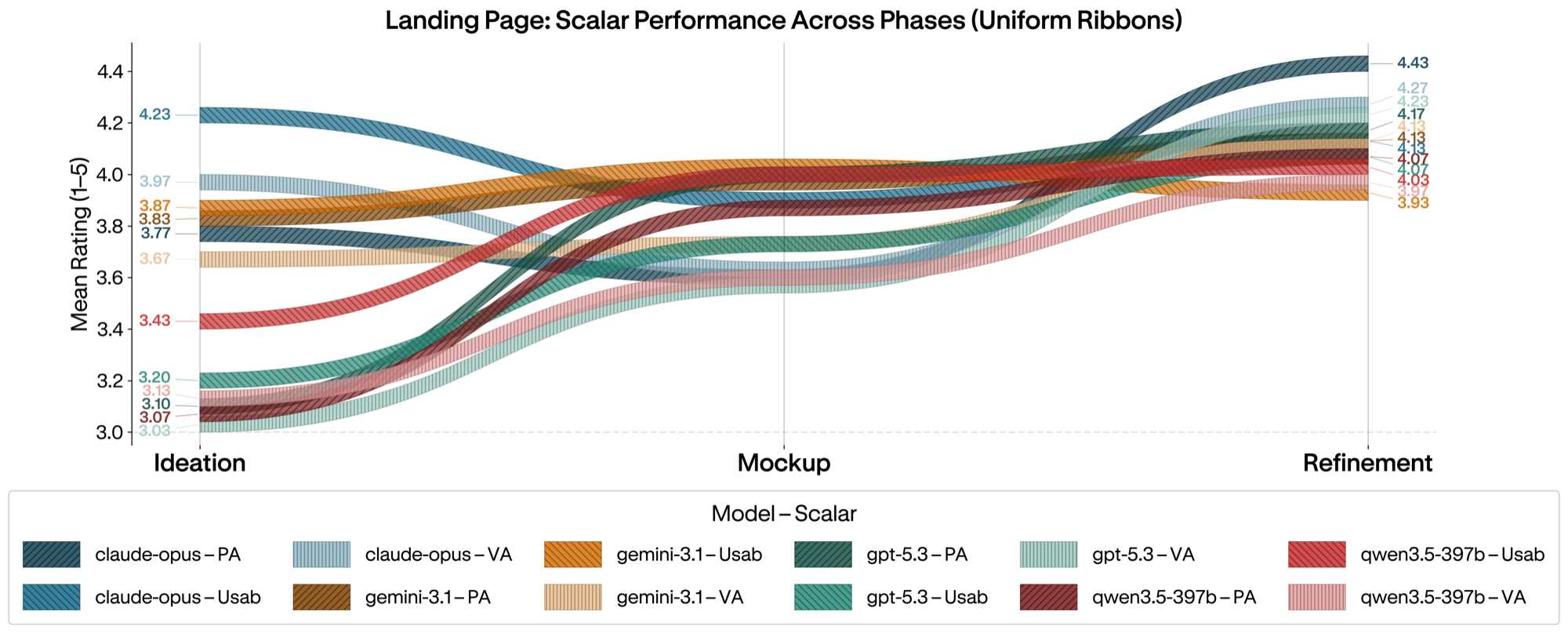}
    % \caption{Mean scalar ratings for Landing Page generation across three workflow phases. Colored ribbons differentiate between models and evaluation dimensions (PA, Usab, VA).}
    \caption{Scalar performance across the three phases for landing-page generation, with one ribbon for each model–metric combination. Color encodes the model (Claude-Opus in blue, Gemini-3.1 in orange, GPT-5.3 in teal, and Qwen3.5-397b in red), while shade encodes the evaluation metric, with the lightest ribbon denoting Visual Appeal (VA), the medium shade Usability (Usab), and the darkest shade Prompt Adherence (PA); these abbreviations (PA, Usab, VA) label the ribbons in the legend. Ribbon width is uniform and does not encode uncertainty.
All twelve ribbons exhibit an upward trend from Ideation to Refinement. The wide dispersion observed at Ideation, spanning approximately 3.0 to 4.2, converges into a substantially narrower band of roughly 3.9 to 4.4 by Refinement. Claude-Opus on Prompt Adherence is the strongest performer throughout, beginning at 4.23 and reaching 4.43 at Refinement. The lowest-scoring models at Ideation, GPT-5.3 and Qwen3.5-397b at approximately 3.0 to 3.2, demonstrate the largest gains across phases, closing much of the initial gap by Refinement. The dashed line denotes the neutral midpoint of 3.0.}
    \label{fig:lp-scalar-main-appendix}
\end{figure}

\begin{figure}[H]
    \centering
    \includegraphics[width=1\textwidth]{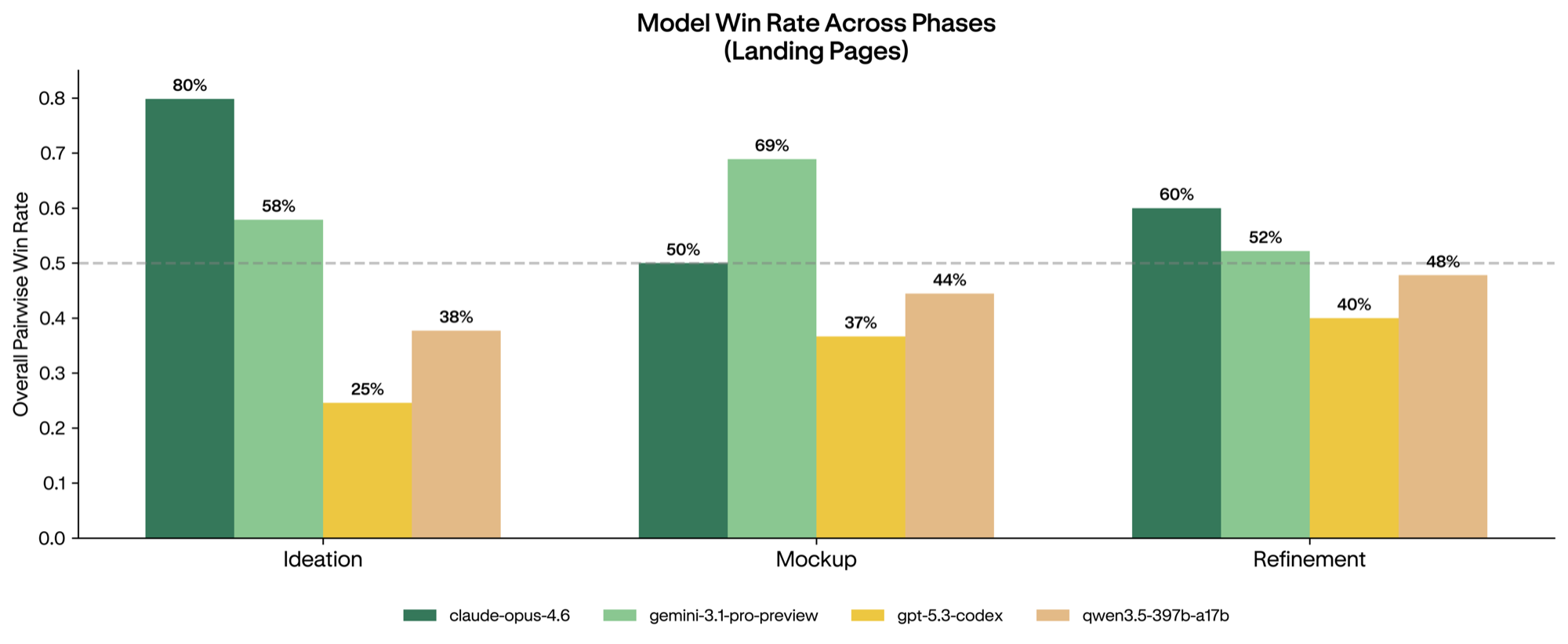}
    % \caption{Overall pairwise win rates for Landing Pages by phase.}
    \caption{Model win rates across the three phases for landing-page generation. Claude-Opus-4.6 leads Ideation at 80\%, followed by Gemini-3.1-Pro-Preview at 58\%, Qwen3.5-397b-a17b at 38\%, and finally GPT-5.3-Codex at 25\%. In Mockup, Gemini-3.1-Pro-Preview leads at 69\%, followed by Claude-Opus-4.6 at 50\%, Qwen3.5-397b-a17b at 44\%, and GPT-5.3-Codex at 37\%. By Refinement, Claude-Opus-4.6 leads again at 60\%, followed by Gemini-3.1-Pro-Preview at 52\%, Qwen3.5-397b-a17b at 48\%, and GPT-5.3-Codex at 40\%.}
    \label{fig:lp-overall-win}
\end{figure}

\begin{figure}[H]
    \centering
    \includegraphics[width=1\textwidth]{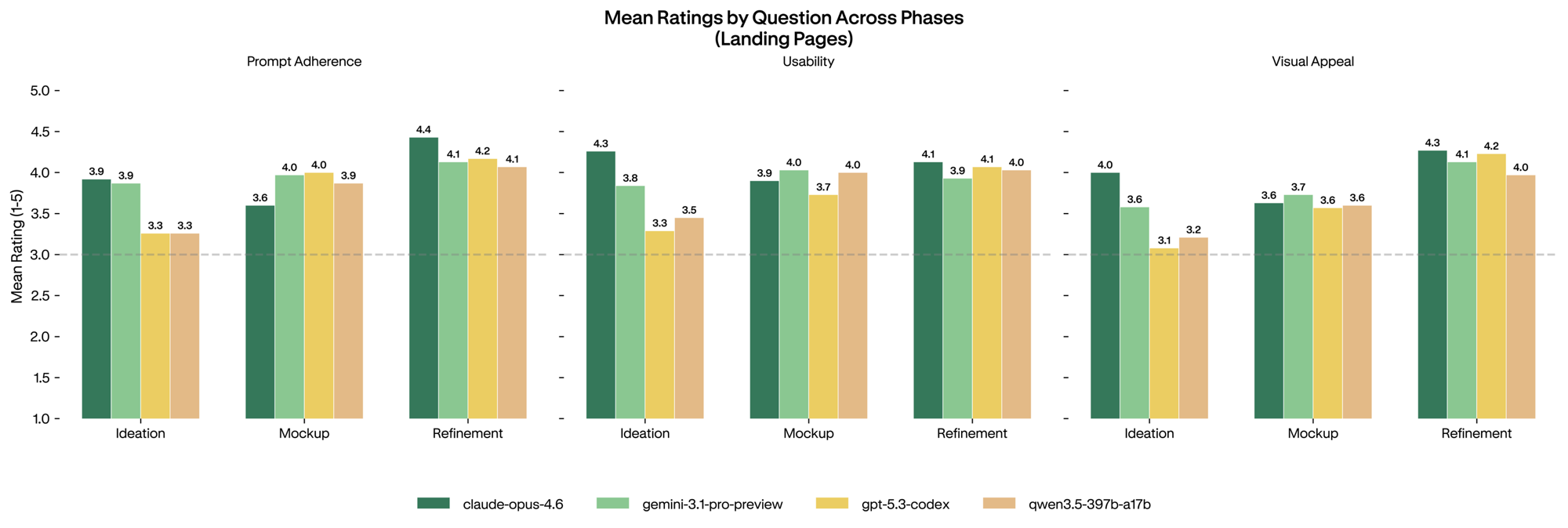}
    % \caption{Mean scalar ratings (1--5) for Landing Page deliverables across dimensions and phases.}
    \caption{Mean scalar ratings (1–5) for landing-page deliverables across the three pipeline stages, broken out by evaluation question. The dashed line marks the neutral midpoint (3.0).
Prompt Adherence. Claude-Opus-4.6 ties Gemini-3.1-Pro-Preview for the Ideation lead at 3.9 (ahead of GPT-5.3-Codex and Qwen3.5-397b-a17b, both 3.3), dips to last in Mockup at 3.6 while the others sit at 3.9–4.0, then jumps to the top in Refinement at 4.4, followed by GPT-5.3-Codex (4.2), and Gemini-3.1-Pro-Preview and Qwen3.5-397b-a17b (both 4.1). Gemini-3.1-Pro-Preview is the most stable, holding 3.9 $\rightarrow$ 4.0 $\rightarrow$ 4.1 across the three stages.
Usability. Claude-Opus-4.6 starts strongest in Ideation at 4.3, ahead of Gemini-3.1-Pro-Preview (3.8), Qwen3.5-397b-a17b (3.5), and GPT-5.3-Codex (3.3). The field tightens in Mockup, where Gemini-3.1-Pro-Preview and Qwen3.5-397b-a17b lead at 4.0, followed by Claude-Opus-4.6 (3.9) and GPT-5.3-Codex (3.7). By Refinement, Claude-Opus-4.6 and GPT-5.3-Codex tie at 4.1, ahead of Qwen3.5-397b-a17b (4.0) and Gemini-3.1-Pro-Preview (3.9).
Visual Appeal. Claude-Opus-4.6 leads Ideation at 4.0, followed by Gemini-3.1-Pro-Preview (3.6), Qwen3.5-397b-a17b (3.2), and GPT-5.3-Codex (3.1). In Mockup the models bunch together, with Gemini-3.1-Pro-Preview just ahead at 3.7 and Claude-Opus-4.6, GPT-5.3-Codex, and Qwen3.5-397b-a17b all at 3.6. In Refinement, Claude-Opus-4.6 leads at 4.3, followed by GPT-5.3-Codex (4.2), Gemini-3.1-Pro-Preview (4.1), and Qwen3.5-397b-a17b (4.0).}
    \label{fig:lp-mean-scalar}
\end{figure}

\begin{figure}[htbp]
    \centering
    \includegraphics[width=1\textwidth]{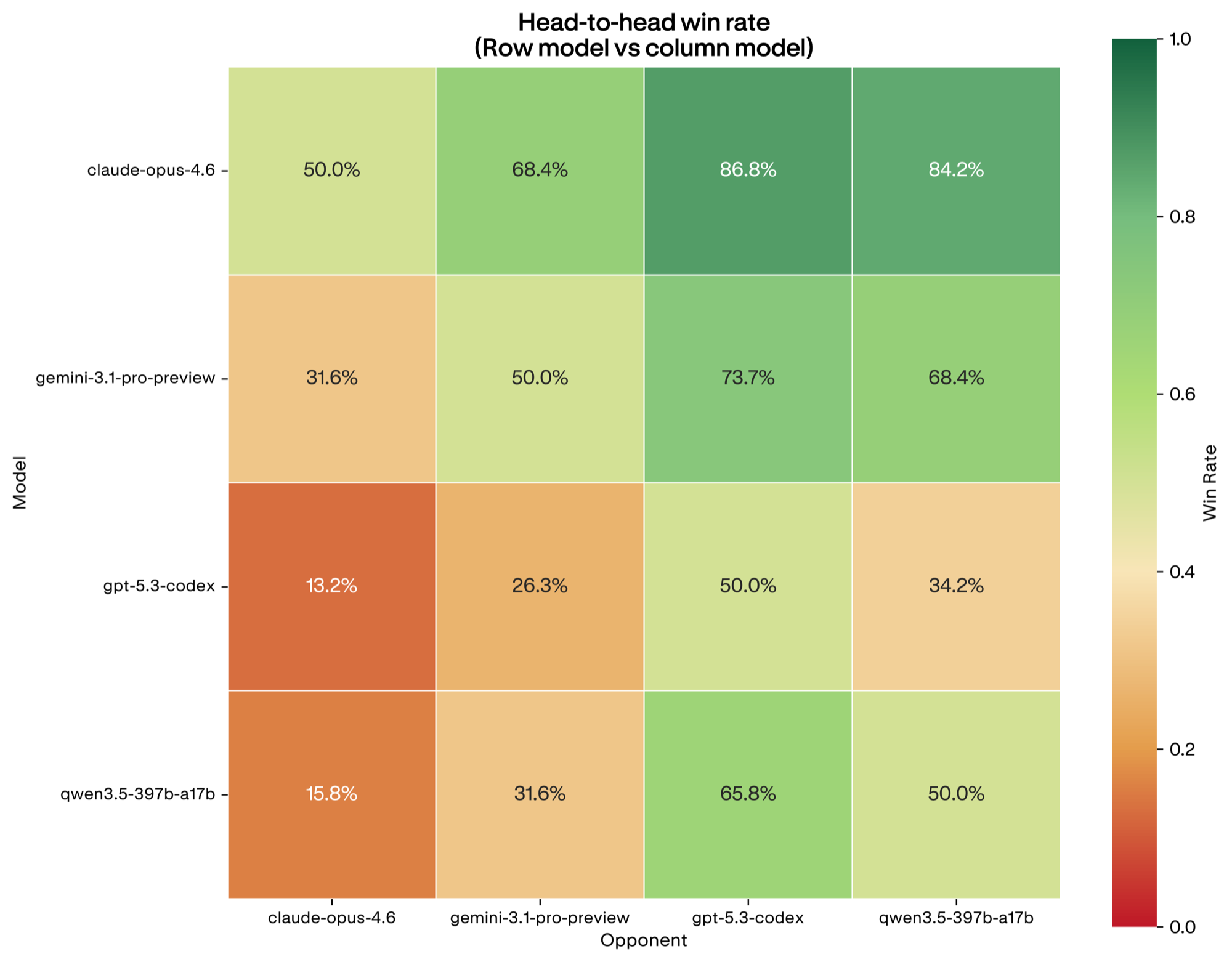}
    \caption{Head-to-head pairwise win rates for landing-page generation in the Ideation stage. Each cell reports the row model's win rate against the column model; diagonal self-matches are excluded.}
    \label{fig:lp-h2h-ideation}
\end{figure}

\begin{figure}[htbp]
    \centering
    \includegraphics[width=1\textwidth]{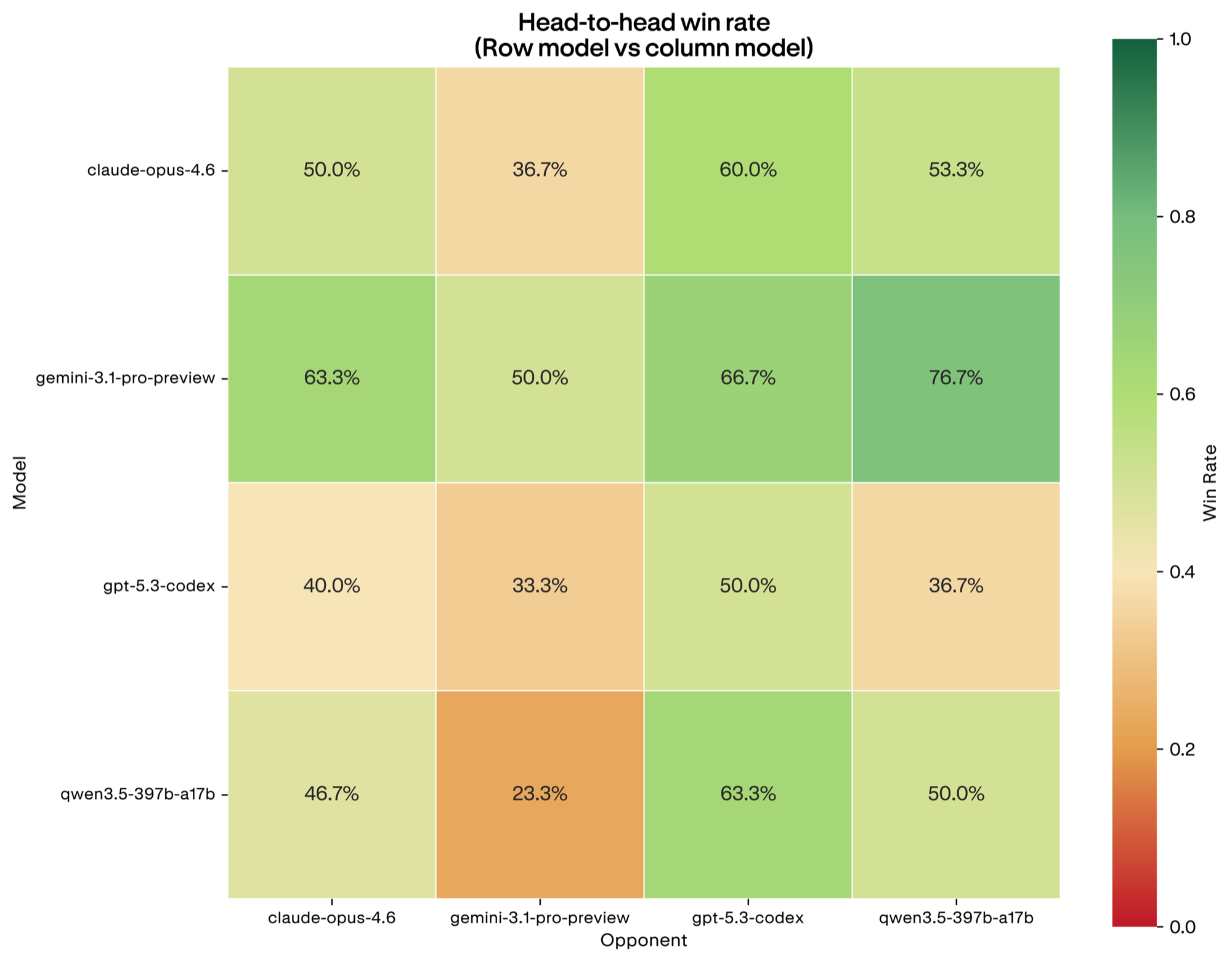}
    \caption{Head-to-head pairwise win rates for landing-page generation in the Mockup stage. Each cell reports the row model's win rate against the column model; diagonal self-matches are excluded.}
    \label{fig:lp-h2h-mockup}
\end{figure}

\begin{figure}[htbp]
    \centering
    \includegraphics[width=1\textwidth]{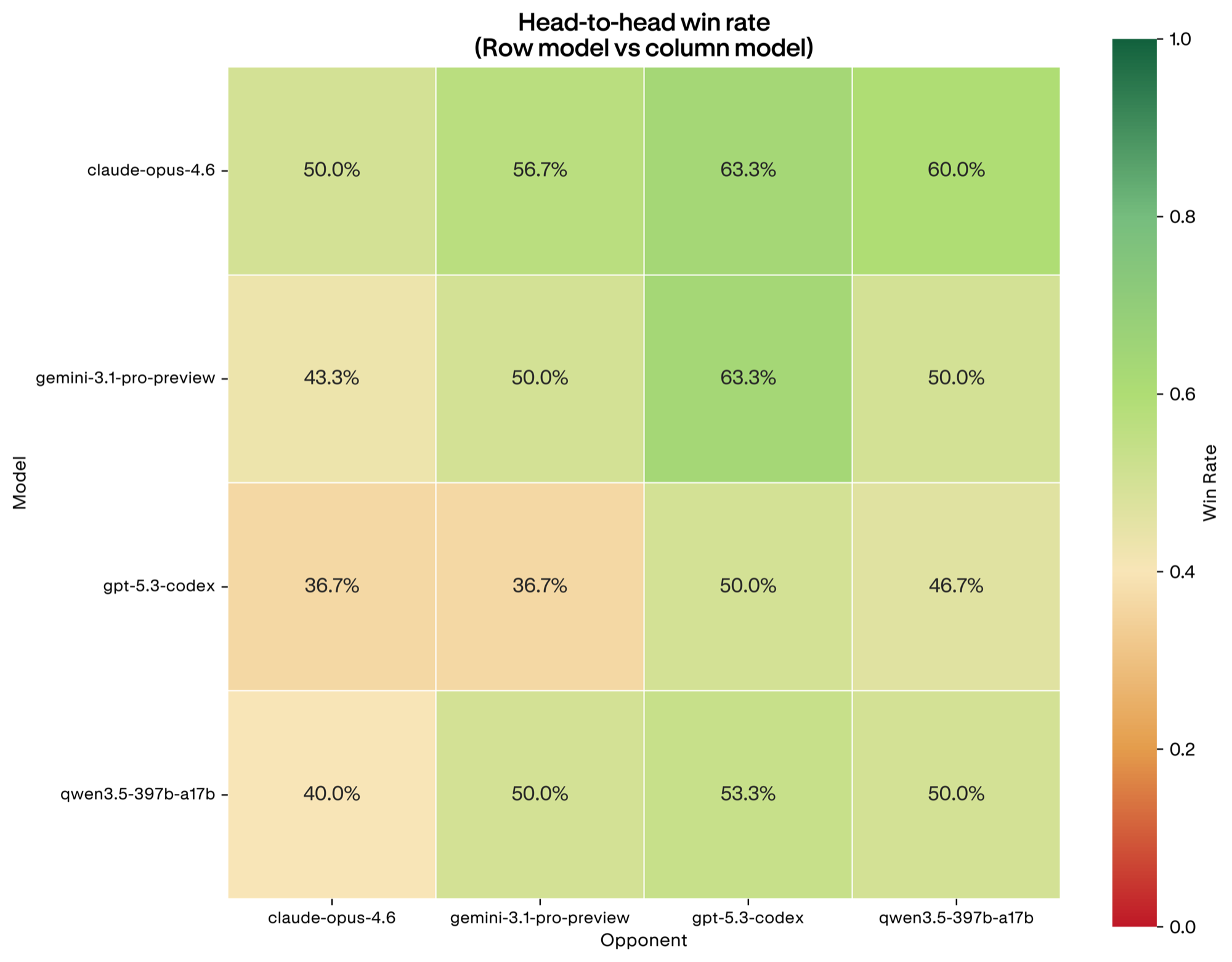}
    \caption{Head-to-head pairwise win rates for landing-page generation in the Refinement stage. Each cell reports the row model's win rate against the column model; diagonal self-matches are excluded.}
    \label{fig:lp-h2h-refinement}
\end{figure}

\begin{figure}[htbp]
    \centering
    \includegraphics[width=1\textwidth]{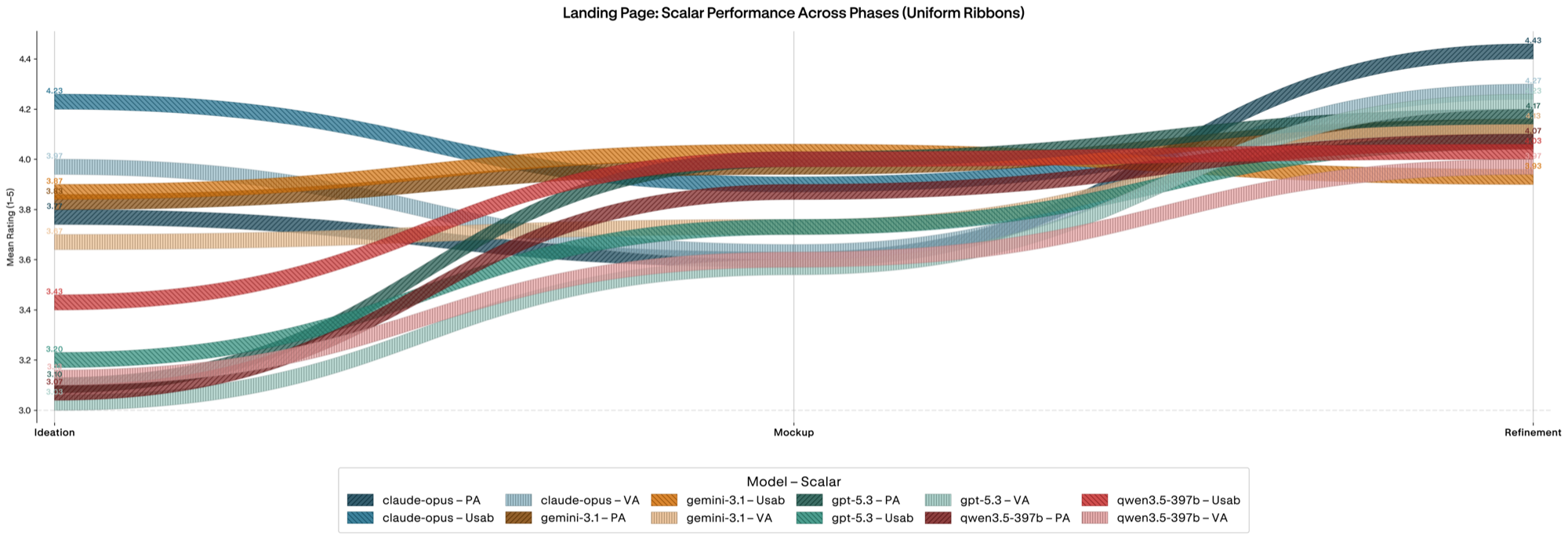}
    % \caption{Scalar performance ribbons for Landing Page generation across the creative workflow. The chart tracks mean ratings for Prompt Adherence (PA), Usability (Usab), and Visual Appeal (VA). }
    \caption[Scalar performance across landing-page phases]{Scalar performance across the three phases for landing-page generation, comprising one ribbon for each model–metric combination. Color encodes the model (Claude-Opus in blue, Gemini-3.1 in orange, GPT-5.3 in teal, and Qwen3.5-397b in red), while shade encodes the evaluation metric, with the lightest ribbon denoting Visual Appeal (VA), the medium shade Usability (Usab), and the darkest shade Prompt Adherence (PA); these abbreviations (PA, Usab, VA) label the ribbons in the legend. Ribbon width is uniform.
The wide dispersion observed at Ideation, spanning approximately 3.0 to 4.2, narrows into a tighter band of roughly 3.9 to 4.4 by Refinement. Claude-Opus on Prompt Adherence records the highest score at both Ideation (4.23) and Refinement (4.43), though it falls toward the middle of the range at Mockup before climbing back. The lowest-scoring models at Ideation, GPT-5.3 and Qwen3.5-397b at approximately 3.0 to 3.2, show the largest gains across phases, closing much of the gap by Refinement. The dashed line denotes the neutral midpoint of 3.0.}
    \label{fig:lp-scalar-ribbons-appendix}
\end{figure}

Figures~\ref{fig:lp-scalar-main-appendix}, \ref{fig:lp-overall-win}, \ref{fig:lp-mean-scalar}, and~\ref{fig:lp-scalar-ribbons-appendix} summarize scalar trajectories, win rates, mean ratings, and phase ribbons for landing-page generation, and Figures~\ref{fig:lp-h2h-ideation}, \ref{fig:lp-h2h-mockup}, and~\ref{fig:lp-h2h-refinement} give head-to-head pairwise win rates for the Ideation, Mockup, and Refinement stages.

\begin{figure*}[t]
    \centering
    \includegraphics[width=\textwidth]{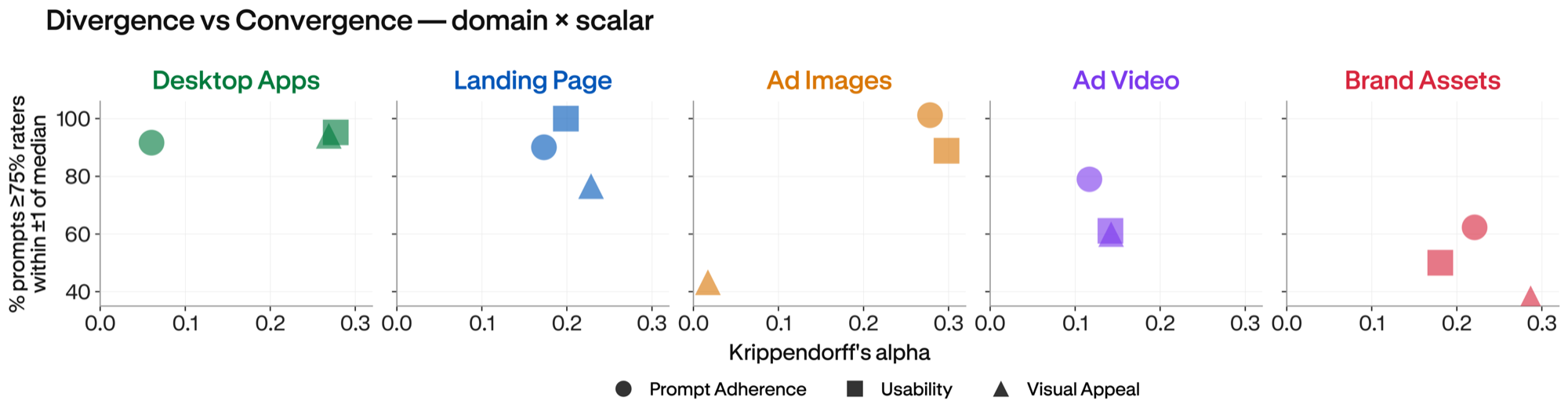}
    \caption{Spread of axes agreement across domains.}
    \label{fig:axes_domains}
\end{figure*}

\end{document}